\documentclass{article}
\usepackage{arxiv}
\usepackage[usenames]{color}
\usepackage[ansinew]{inputenc}
\usepackage{inputenc}
\usepackage{graphicx}
\usepackage{amsfonts}
\usepackage{amssymb}
\usepackage{amsbsy}
\usepackage{amsmath}
\usepackage{multirow}
\usepackage{array}
\usepackage{color}
\usepackage{soul}
\usepackage{cite}
\usepackage[normalem]{ulem} 
\usepackage{times}
\usepackage[all]{xy}
\usepackage{lscape}
\usepackage{flushend}
\usepackage{url}
\usepackage{float}
\usepackage{rotating}
\usepackage[dvipsnames]{xcolor}
\usepackage{colortbl}
\usepackage{hyperref}

\usepackage[caption = false]{subfig}
\usepackage{xspace}

\graphicspath{{./images/}}
\DeclareGraphicsExtensions{.pdf,.png}

\def\mH{\mathcal{H}}

\newcommand{\bphi}{\boldsymbol{\phi}}

\def\Real{\mathbb{R}}

\def\C{\ensuremath{\mathbf{C}}}

\def\K{\ensuremath{\mathbf{K}}}

\def\X{\ensuremath{\mathbf{X}}}
\def\mX{\ensuremath{\mathcal{X}}}
\def\mH{\ensuremath{\mathcal{H}}}

\def\a{\ensuremath{\mathbf{a}}}

\def\x{\ensuremath{\mathbf{x}}}
\def\y{\ensuremath{\mathbf{y}}}
\def\z{\ensuremath{\mathbf{z}}}

\newcommand{\red}[1]{\textcolor[rgb]{0.5,0,0}{#1}}           

\newcommand{\degman}{\ensuremath{^\circ}\xspace}

\hypersetup{
    bookmarks=true,         
    unicode=false,          
    pdftoolbar=true,        
    pdfmenubar=true,        
    pdffitwindow=false,     
    pdfstartview={FitH},    
    pdftitle={KACD},    
    pdfauthor={Jose A. Padron},     
    pdfcreator={JAP}   
    pdfproducer={JAP}, 
    pdfkeywords={}, 
    pdfnewwindow=true,      
    colorlinks=true,       
    linkcolor={blue},          
    citecolor=blue,        
    filecolor=blue,      
    urlcolor=blue           
}

\title{Kernel Anomalous Change Detection\\ for Remote Sensing Imagery}
\author{
  Jos\'e A. Padr\'on-Hidalgo\\
  Image Processing Laboratory \\
  Universitat de Val{\`e}ncia\\
  Val{\`e}ncia, Spain\\
  \texttt{joseantoniopadronhidalgo@gmail.com} \\
  \And
  Valero Laparra\\
  Image Processing Laboratory \\
  Universitat de Val{\`e}ncia\\
  Val{\`e}ncia, Spain\\
  \texttt{valero.laparra@uv.es} \\
  \And
  Nathan Longbotham \\
  Descartes Lab \\
  Santa Fe, NM, USA\\
  \texttt{nathan@descarteslabs.com} \\
  \And
  Gustau Camps-Valls \\
  Image Processing Laboratory \\
  Universitat de Val{\`e}ncia\\
  Val{\`e}ncia, Spain\\
  \texttt{gcamps@uv.es} \\
}

\begin{document}

\begin{center}
©IEEE. ACCEPTED FOR PUBLICATION IN IEEE TGARS 2019. DOI 10.1109/TGRS.2019.2916212\footnote{
©IEEE. Personal use of this material is permitted.  Permission from IEEE must be obtained for all other users,including reprinting/republishing this material for advertising or promotional purposes, creating new collective works for resale or redistribution to servers or lists, or reuse of any copyrighted components of this work in other works.  DOI: 10.1109/TGRS.2019.2916212.
}
\end{center}
\maketitle

\begin{abstract}
Anomalous change detection is an important problem in remote sensing image processing. Detecting not only pervasive but anomalous or extreme changes has many applications for which methodologies are available. This paper introduces a nonlinear extension of a full family of anomalous change detectors based on covariance operators.  
In particular, this paper focuses on algorithms that utilize Gaussian and elliptically contoured distributions, and extend them to their nonlinear counterparts based on the theory of reproducing kernels in Hilbert spaces. 
The presented methods generalize their linear counterparts, based on the assumption of either Gaussian or elliptically-contoured distribution. 
We illustrate the performance of the introduced kernel methods in both pervasive and anomalous change detection problems involving both real and simulated changes in multi and hyperspectral imagery with different resolutions (AVIRIS, Sentinel-2, WorldView-2, Quickbird). A wide range of situations are studied, involving droughts, wildfires, and urbanization in real examples. 
Excellent performance in terms of detection accuracy compared to linear formulations is achieved, resulting in improved detection accuracy and reduced false alarm rates.
Results also reveal that the elliptically-contoured assumption may be still valid in Hilbert spaces. 
We provide an implementation of the algorithms as well as a database of natural anomalous changes in real scenarios. 
\end{abstract}


\section{Introduction}
\label{sec:intro}

The problem of change detection deals with identifying transitions between a pair (or a series) of co-registered images~\cite{Sin89,Rad05}. Change detection in remote sensing images is of paramount relevance because it automatizes traditionally manual tasks in disaster management (floods, droughts and wildfires), helps in designing development and settlement plans as well as in urban and crop monitoring. Multitemporal classification and change detection are very active fields nowadays because of the increasing availability of complete time series of images and the interest in monitoring changes occurring on the Earth's cover due to either natural or anthropogenic activities. Complete constellations of civil and military satellites sensors currently provide with high spatial resolution and revisiting frequency. The Copernicus' Sentinels\footnote{\url{http://www.esa.int/esaLP/SEMZHM0DU8E_LPgmes_0.html}} or NASA's A-train\footnote{\url{http://www.nasa.gov/mission_pages/a-train/a-train.html}} programs are producing near real-time coverage of the globe. NASA is currently producing a Harmonized Landsat Sentinel-2 (HLS) data set, which can be used for monitoring agricultural resources with an unprecedented combination of 30 m spatial resolution and 2-3 days revisit. In parallel, new commercial satellite missions are being deployed to provide multispectral data at both high spatial and high temporal resolution. For example, the PlanetScope constellation by Planet Labs, Inc. can provide 5 m data daily for sites requested by the client, and the recently announced UrtheDaily constellation, specifically designed for operational agricultural applications, will acquire S2-like data also at 5-m spatial resolution and with an impressive full global coverage every day. It goes without saying that closed-range applications using drones and all kind of unmanned automated vehicles (UAVs) also challenge the field of automatic change detection. All in all, automatic image analysis in general and change detection in particular are becoming strictly necessary in the current era of data deluge.


The field of {\bf change detection} (CD) methods is vast and many approaches are available in the literature~\cite{Lu04,Radke,Manolakis2,Manolakis03}. A simple taxonomy could organize them according to three types of products~\cite{Sin89,Cop04}: 1) binary maps, 2) detection of types of changes, and 3) full multiclass change maps, thus including classes of changes and unchanged land-cover classes. Each type of product can be achieved using different sources of information retrieved from the initial spectral images at time instants $t_1$ and $t_2$. Unsupervised CD has been widely studied, mainly because it meets the requirements of most applications: i) the speed in retrieving the change map and ii) the absence of labeled information in general applications. However, the lack of labeled information makes the problem of detection more difficult and thus unsupervised methods typically consider binary change detection problems only. 
In the last decade, change vector analysis (CVA) techniques have been widely applied: CVA converts the difference image to polar coordinates and operate in such representation space~\cite{Mal80,Bov07}. In~\cite{Mur08}, morphological operators were successfully applied to increase the discriminative power of the CVA method. In~\cite{Bov09}, a contextual parcel-based multiscale approach to unsupervised CD was presented. Traditional CVA relies on the experience of the researcher for the threshold definition, and is still on-going research~\cite{Im08,Che11}. The method has been also studied in terms of sensitivity to differences in registration and other radiometric factors~\cite{Bov09b}. Another interesting approach based on spectral transforms is the multivariate alteration detection (MAD)~\cite{Nie98,Nie06}, where canonical correlation is computed for the points at each time instant and then subtracted. The method consequently reveals changes invariant to linear transformations between the time instants. Radiometric normalization issues for MAD has been recently considered in~\cite{Can08}, and nonlinear extensions have been also realized via kernel machines (KM)~\cite{Nielsen15,gomezchovaigarss11ksnr}. Other approaches based on KM have proposed to use dimensionality reduction via principal components~\cite{Ding} or slow features~\cite{WUandZhang} of the difference image.
Clustering has been used in recent binary CD. In~\cite{Cel09}, rather than converting the difference image in the polar domain, local PCAs are used in sub-blocks of the image, followed by a binary $k$-means clustering to detect changed/unchanged areas locally. Kernel clustering has been also studied in~\cite{Vol10c,Volpi}, where kernel $k$-means with parameters optimized using an unsupervised ANOVA-like cost function is used to separate the two clusters in a fully unsupervised way. 
Finally, unsupervised neural networks have been considered for binary CD~\cite{Pac09,Pac10}. In~\cite{Gho07}, a Hopfield neural network, where each neuron is connected to a single pixel is used to enforce neighborhood relationships. Lately, many efforts have been conducted in using deep convolutional neural networks as well~\cite{liu,CHA,Ouyang}.

When the signature of the class to be detected is unknown, {\bf anomaly detection} (AD) algorithms are more often used as an alternative to traditional CD. AD reduces to assuming (and/or defining) a model for the background class, and then looking for the test samples lying far from the mean of the background distribution. Depending on the assumptions made on the background, different anomaly detectors can be used. If the background is modeled as a Gaussian distribution, the Reed-Xiaoli (RX)~\cite{Ree90,Lu97} detector can be used. The RX detector identifies outliers by computing the Mahalanobis distance between the considered pixel and the background mean vector. If the background is believed to be more complex, for instance multi-Gaussian, it can be modeled with a Gaussian mixture model~\cite{Bea00}. Application of these principles to the detection of small objects in series of remote sensing images can be found in~\cite{Car05}. The RX detector has been extended to its nonlinear kernelized version in~\cite{Kwo05}, and many variants have been introduced in the last years~\cite{Kwan}. 


A related field of investigation in this direction is the so-called {\bf anomalous change detection} (ACD)~\cite{The08b}: in this field, one looks for changes that are interestingly anomalous in multitemporal series of images, and try to highlight them in contrast to acquisition condition changes, registration noise, or seasonal variation. 
The interest in ACD is high, and many methods have been proposed in the literature, ranging from regression-based approaches like in the chronocrome~\cite{Schaum97} where big (`influential') residuals are associated with anomalies~\cite{Behnken,Cook}, to equalization-based approaches that rely on whitening principles~\cite{Mayer03}, as well as multivariate methods~\cite{Arenas13} that reinforce directions in feature spaces associated with noisy or rare events~\cite{Green88,Nielsen98}. 
The work~\cite{Theiler06} formalized the field by introducing a framework for ACD, which assumes Gaussianity, yet the derived detector delineates hyperbolic decision functions. Even though the Gaussian assumption reports some advantages (e.g. tractability and generally good performance) it is still an {\em ad hoc} assumption that it is not necessarily fulfilled in practice. This is the motivation in~\cite{Theiler10}, where the authors introduced elliptically-contoured (EC) distributions that generalize the Gaussian distribution and proved more appropriate to modeling fatter tail distributions and thus detect anomalies more effectively. 
The EC decision functions are point-wise nonlinear and still rely on second-order feature relations. 
Recent advances in ACD have considered methods robust to pixel misregistration~\cite{Theiler5} and sequences of several images in RX and chronocrome settings~\cite{Theiler7}. 

In this paper, we extend the family of ACD methods in~\cite{Theiler10} to cope with higher-order feature relations through the theory of reproducing kernels. 
Kernel methods allow the generalization of algorithms that are expressed in terms of dot products to account for higher-order (nonlinear) feature relations, yet still relying on linear algebra~\cite{shawetaylor04,CampsValls09wiley,Rojo17dspkm}. 
We illustrate the performance of the introduced kernel ACD methods in different experiments involving synthetic, artificially enforced and natural anomalous changes in multi- and hyperspectral imagery with different spatial resolutions (AVIRIS, Sentinel-2, WorldView-2, Quickbird). A wide range of situations are studied, involving droughts, wildfires, and urbanization in real examples. {Very good performance is achieved in terms of detection accuracy compared to the linear formulations.} 
Results also reveal that the elliptically-contoured assumption may be still valid in Hilbert spaces, even when high pervasive distortions mask anomalous targets. 

The rest of the paper is outlined as follows. Section~\ref{sec:kacd} reviews the family of (Gaussian and elliptically-contoured) ACD algorithms introduced in~\cite{Theiler10}, and introduces the proposed nonlinear (kernel-based) versions. The introduced methods generalize the previous ones for the linear kernel, and provide more flexible mappings to account for higher-order dependencies between features. 
Section~\ref{sec:experiments} presents experiments comparing the performance of the proposed algorithms with their linear counterpart in different scenarios. Finally, section~\ref{sec:conclusions} concludes the paper.

\section{Proposed Family of Kernel Anomaly Change Detection Methods}\label{sec:kacd}


An anomaly can be loosely defined as a sample with small probability to occur. An anomalous change is thus a rare, unexpected, change between two consecutive observations of the same phenomenon. 
In this paper we want to find samples that can be interpreted as anomalous changes between two multidimensional images. This calls for studying and characterizing differences between multivariate distributions, and in particular those features that account for the anomalous changes. In~\cite{Theiler10} a framework to define different anomalous change detectors based on probability distributions was formalized.

Let treat the two images as random variables $X$ and $Y$, with probability distributions $\mathbb{P}_X$ and $\mathbb{P}_Y$ respectively. Let us indicate the joint distribution as $\mathbb{P}_{X,Y}$ which accounts for how probable particular joint events are. The marginal distributions $\mathbb{P}_X$ and $\mathbb{P}_Y$ can be used in order to mitigate the problem of detecting a joint anomaly as an anomalous change. Given two pixels $\x_i,\y_i\in\Real^d$ from the same spatial location $i$ but each one from one image, the general formula to compute the amount of anomalousness of a change is:
\begin{equation}
    {\mathcal A^*}_{X,Y}(\x_i,\y_i)= \frac{\mathbb{P}_{X}(\x_i) \mathbb{P}_Y(\y_i)}{\mathbb{P}_{X,Y}(\x_i,\y_i)}.
    \label{eq:probability}
\end{equation}
A sample is detected as anomalous change when it is anomalous with respect to the joint distribution but not anomalous with respect to the distributions of each isolated image. We are using here all three distributions, however different combinations can be used as we will see below.

Usually equation \ref{eq:probability} is applied by taking logarithms, ${\mathcal A}_{[X,Y]}(\x_i,\y_i) = \log({\mathcal A^*}_{[X,Y]}(\x_i,\y_i))$, which can be interpreted in information theoretic terms by noting the relation between probability and information. Elaborating on Shannon's information~\cite{Shannon48} we have: 
$${\mathcal A}_{X,Y}(\x_i,\y_i)= I_{X,Y}(\x_i,\y_i) - I_X(\x_i) - I_Y(\y_i),$$
where $I_A(B_i)$ is the amount of information in Shannon's terms the sample $B_i$ provides assuming it follows the distribution $\mathbb{P}_{A}$. A sample could be interpreted as an anomalous change if the information obtained by observing both images simultaneously gives high information with respect to the information given by observing each isolated image.

\subsection{Linear ACD algorithms}

Assuming that all three distributions follow a {\em multivariate Gaussian} we can express the formula only in terms of covariance matrices. The amount of {\em anomalousness} is given by:
\begin{equation}
  {\mathcal A}_{\mathcal G}(\x_i,\y_i)=\xi(\z_i) - \beta_x\xi(\x_i) - \beta_y\xi(\y_i),  
  \label{eq:ACDgauss}
\end{equation}
where $\xi(\a) = \a^\top\C_a^{-1}\a$, $\C_a$ is the estimated covariance matrix with the available data, and being $\z=[\x,\y]\in\Real^{2d}$. The value of $\beta_x,\beta_y\in\{0,1\}$ parameters defines which distributions are taken into account to define our anomaly. The different combinations give rise to different anomaly detectors (see Table~\ref{family}). These methods and some variants have been widely used in many hyperspectral image analysis settings because of its simplicity and generally good performance~\cite{Reed90,Chang02,Kwon03}. 

However, these methods are hampered by a fundamental problem: the (typically strong) assumption of Gaussianity that is implicit in the formulation. 
Accommodating other data distributions may not be easy in general. Theiler et al.~\cite{Theiler10} introduced alternative ACD to cope with elliptically-contoured distributions~\cite{Cambanis81}: roughly speaking, the idea is to model the data using an {\em  elliptically-contoured (EC) distribution}. EC distributions are particularly convenient in the case of images \cite{Lyu08c}. In particular the formulation introduced in \cite{Theiler10} uses the multivariate Student's t-distribution, giving rise to the following formula for computing the amount of {\em EC anomalousness}:
\begin{eqnarray}\label{anomal}
\begin{array}{lll}
{\mathcal A}_{\text{EC}}(\x_i,\y_i) &=& (2d+\nu) \log\bigg(1+\dfrac{\xi(\z_i)}{\nu}\bigg) \\
&-& \beta_x (d+\nu) \log\bigg(1+\dfrac{\xi(\x_i)}{\nu}\bigg)\\
&-& \beta_y (d+\nu) \log\bigg(1+\dfrac{\xi(\y_i)}{\nu}\bigg),
\end{array}
\end{eqnarray}
where $\nu$ controls the shape of the Student's t-distribution: for $\nu\to\infty$ the solution approximates the Gaussian and for $\nu\to 0$ it diverges. 

\begin{table}[t!]
\small
\begin{center}
\caption{\small A family of ACD algorithms.}
\label{family}
\begin{tabular}{||l|c|c||}
\hline\hline
 \bf{ACD algorithm} &  $\beta_x$  & \bf{$\beta_y$}  \\
\hline\hline
RX  & 0 & 0\\
Chronocrome $\y|\x$  & 0 & 1\\
Chronocrome $\x|\y$  & 1 & 0\\
Hyperbolic ACD  & 1 & 1\\
\hline
\end{tabular}
\end{center}
\end{table}

An interesting particular case is the RX algorithm which brings to the same result for the Gaussian and the elliptical case (independently of the $\nu$ value). All extra operations applied by the EC formulation with regard to the Gaussian version are increasing monotonic functions which do not change the ordering of the values. Therefore, although the value of anomalousness are different (i.e. ${\mathcal A}_{\mathcal G}(\x_i,\y_i) \neq {\mathcal A}_{\text{EC}}(\x_i,\y_i)$), the values are sorted in the same way which makes the detection curves also equal. The same effect happen between the RX methods based on kernels proposed in the next section. 

\begin{figure*}[t!]
\setlength{\tabcolsep}{2pt}
\begin{center}
\begin{tabular}{ccc}
\includegraphics[width=6cm]{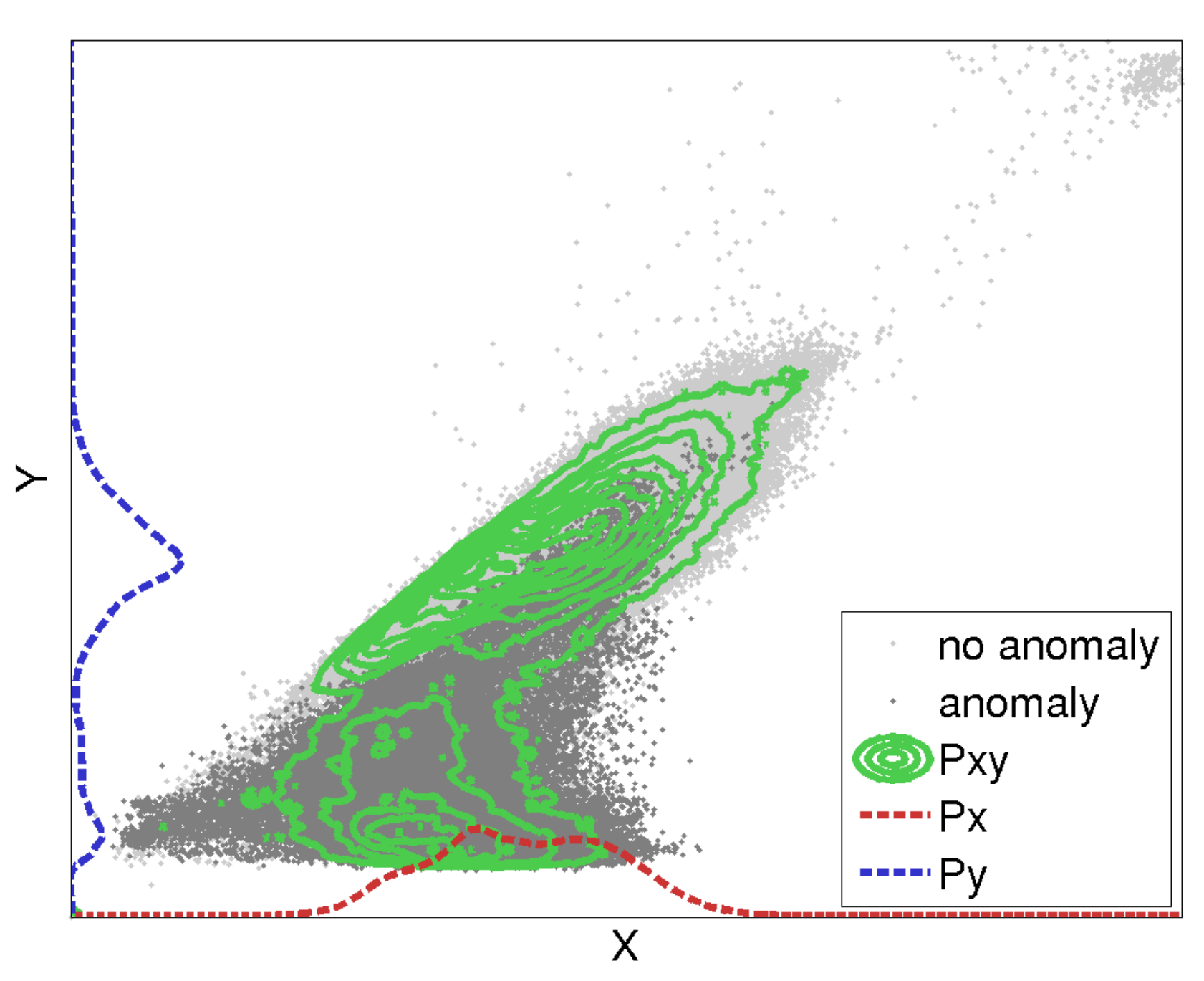} 
 & \includegraphics[width=6cm]{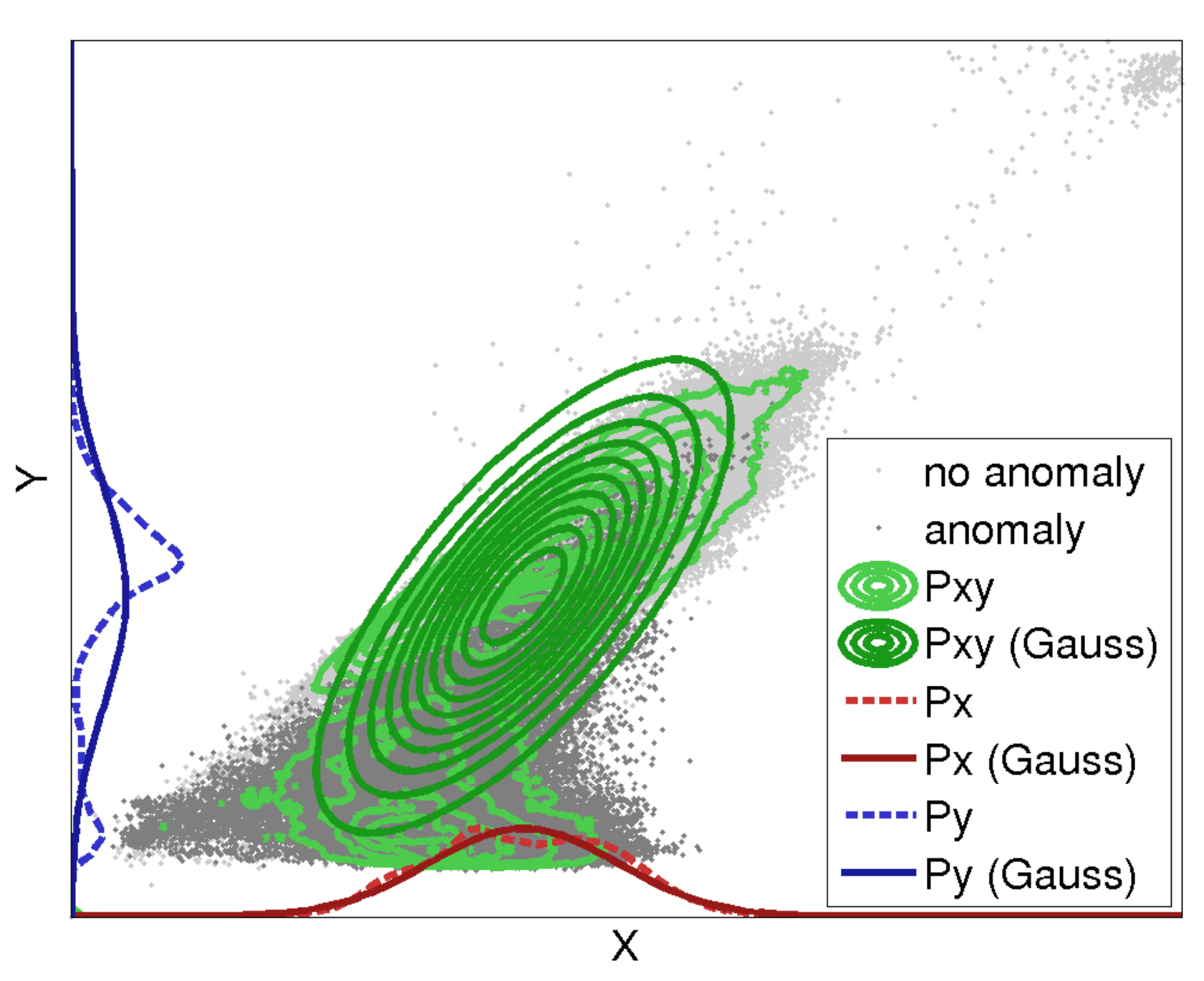} 
 & \includegraphics[width=6cm]{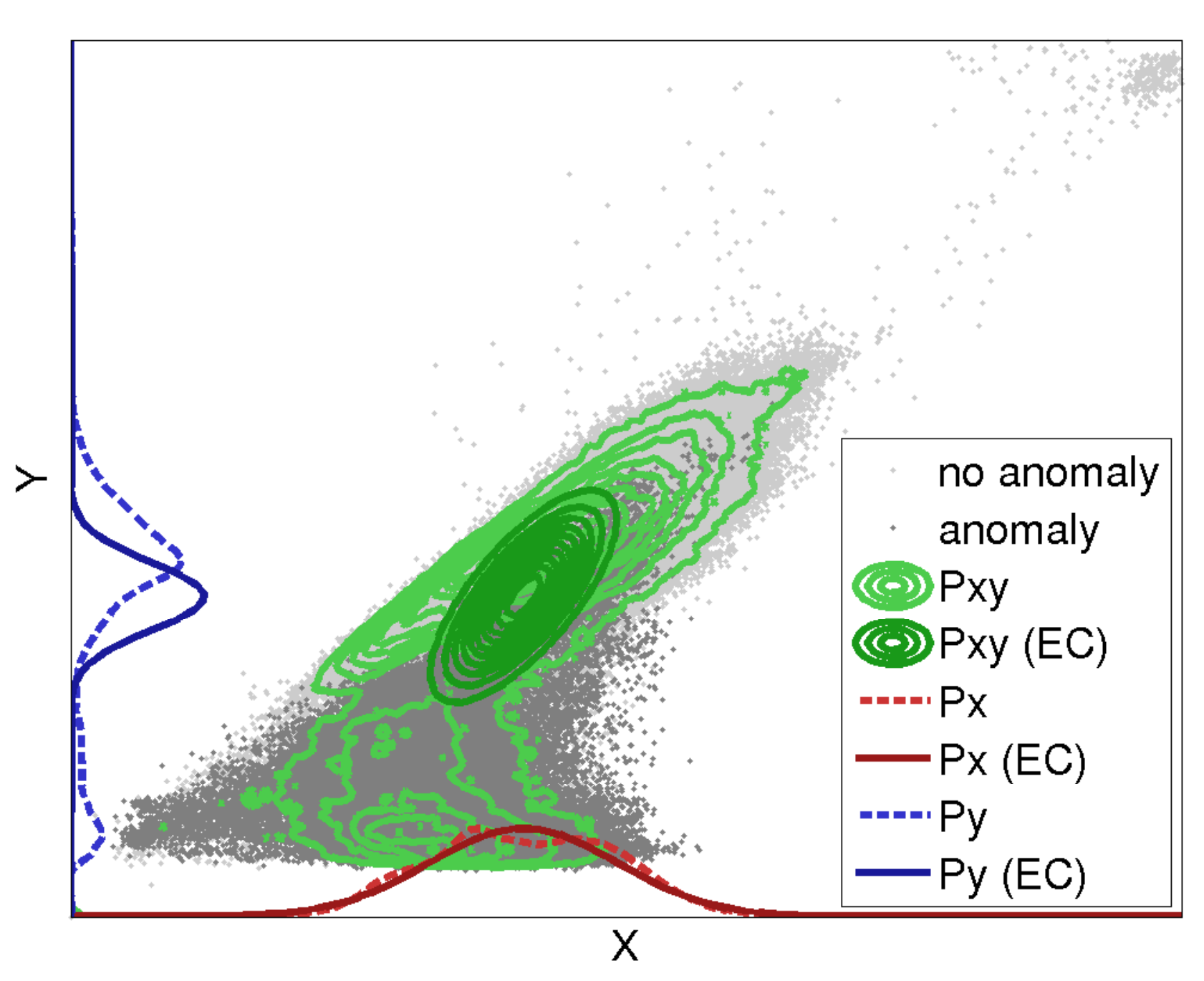} 
\end{tabular}
\end{center}
\caption{\small Illustration of ACD probabilistic framework. From left to right joint and marginal probability distributions of: the original data, Gaussian model, and Elliptically Contoured model. See text for details on the data.}
\label{fig:toy_ACD}
\end{figure*}

Figure \ref{fig:toy_ACD} shows an example of the distributions involved in the anomalous change detection framework. In order to be able to visualize the distributions we restrict ourselves to the most simple situation where each image contains just one band. In particular we show the distributions for the band $9$ of a Sentinel-2 image over Australia, see table \ref{ROC_table}. We show results for the distribution of the data estimated using histograms, and when assuming Gaussian or EC distributions. Note that the estimation of the distribution based on histograms is only feasible in the low dimensional (i.e. 2D) case: when the number of bands increases the computation of the histogram becomes unfeasible due to the curse of dimensionality. However, the Gaussian and the EC model can be estimated easily for multiple dimensions. The difference between the Gaussian and the EC model relies in the kurtosis of the distribution, while in the Gaussian case is fixed in the EC case can be controlled with the $\nu$ parameter. By comparing the marginal distributions in the central and the right panels we can easily spot the differences between the Gaussian and the EC model. For the horizontal axes the data follows quite well the Gaussian model, i.e. the red solid line and the dashed red line are very similar in the central panel. However the Gaussian model fails when reproducing the probability for the vertical axes (central panel blue lines). Although it is not a perfect model, the EC distribution allows to get better description of the vertical axes (right panel blue lines), and simultaneously obtains a similar (good) result as the Gaussian model in the horizontal axes right panel red lines).

\subsection{Kernel ACD algorithms}

Previous methods are linear and depend on estimating covariance matrices with the available data, and use them as a metric for testing anomalousness. These methods are fast to apply, delineate point-wise nonlinear decision boundaries, but still rely on second-order statistics. This restricts the class of functions that can be implemented and thus the generalization capabilities of the algorithm. For instance in Fig.~\ref{fig:toy_ACD} the assumed joint distributions (dark green) for both Gaussian and EC models clearly differ from the {\emph real} distribution (light green). We here address this issue through the theory of reproducing kernel functions~\cite{shawetaylor04}, which allow us to capture higher-order feature relations while still relying on linear algebra. Kernel methods are particularly robust to reduced sample sizes and high-dimensional feature spaces, situations often encountered in hyperspectral image detection problems.

As the covariance matrix, Kernel methods rely on the notion of {similarity} between points in a higher (possibly infinite) dimensional space. They assume the existence of a Hilbert space $\mH$ equipped with an inner product $\langle\cdot,\cdot\rangle_\mH$. Samples in $\mX$ are mapped into $\mH$ by means of a feature map $\bphi:\mX\rightarrow\mH, \x_i\mapsto \bphi(\x_i)$, $1\leq i\leq n$. 
The mapping function $\bphi$ can be defined explicitly or implicitly, which is usually the case in kernel methods. The similarity between the elements in $\mH$ can be estimated using its associated inner product $\langle\cdot,\cdot\rangle_{\mH}$ via reproducing kernels in Hilbert spaces, $k:\mX\times\mX\rightarrow\mathbb{R}$, such that pairs of points $(\x_i,\x_j)$ $\mapsto$ $k(\x_i,\x_j)$. So we can estimate similarities in $\mH$ without the explicit definition of the \emph{feature map} $\bphi$, and hence without the need of having access to the points in $\mH$. The {\em kernel function} $k$ requires to satisfy Mercer's Theorem~\cite{Aizerman64}.

\begin{figure*}[t!]
\setlength{\tabcolsep}{2pt}
\begin{center}
\begin{tabular}{cccc}
RX (0.92) & XY (0.54) & YX (0.81) & HACD (0.33)\\ 
\includegraphics[width=4cm]{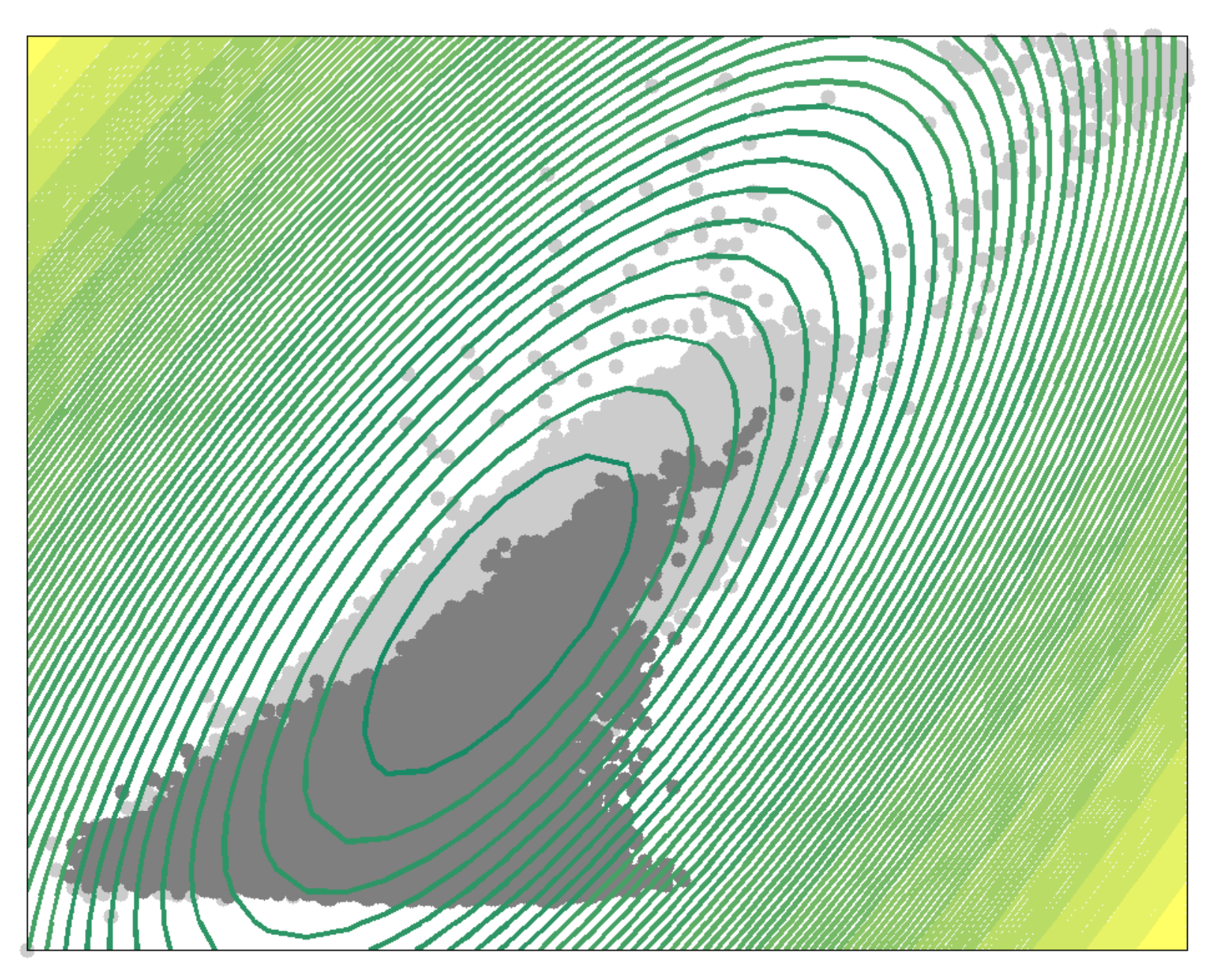} 
 & \includegraphics[width=4cm]{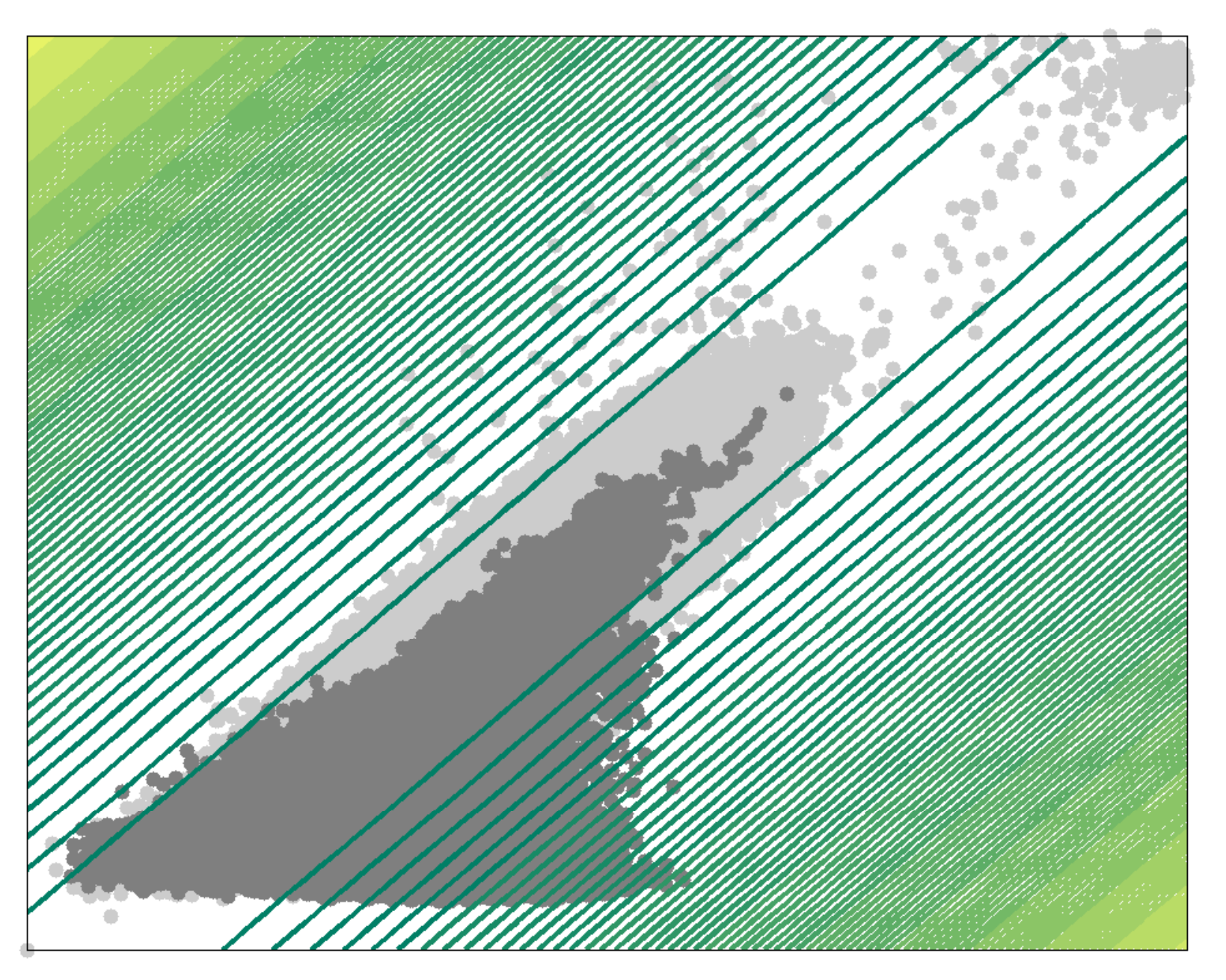} 
 & \includegraphics[width=4cm]{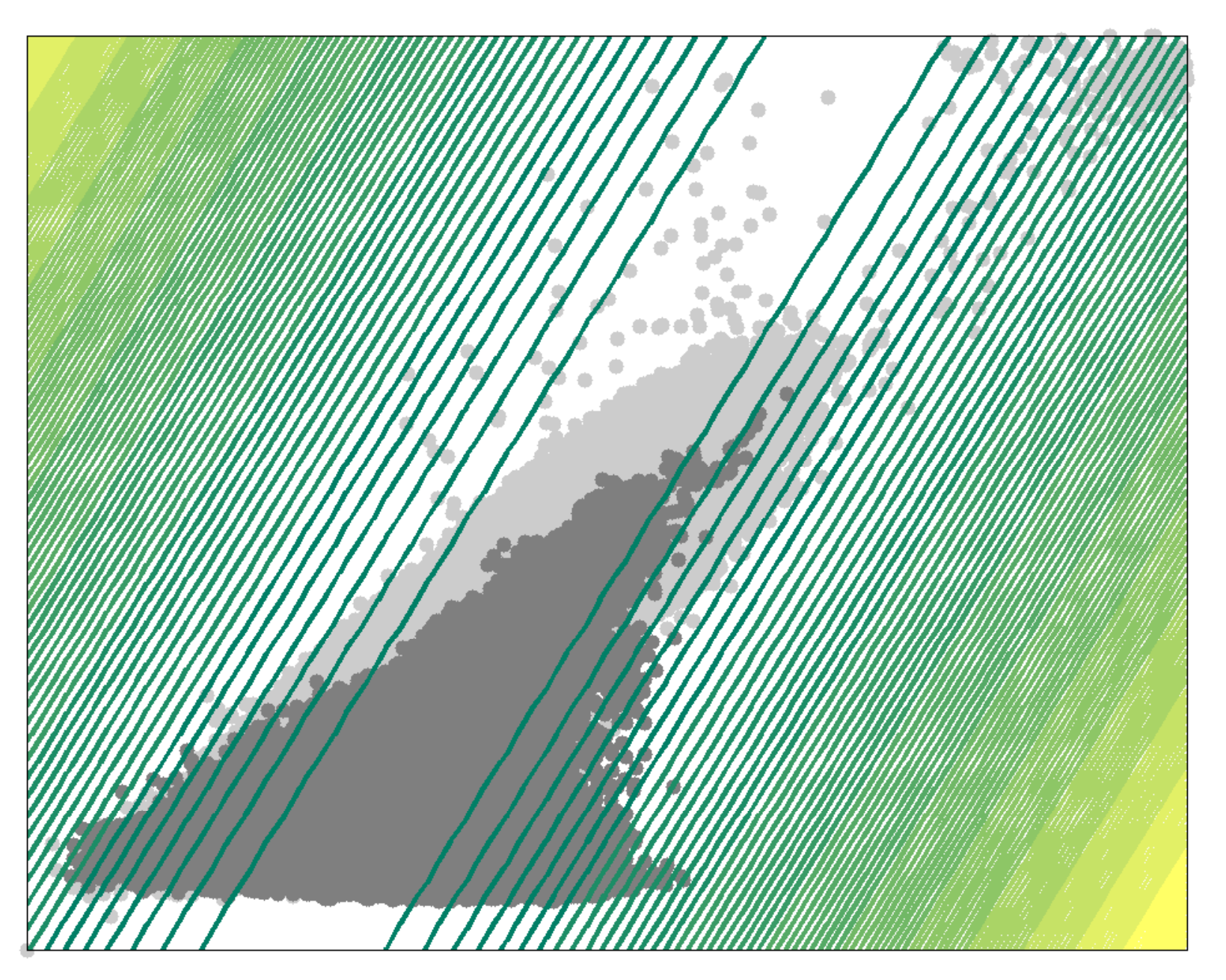} 
 & \includegraphics[width=4cm]{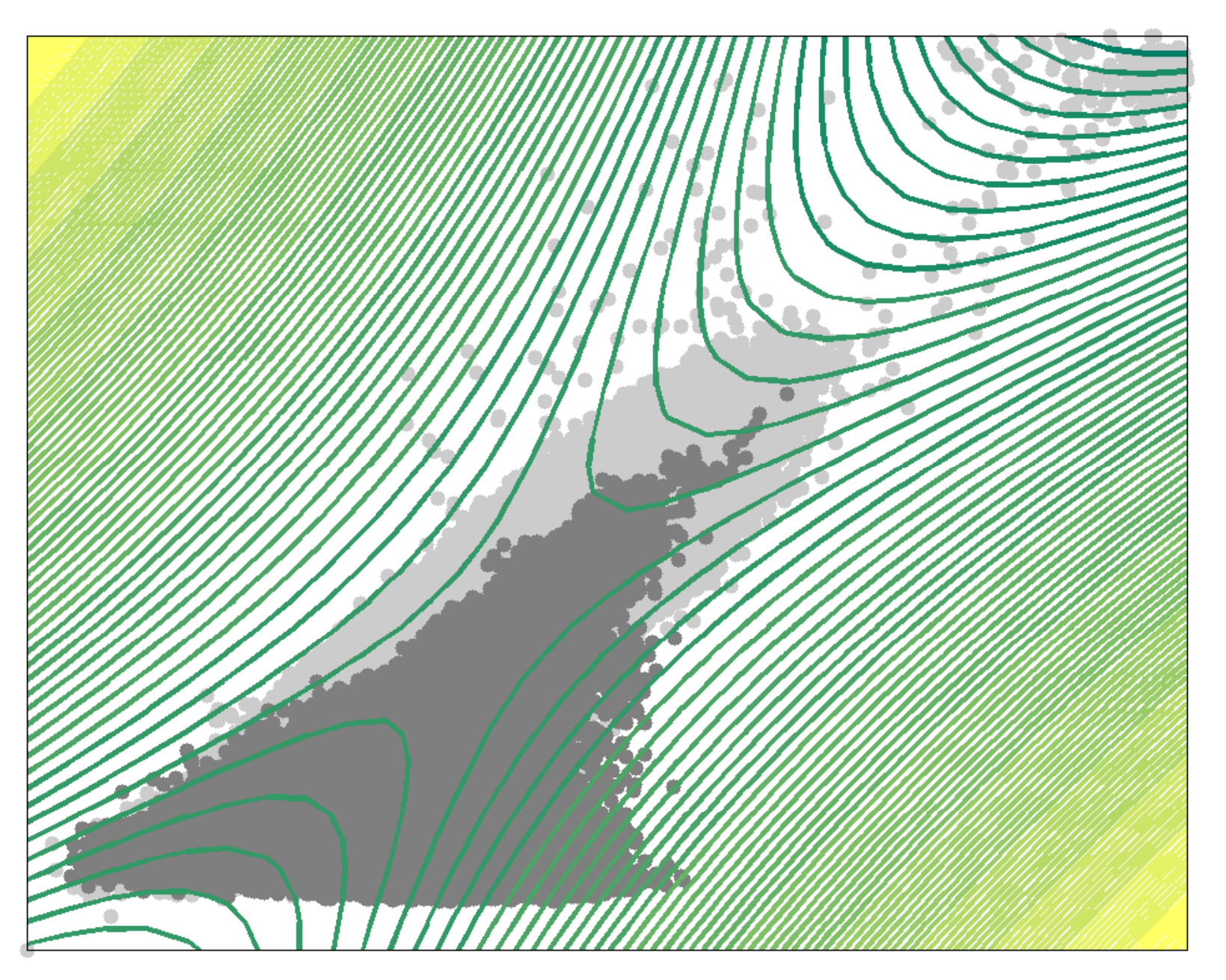} \\
EC-RX (0.92) & EC-XY (0.68) & EC-YX (0.88) & EC-HACD (0.43)\\
\includegraphics[width=4cm]{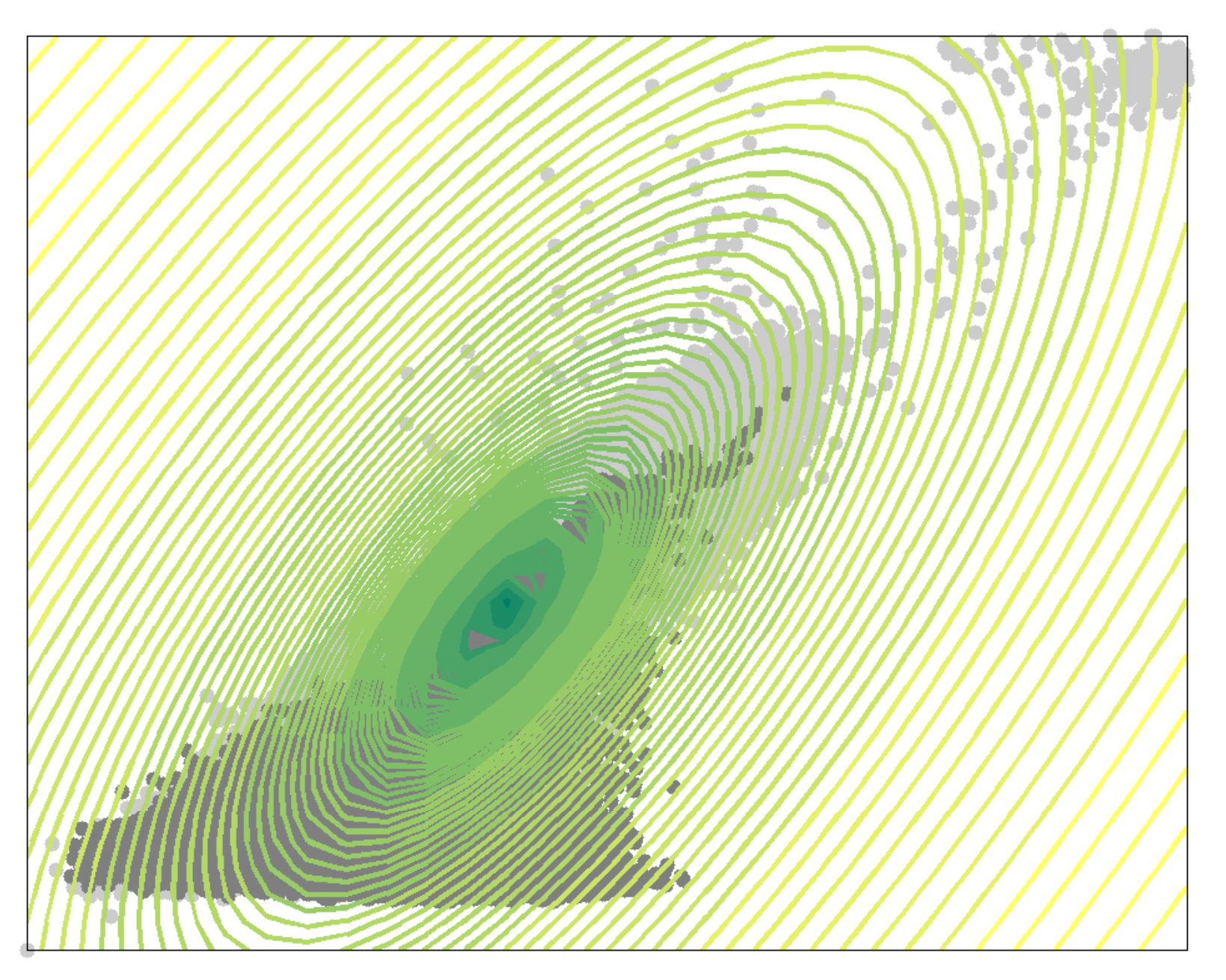} 
 & \includegraphics[width=4cm]{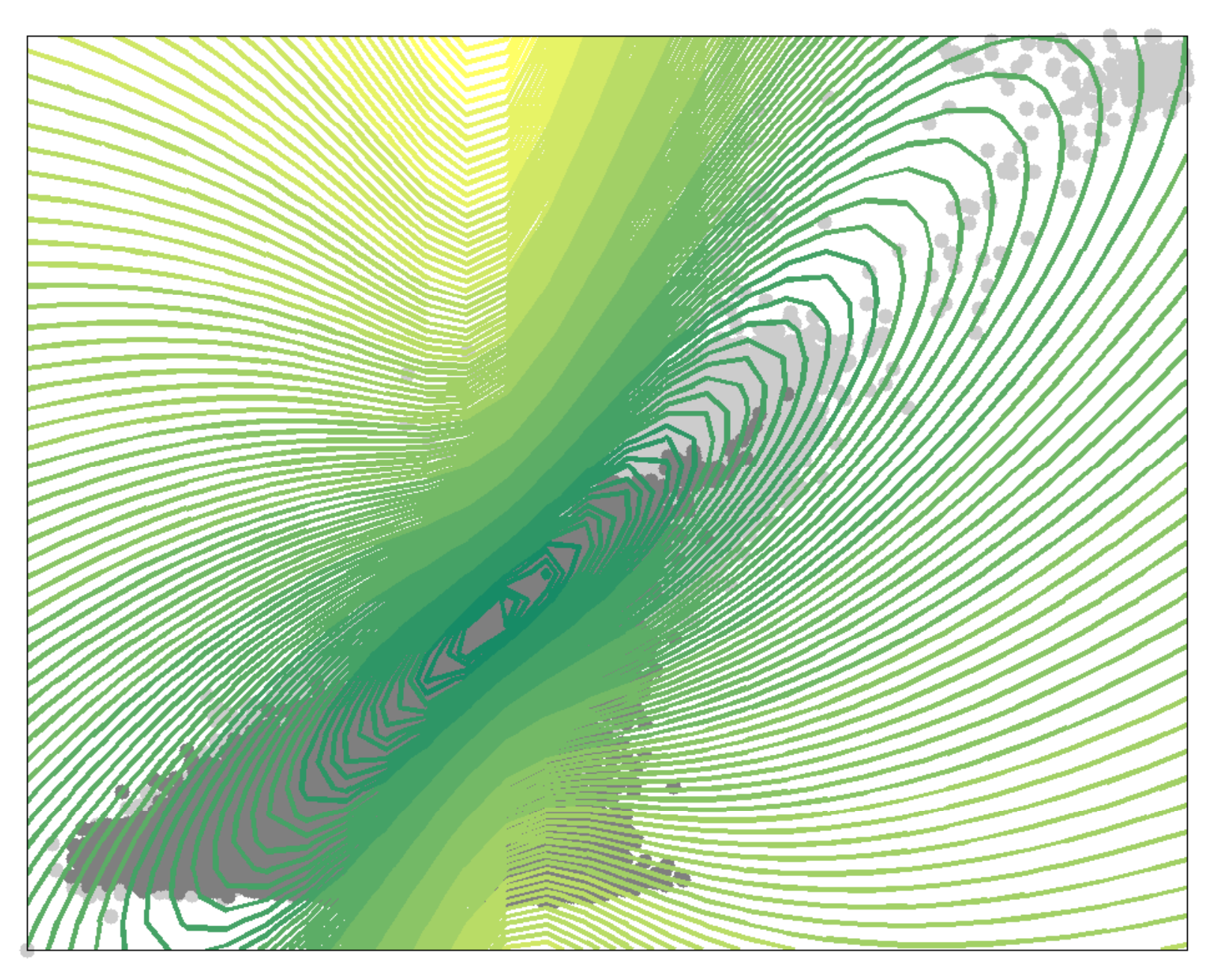} 
 & \includegraphics[width=4cm]{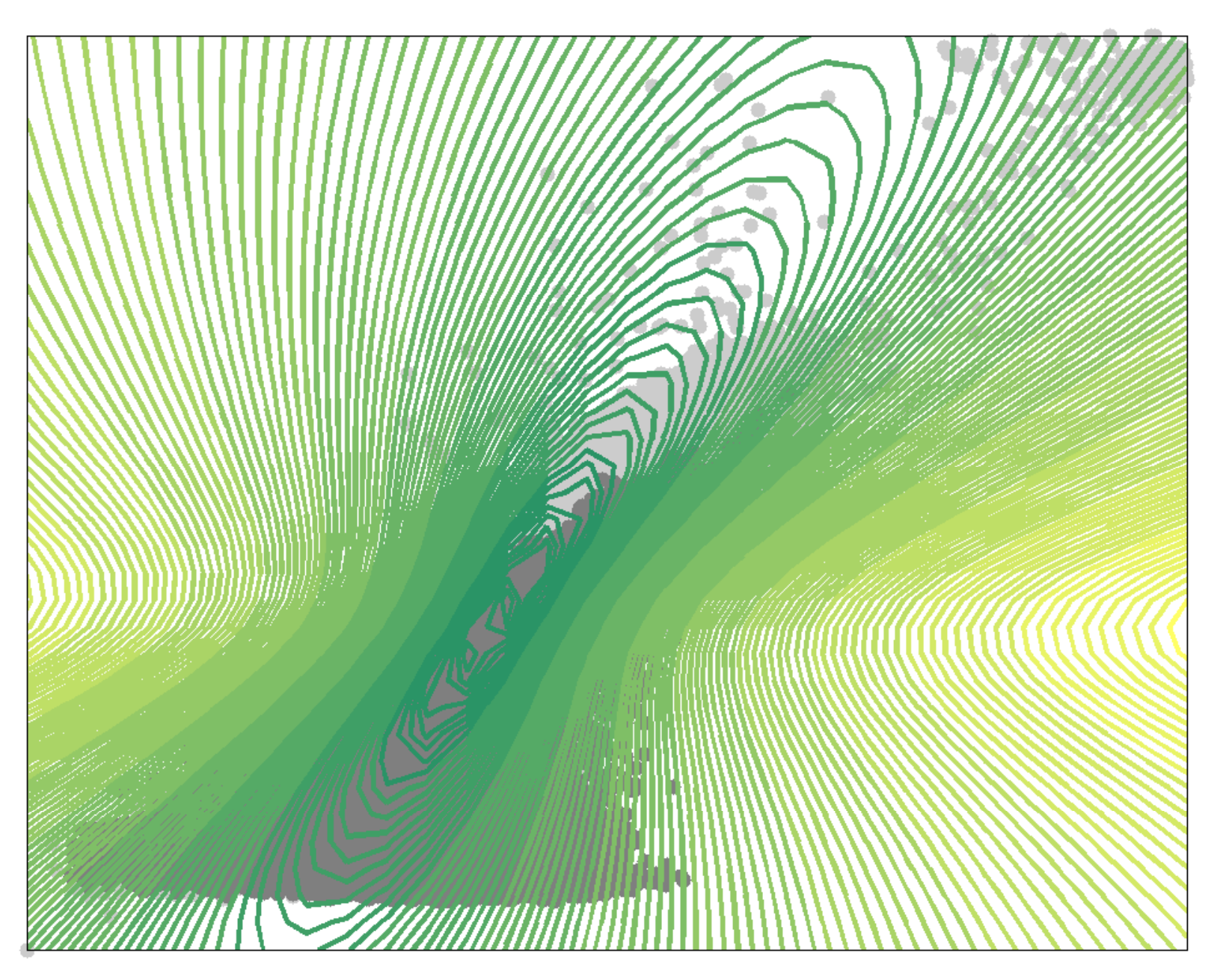} 
 & \includegraphics[width=4cm]{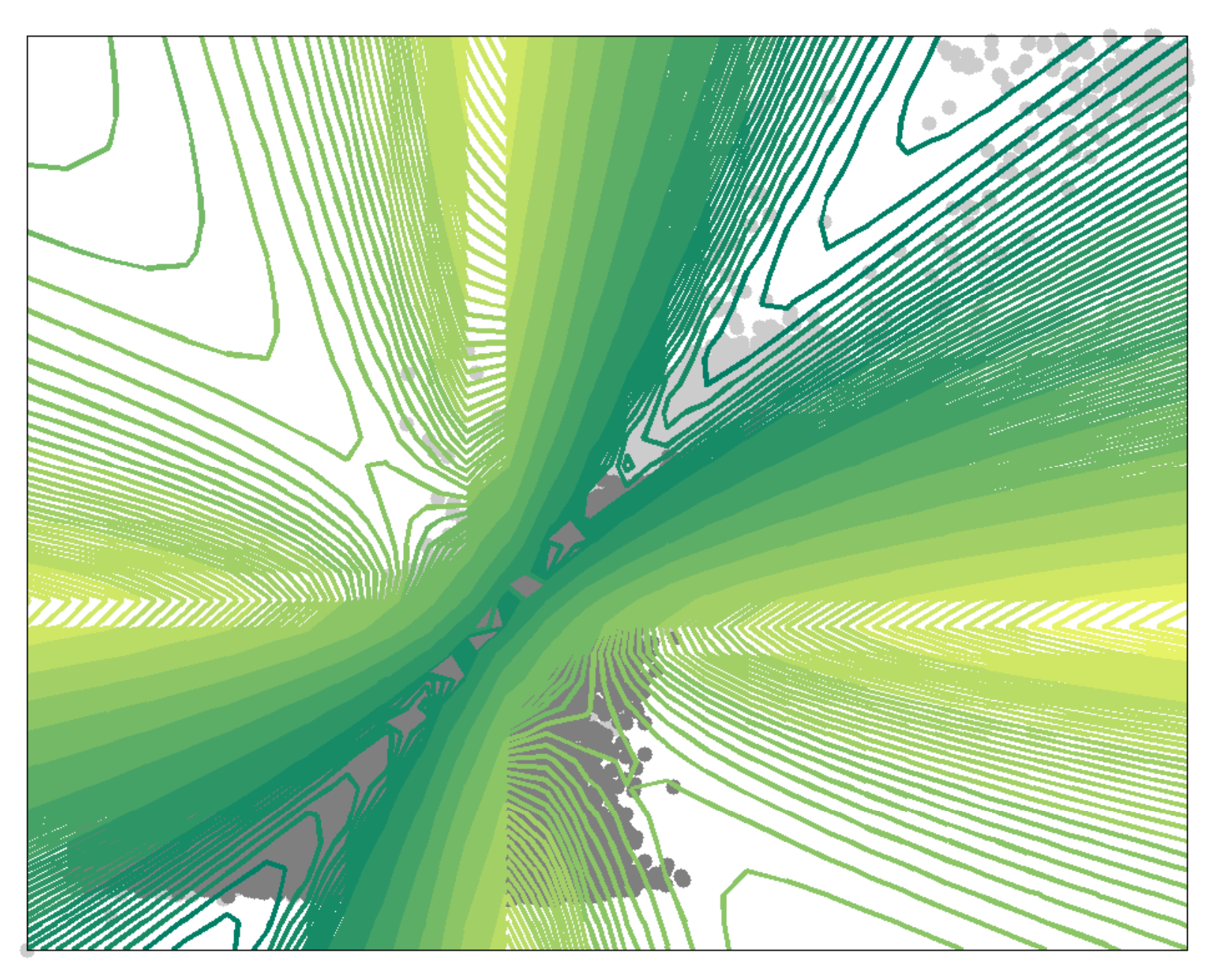}  \\
K-RX (0.95) & K-XY (0.89) & K-YX (0.92) & EC-HACD (0.95)\\
\includegraphics[width=4cm]{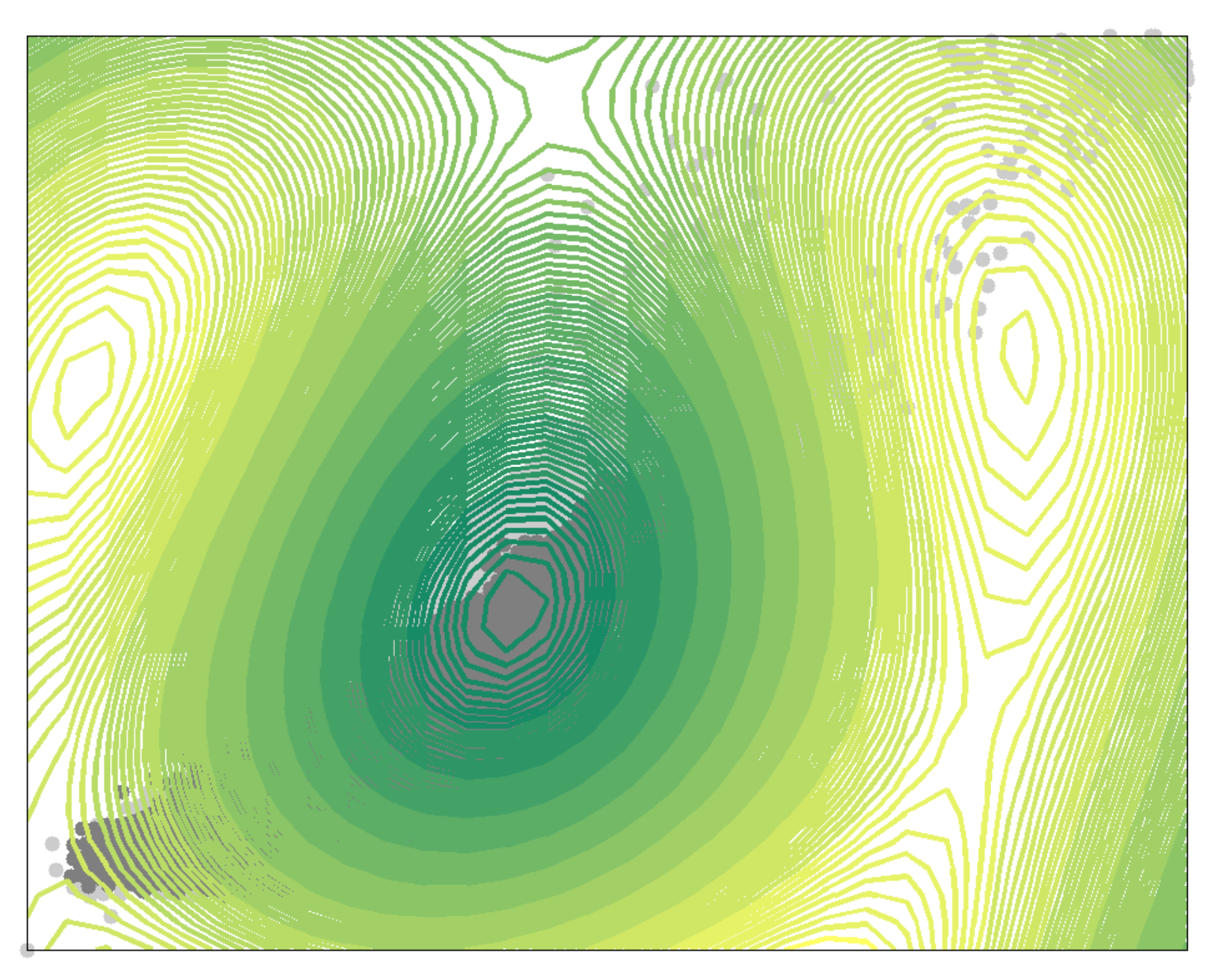} 
 & \includegraphics[width=4cm]{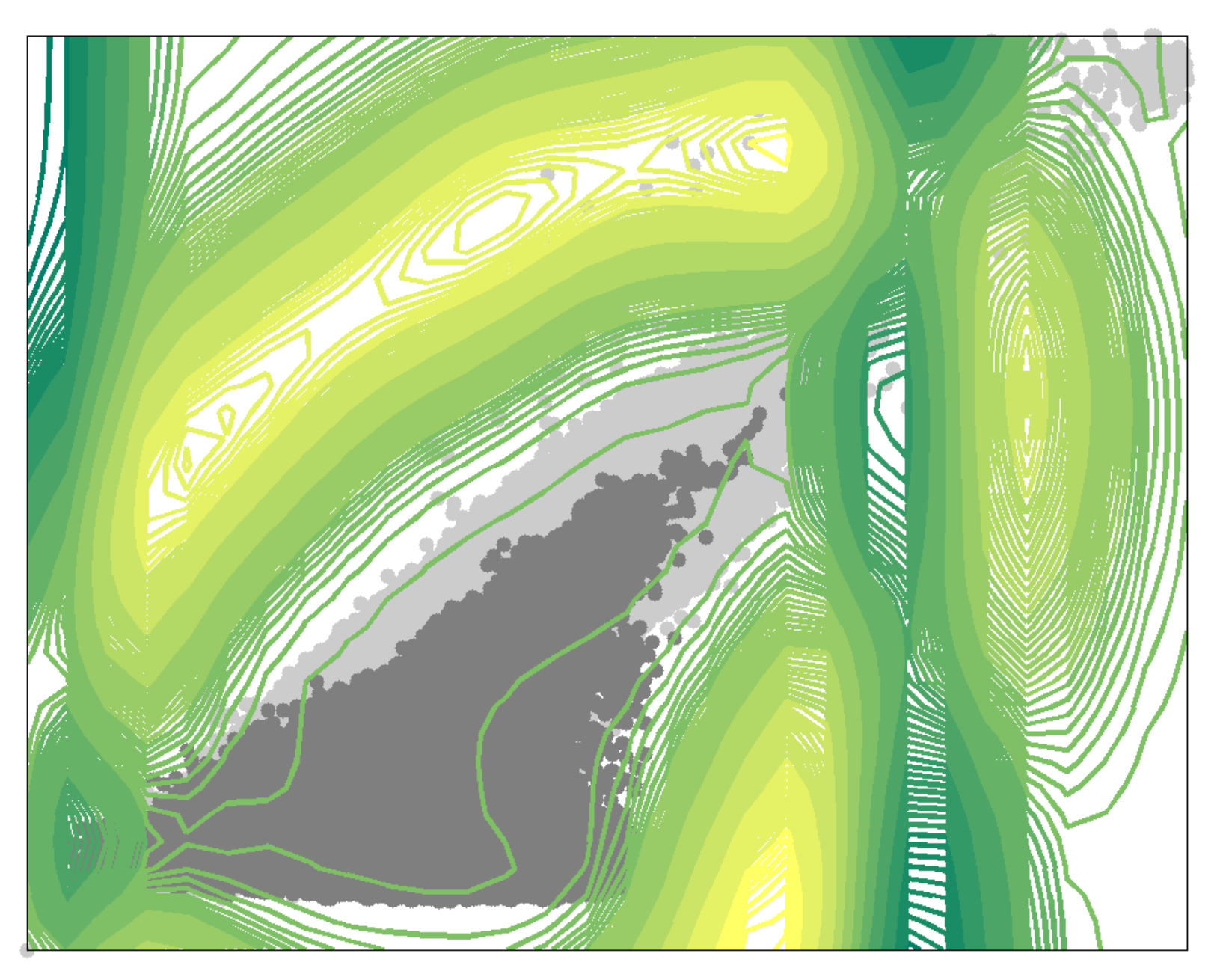} 
 & \includegraphics[width=4cm]{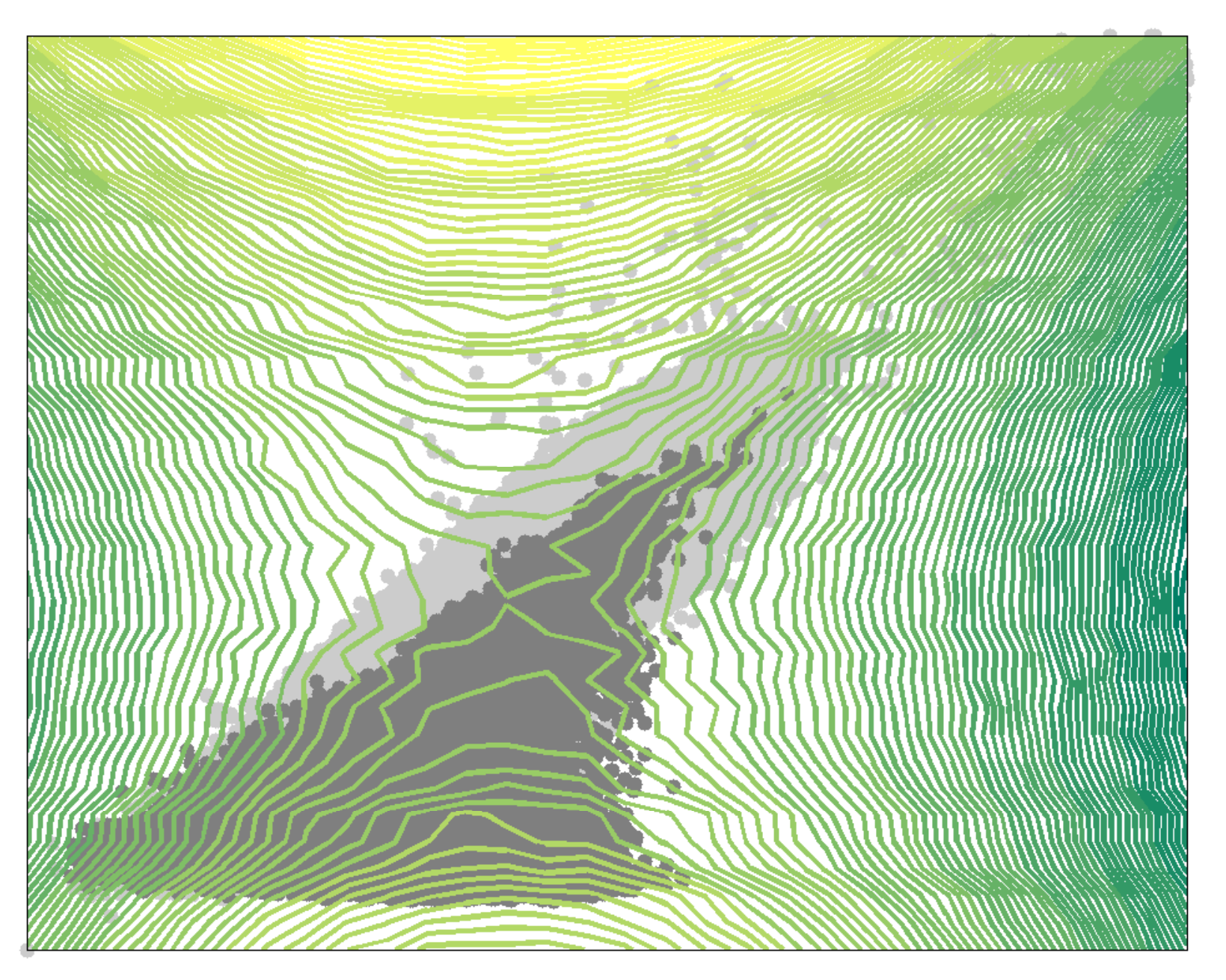} 
 & \includegraphics[width=4cm]{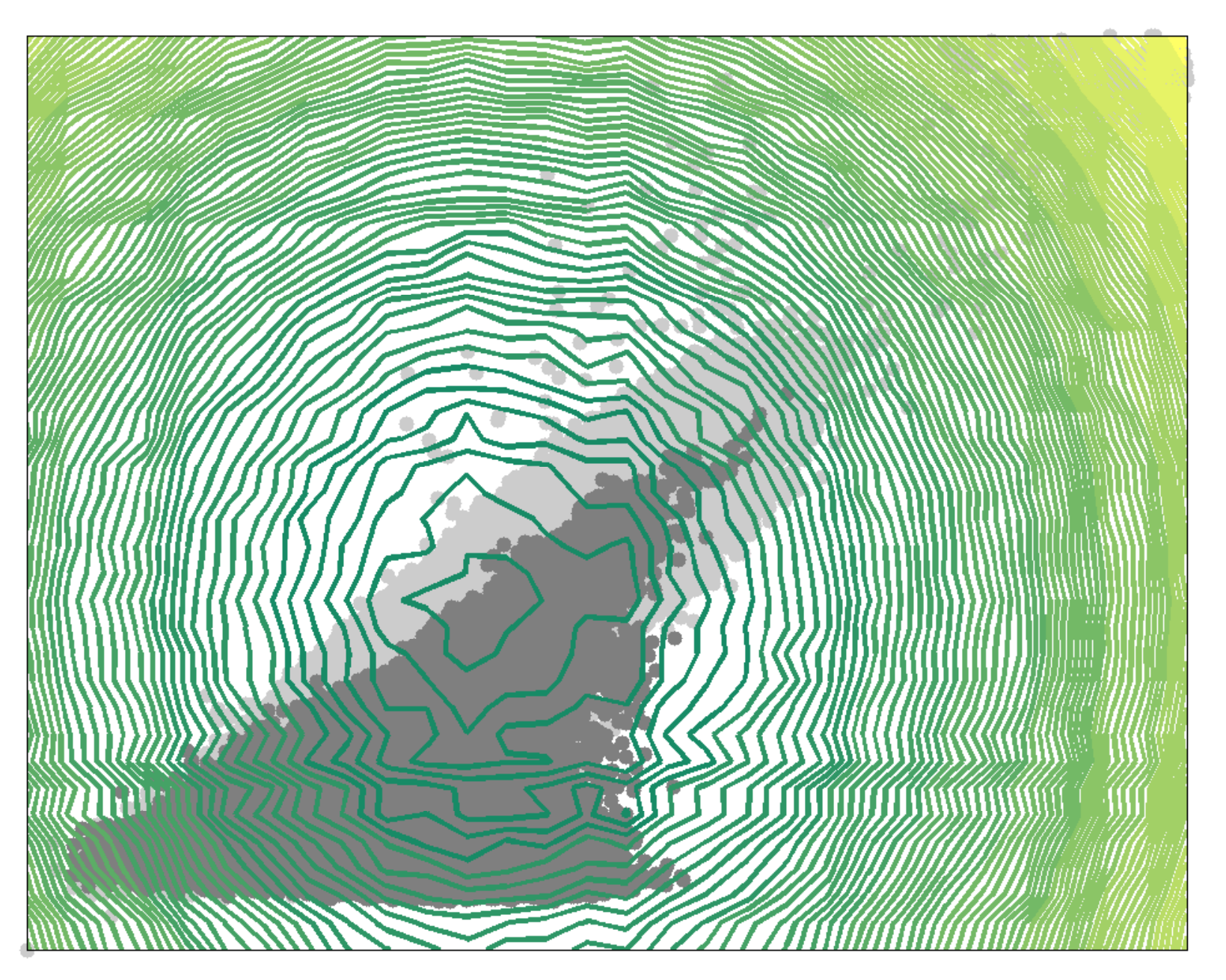} \\
K-EC-RX (0.95) & K-EC-XY (0.90) & K-EC-YX (0.97) & K-EC-HACD (0.96)\\
\includegraphics[width=4cm]{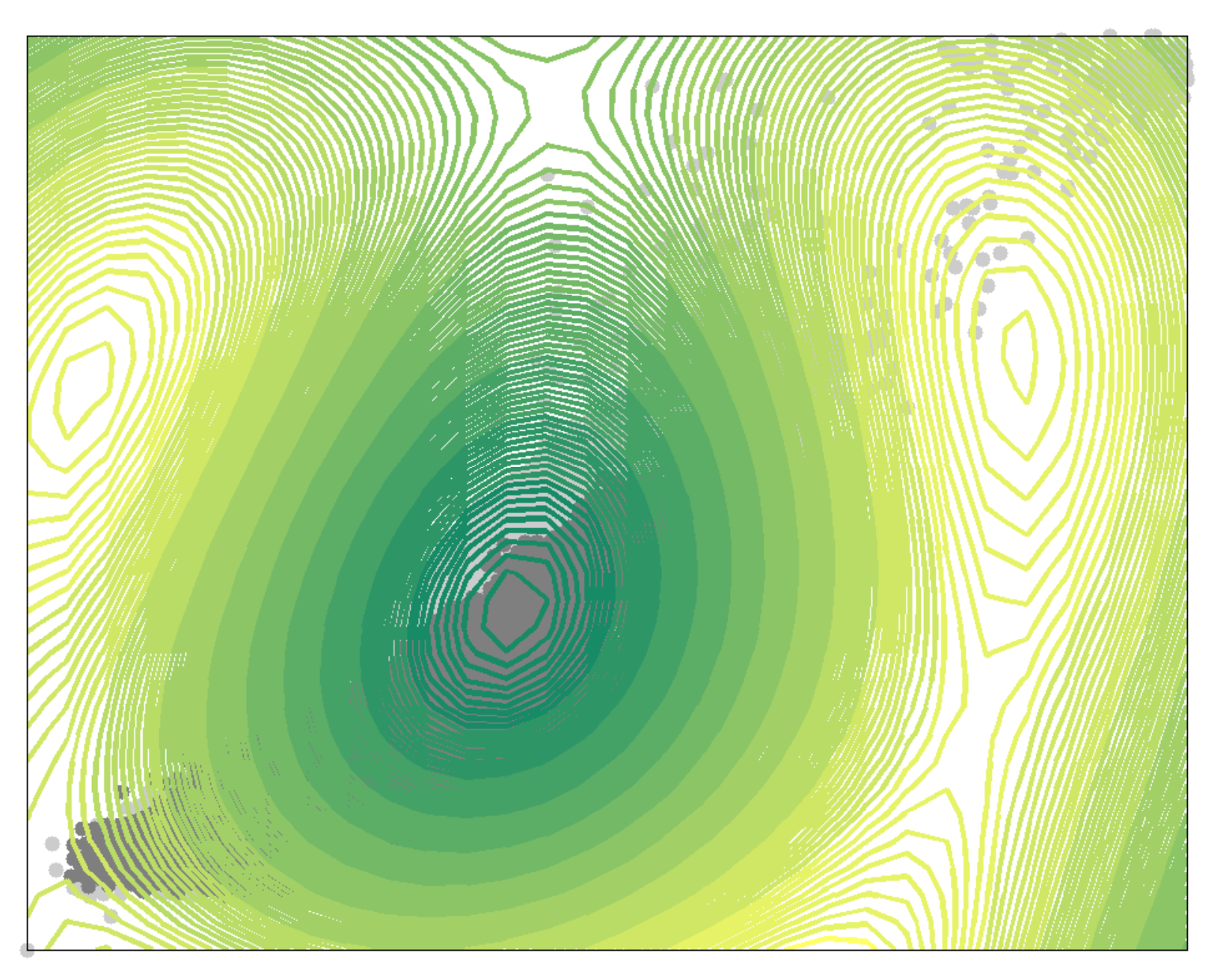} 
 & \includegraphics[width=4cm]{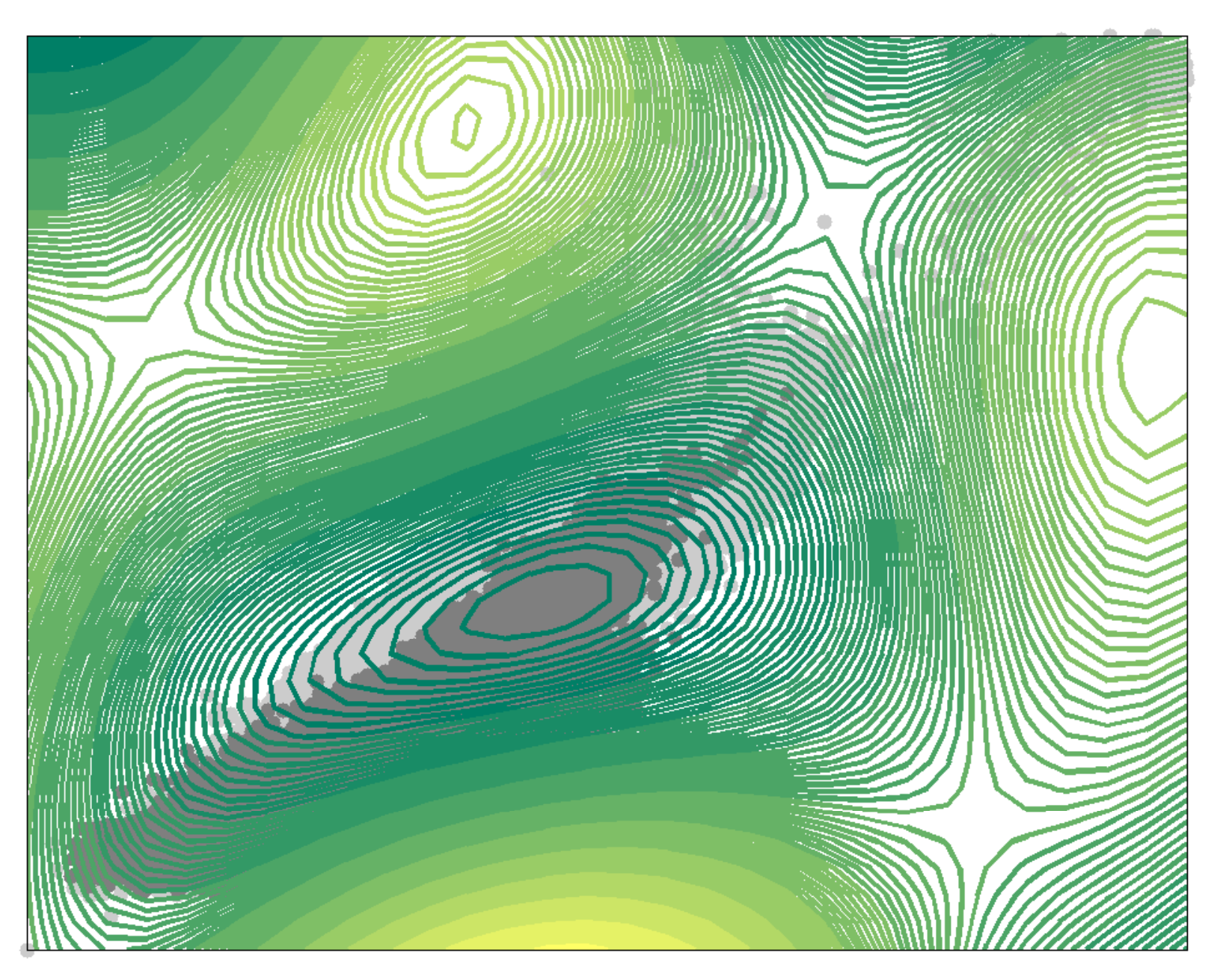} 
 & \includegraphics[width=4cm]{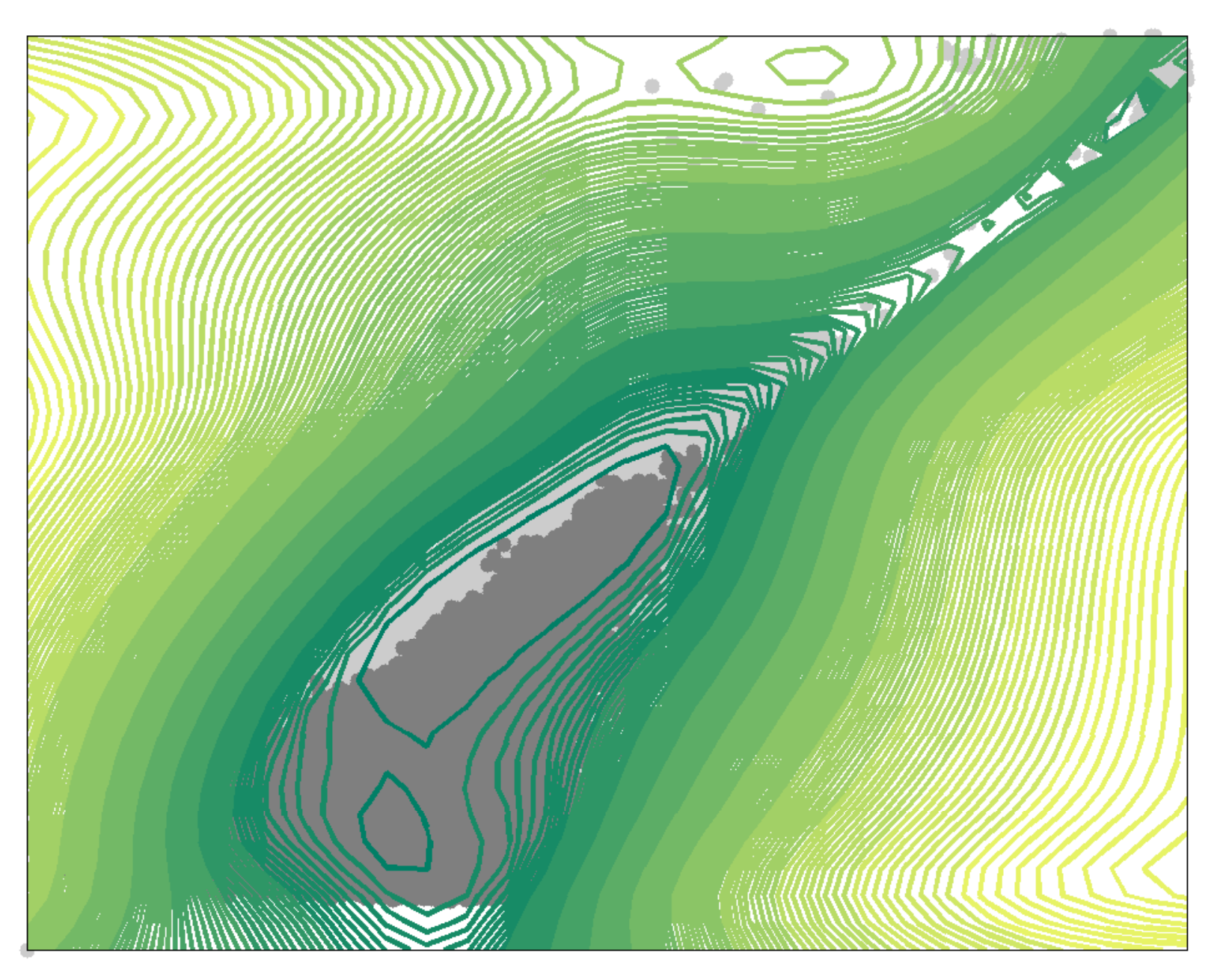} 
 & \includegraphics[width=4cm]{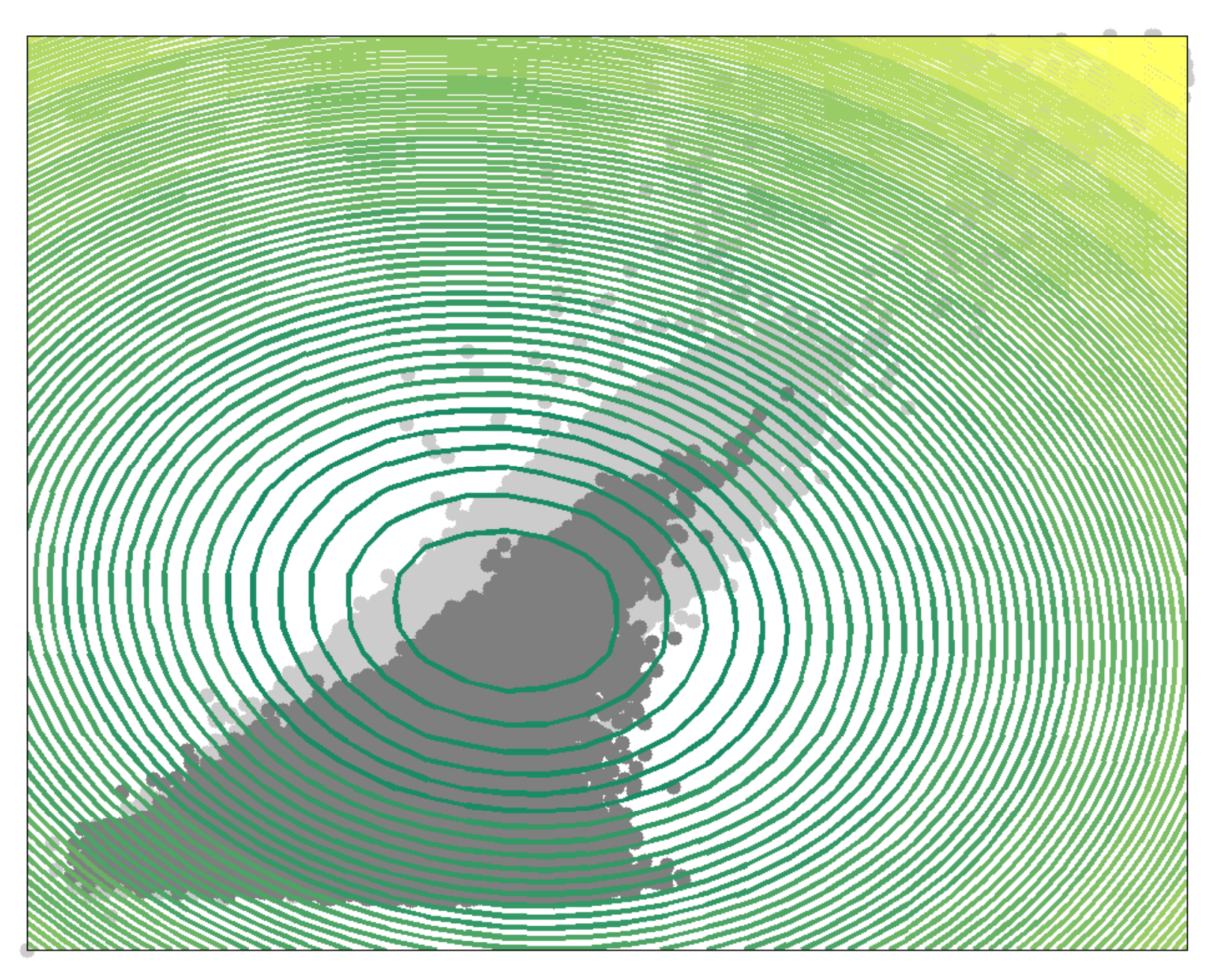} 
\end{tabular}
\end{center}
\caption{\small Toy example of anomalous detection surfaces for different methods, only the band 9 of a Sentinel-2 image was employed. Yellowish values mean more anomalous sample (i.e. bigger ${\mathcal A}$). Bright grey dots represent the non-anomalous data, while the dark gray are the anomalous data. Overall area under curve (AUC) of the receiver operating characteristic (ROC) values are given in parenthesis.}
\label{fig:toy_KACD}
\end{figure*}
     
The mapped training data matrix $\X = [\x_1, \dotsc \x_n]\in\Real^{n\times d}$ is now denoted as $\boldsymbol{\Phi}=[\bphi(\x_1), \dotsc \bphi(\x_n)]\in\Real^{n\times d_{\mathcal H}}$.
In the following we show how to estimate the $\xi(\x_i)$ function in the Hilbert space, i.e. $\xi^{\mathcal H}(\x_i) = \xi(\bphi(\x_i))$. The other terms are derived equivalently. Note that one could think of different mappings for each image, $\boldsymbol{\phi}:\x\to\boldsymbol{\phi}(\x)$ and $\boldsymbol{\psi}:\y\to\boldsymbol{\psi}(\y)$, $\boldsymbol{\Psi}\in\Real^{n\times d_{\mathcal F}}$, respectively. However, in our case, we are forced to consider mapping to the same Hilbert space because we have to stack the mapped vectors to estimate $\xi(\bphi(\z))$, i.e. ${\mathcal F}={\mathcal H}$. The mapped training data to Hilbert spaces are denoted as $\boldsymbol{\Phi}$. 
In order to estimate $\xi(\bphi(\x_i))$ we follow the same procedure as in the linear case but first mapping the points to the Hilbert space  
\begin{equation}
\xi^{\mathcal H}(\x_i) = \boldsymbol{\phi}(\x_i)(\boldsymbol{\Phi}^\top\boldsymbol{\Phi})^{-1}\boldsymbol{\phi}(\x_i)^\top.
\label{eq:Hilbert}
\end{equation}
Note that we do not have access to either the samples or the covariance in the Hilbert. 
However note that $(\boldsymbol{\Phi}^\top\boldsymbol{\Phi})^{-1} = \boldsymbol{\Phi}^\top~~(\boldsymbol{\Phi}\boldsymbol{\Phi}^\top\boldsymbol{\Phi}\boldsymbol{\Phi}^\top)^{-1}\boldsymbol{\Phi} $. This can be easily shown by right multipliying by the term $\boldsymbol{\Phi}^\top \boldsymbol{\Phi} \boldsymbol{\Phi}^\top$ and applying some linear algebra. By substituting in eq.~\ref{eq:Hilbert} we get
$$\xi^{\mathcal H}(\x_i) = 
\boldsymbol{\phi}(\x_i) \boldsymbol{\Phi}^\top~~(\boldsymbol{\Phi}\boldsymbol{\Phi}^\top\boldsymbol{\Phi}\boldsymbol{\Phi}^\top)^{-1}\boldsymbol{\Phi} \boldsymbol{\phi}(\x_i)^\top.$$ 
In this equation we can replace all dot products by reproducing kernel functions using the representer theorem~\cite{shawetaylor04}, and hence 

$$\xi^{\mathcal H}(\x_i) = \xi(\boldsymbol{\phi}(\x_i)) = {\bf k}_i(\K\K)^{-1}{\bf k}_i^\top,$$ 
where ${\bf k}_i\in\Real^{1\times n}$ contains the similarities between $\x_i$ and all training data, $\X$, and $\K\in\Real^{n\times n}$ stands for the kernel matrix containing all training data similarities. The solution may need extra regularization $\xi^{\mathcal H}(\x_i) = {\bf k}_i(\K\K+\lambda{\bf I}_n)^{-1}{\bf k}_i^\top$, $\lambda\in\Real^+$. 
Therefore the kernel version of~Eq.~\eqref{eq:ACDgauss} is:
\begin{equation*}
    {\mathcal A}_{\mathcal G}^{\mathcal H}(\x_i,\y_i)=\xi^{\mathcal H}(\z_i) - \beta_x\xi^{\mathcal H}(\x_i) - \beta_y\xi^{\mathcal H}(\y_i).
\end{equation*}
By following a similar procedure for~Eq.~\eqref{anomal}, one obtains kernel versions of the elliptically-contoured linear solution:
\begin{eqnarray*}
\begin{array}{lll}
{\mathcal A}_{\text{EC}}^{\mathcal H}(\x_i,\y_i) &=& (2d+\nu) \log\bigg(1+\dfrac{\xi^{\mathcal H}(\z_i)}{\nu}\bigg) \\
&-& \beta_x (d+\nu) \log\bigg(1+\dfrac{\xi^{\mathcal H}(\x_i)}{\nu}\bigg)\\
&-& \beta_y (d+\nu) \log\bigg(1+\dfrac{\xi^{\mathcal H}(\y_i)}{\nu}\bigg),
\end{array}
\end{eqnarray*}
Note that in the case of $\beta_x=\beta_y=0$, the algorithm reduces to kernel RX which was previously introduced in~\cite{Kwon05krx}. 


Figure \ref{fig:toy_KACD} shows an illustrative example of the surface, ${\mathcal A}$, for different methods. As in Fig.~\ref{fig:toy_ACD} the experiment is done using one band in order to be able of represent the surfaces. Different methods obtain different results, as we will see in the experiments the kernel methods obtain better results than their linear counterpart. The surfaces are direct consequence of the probabilistic model assumed, for instance in the case of RX for Gaussian and EC assumptions the surfaces are equivalent to the probabilistic distributions of $\mathbb{P}_{X,Y}$ in Fig.~\ref{fig:toy_ACD}. It is clear that the kernel versions have much more capacity to non-linearly adapt the decision surface to the problem.

\section{Experimental Results}\label{sec:experiments}

This section analyzes the proposed methods in several simulated and real examples of pervasive and anomalous changes. We evaluate the performance of the methods by using the area under the curve (AUC) of the detection receiver operating characteristic (ROC) curves. 

\begin{figure*}[t!]
\setlength{\tabcolsep}{2pt}
\begin{center}
\begin{tabular}{cccccc}
\multirow{2}{*}[1.9cm]{\includegraphics[width=5.3cm]{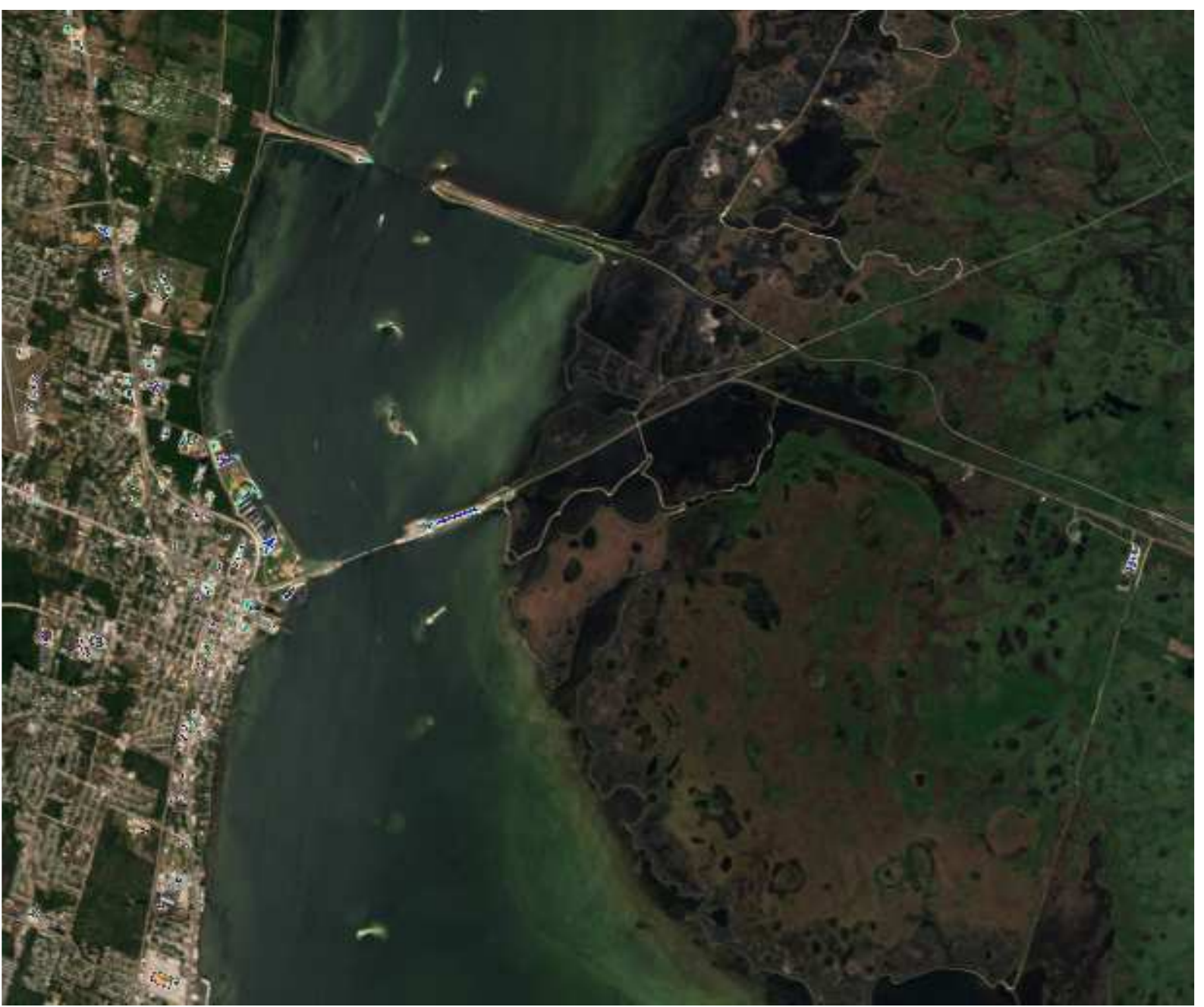}} & & \includegraphics[width=2.1cm]{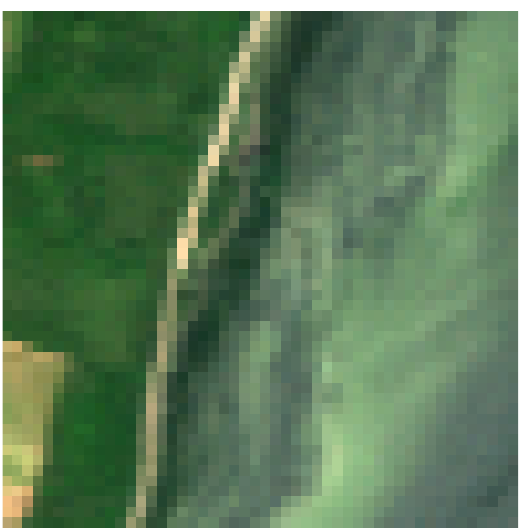} & \includegraphics[width=2.1cm]{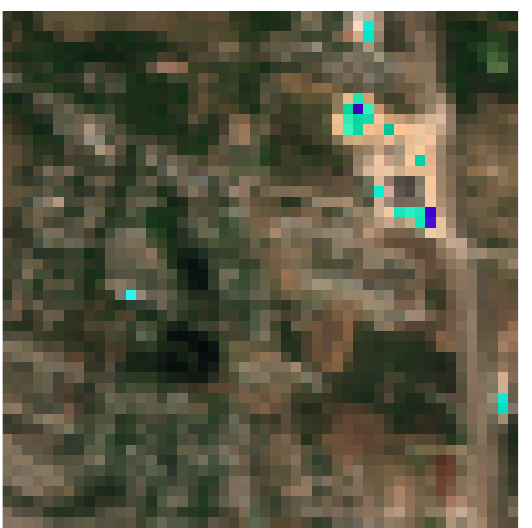} &
\includegraphics[width=2.1cm]{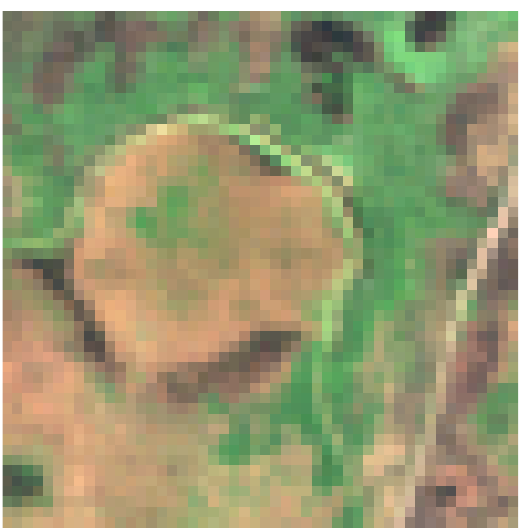} & \includegraphics[width=2.1cm]{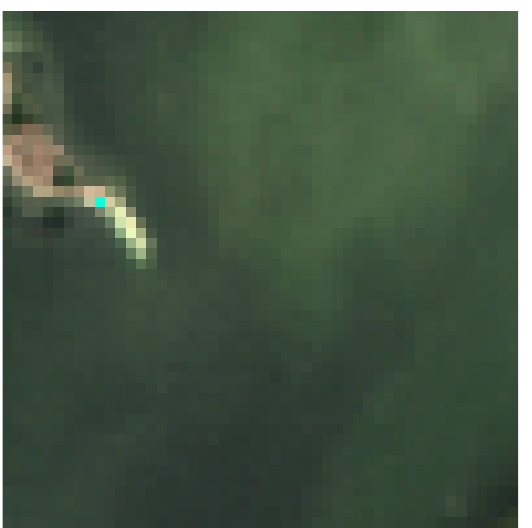} \\
 & & \includegraphics[width=2.1cm]{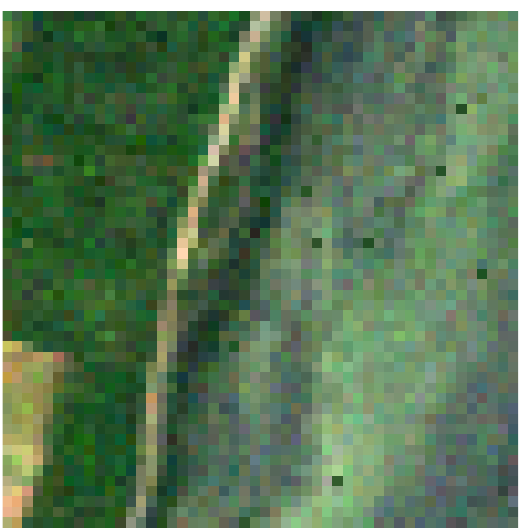} & \includegraphics[width=2.1cm]{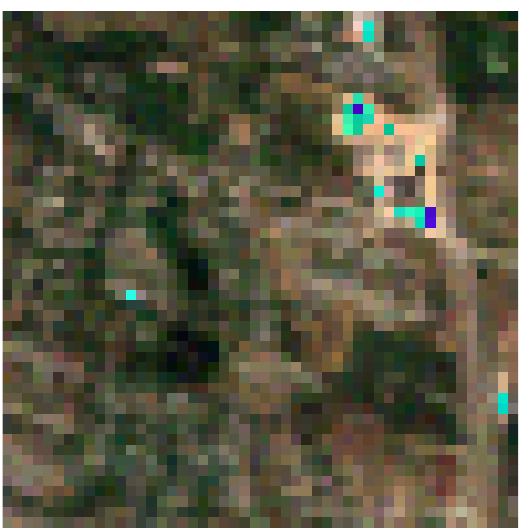} &
 \includegraphics[width=2.1cm]{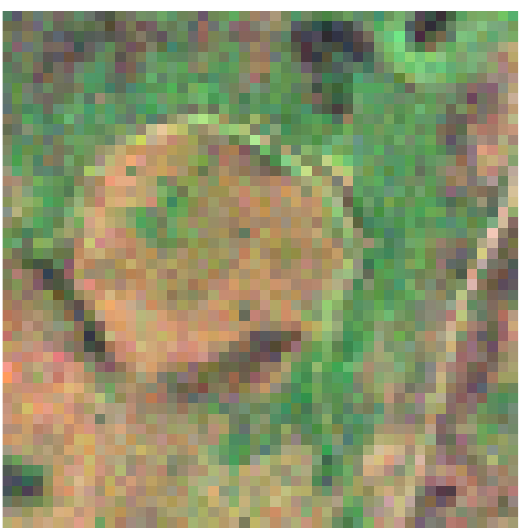} & \includegraphics[width=2.1cm]{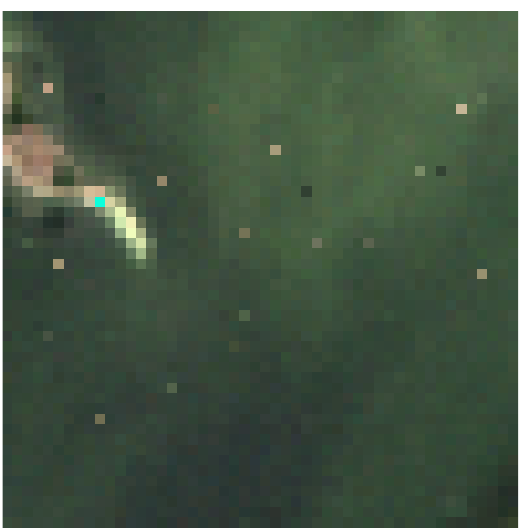} \\ 
\end{tabular}
\end{center}
\caption{\small AVIRIS hyperspectral image 
(left panel), and four illustrative chips of simulated changes (right panel). The original (leftmost) image is used to simulate an anomalous change image (rightmost) by adding Gaussian noise and randomly scrambling 1\% of the pixels. }
\label{fig:scrambleimage}
\end{figure*}

We perform three experiments with datasets with different complexity and control on the analyzed changes. First, we perform an experiment where we control the anomalous change in a synthetic controlled scenario. The second experiment deals with data where the changes were real but manually introduced directly in the images. Finally, in the third battery of experiments we deal with natural changes related to floods, droughts and man-made changes.

We provide Matlab implementations of the methods. Moreover we made available a database with the labeled  images employed in the third experiment\footnote{\url{http://isp.uv.es/kacd.html}.}.

\subsection{Experiment 1: Simulated Changes}

This experiment is devoted to analyzing the capacity of the methods to detect pervasive and anomalous changes in simulated data by reproducing the simulation framework used in~\cite{Theiler08}. The data set (see Fig.~\ref{fig:scrambleimage}) is an AVIRIS 224-channel image acquired over the Kennedy Space Center (KSC), Florida, on March 23rd, 1996. The data was acquired from an altitude of $20$ km and has a spatial resolution of $18$ m. After removing low SNR and water absorption bands, a total of $176$ bands remain for analysis. More information can be found at http://www.csr.utexas.edu/. While the linear methods suffer from the dimensionality of the data and dimensionality reduction based on Principal Components Analysis (PCA) is usually performed as preprocessing~\cite{Theiler08}, kernel methods are less sensitive to the dimensionality. While this type of preprocessing is usually accepted, it also involves the eliminatation of information. 
Here we did not further reduce the dimensionality with PCA and, instead, work directly with the SNR filtered hyperspectral data. 
{\em Pervasive changes} are simulated by adding Gaussian noise with $0$ mean and $0.1$ standard deviation to all the bands and all the pixels. The image with the added noise is taken as the second image. {\em Anomalous changes} are produced by scrambling some pixels in the second image. Note that since we are only switching the position of pixels the global distribution of the image does not change. Since the methods are applied pixel-wise, this yields anomalous changes that can not be detected as anomalies in the individual images. 

In this experiment, we restrict ourselves to the use of hyperbolic detectors (HACD), i.e. $\beta_x=\beta_y=1$, that have shown improved performance for this particular experiment~\cite{Theiler10}. We tuned all the involved parameters (estimated covariance ${\bf C}_z$ and kernel ${\bf K}_z$, $\nu$ for the EC methods, lengthscale $\sigma$ parameter for the kernel versions) through standard cross-validation in the training set and show results on the independent test set.

In this experiment we use the spectral angle mapper (SAM) kernel, $k({\bf x}_i,{\bf x}_j)=\exp(-\text{acos}({{\bf x}_i^\top{\bf x}_j}/({\|{\bf x}_i\|\|{\bf x}_j\|}))^2/(2\sigma^2))$, since it has been proven a good choice for hyperspectral images \cite{CampsValls16ksam}. Two parameters need to be tuned in our kernel versions: the regularization parameter $\lambda$ and the kernel parameter. In this case we used $\lambda=10^{-5}/n$ where $n$ is the number of training samples, and used a isotropic kernel function, whose lengthscale $\sigma$ parameter is tuned in the range of 0.05-0.95 percentiles of the distances between all training samples. We should note that, when a linear kernel is used, $k({\bf x}_i,{\bf x}_j)={\bf x}_i^\top{\bf x}_j$, the proposed algorithms reduce to the linear counterparts proposed in~\cite{Theiler10}. {The SAM kernel approximates the linear kernel for high $\sigma$ values, therefore results should be improved with regard the linear versions}. Working in the dual (or $Q$-mode) with the linear kernel instead of the original linear versions can be advantageous {\em only} in the case of higher dimensionality than available samples, $d\geq n$.

\begin{figure*}[t!]
\setlength{\tabcolsep}{-5pt}
\begin{center}
\begin{tabular}{cc}
\includegraphics[width=9cm]{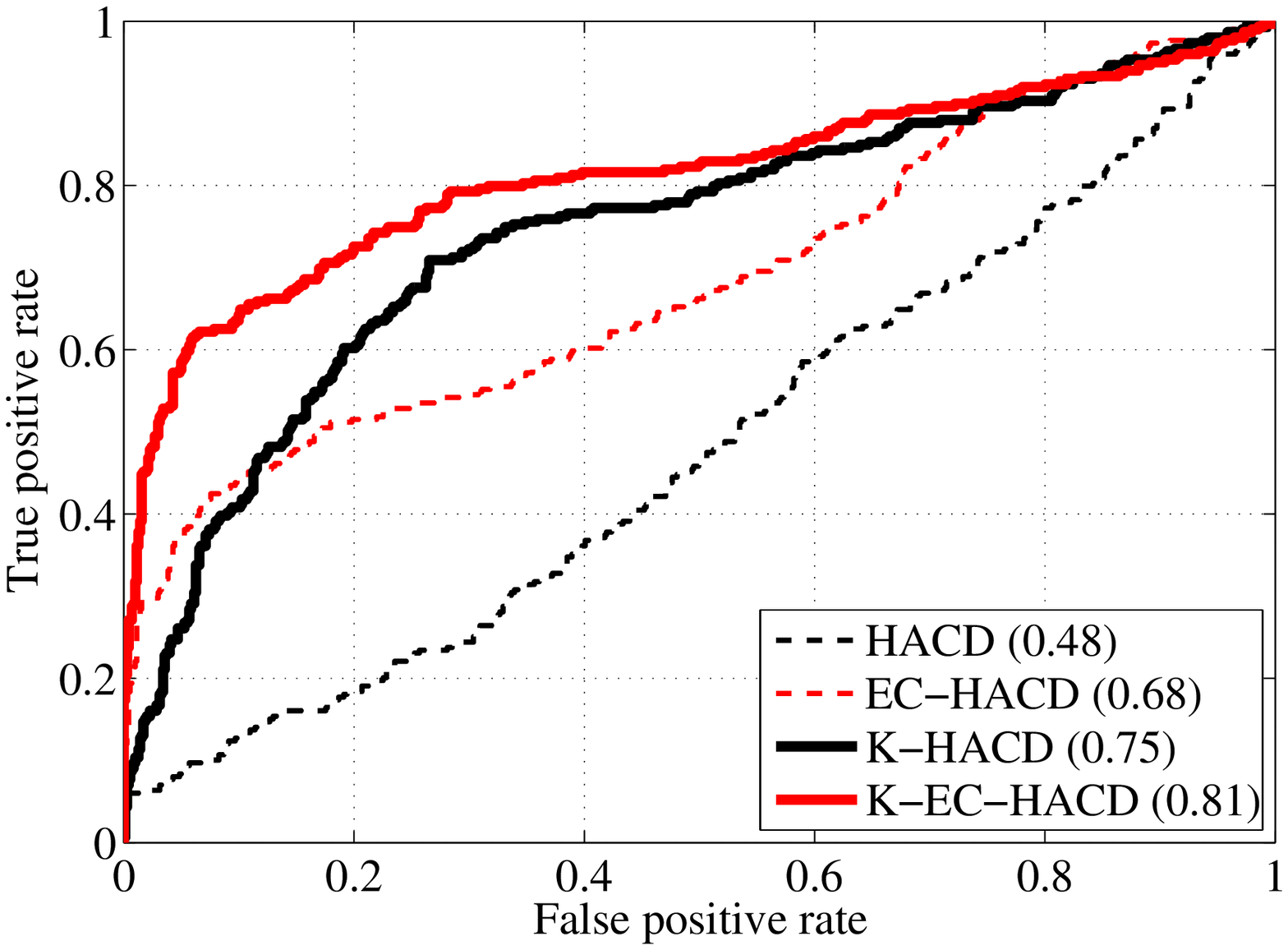} & \includegraphics[width=9cm]{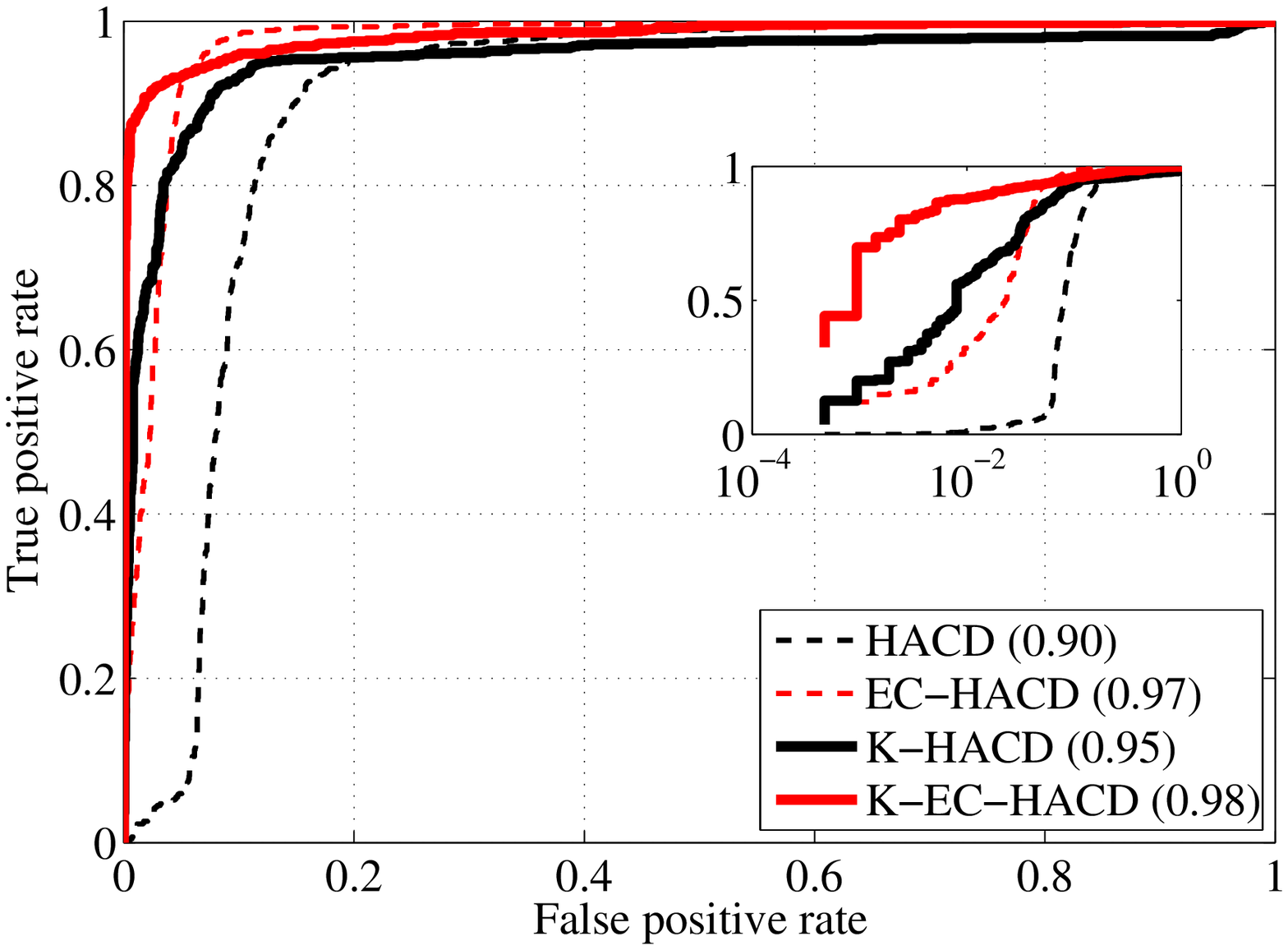} \\
\end{tabular}
\end{center}
\caption{\small ROC curves and AUC obtained for simulated changes on hyperspectral image. Results using the HACD detector in his linear (Gaussian and EC) and kernelized version are given. Left: results for 100 training examples. Right: results 500 training examples, a version of the plot in logarithmic scale is shown in the detailed plot.} 
\label{fig:kscresults}
\end{figure*}

Figure~\ref{fig:kscresults} shows the obtained ROC curves for the linear and kernel HACD methods. The dataset was split into small training sets of only $100$ and $500$ pixels, and results are given for 3000 test samples. The main conclusions are that 1) the kernel versions improve upon their linear counterparts (between 13-26\% in Gaussian and 1-5\% in EC detectors); 2) the EC variants outperform their Gaussian counterparts, especially in the low-sized training sets (+30\% over HACD and +18\% over EC-HACD in AUC terms); and 3) results improve for all methods when using $500$ training samples.  The EC-HACD is very competitive compared to the kernel versions in terms of AUC, but still the proposed K-EC-HACD leads to longer tails of false positive detection rates (right figure, inset plot in log-scale).  

\subsection{Experiment 2: Real and enforced Changes}\label{sec:exp2}
This experiment is designed to analyze the performance of the proposed methods on distortions that are present in real world imagery.  While the distortions that are present in any given pair of image sets are location and sensor dependant, some of the more prevalent distortions are due to seasonality, look-angle, and spatial resolution.  These experiments employee a very-high spatial resolution sensor that was used to image the same target with highly varying view angles (thus, varying distortion and layover) as well as large differences in seasonality.  The ability to detect anomalous changes in these highly distorted image sets illustrates the unique advantage of these types of algorithms and, in particular, the performance advantages of the proposed methods.  

\begin{figure}[h!]
\centering
\subfloat[May. 2013, 14.0\degman off-nadir]{\includegraphics[width= 2.55cm,height = 4cm]{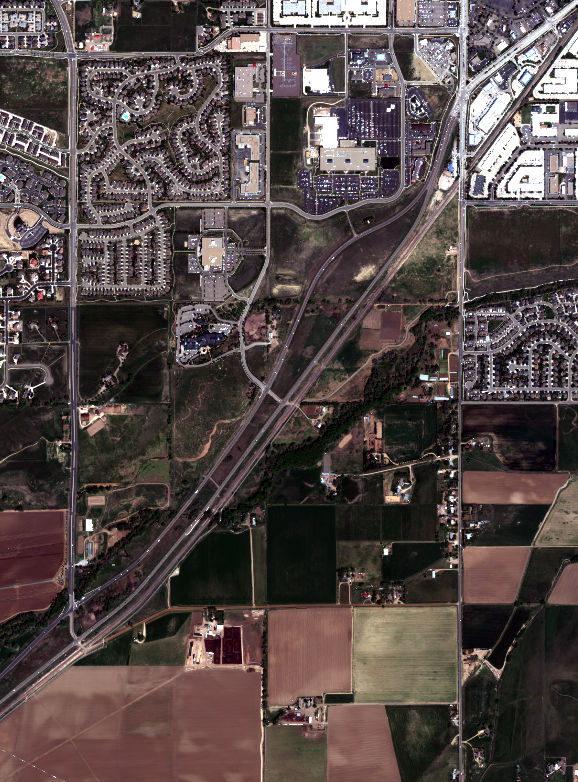}} \hspace{1mm}
\subfloat[Aug. 2013, 43.6\degman off-nadir]{\includegraphics[width= 2.55cm,height = 4cm]{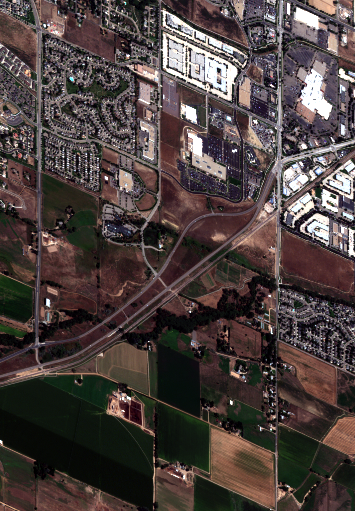}} \hspace{1mm}
\subfloat[Nov. 2013, 29.3\degman off-nadir]{\includegraphics[width= 2.55cm,height = 4cm]{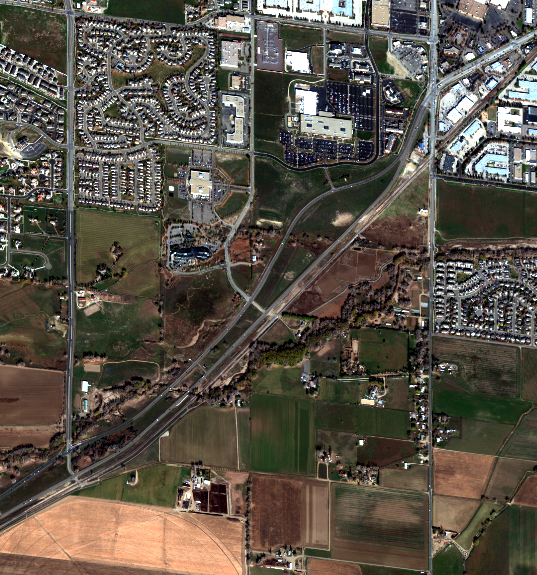}} \hspace{1mm}
\caption{\small The three WorldView-2 images present a wide variety of distortions due to both seasonality and view angle.  In addition to the more obvious changes in agricultural and natural vegetation, the varying view-angles result in variations in ground sample distances (GSD) of 2.0 m (May), 3.6 m (Aug), and 2.4 m (Nov).}
\label{fig:wv2Imgs}
\end{figure}

The experiments utilize three WorldView-2 images collected in May, August, and November of 2013.  All three images (Fig.~\ref{fig:wv2Imgs}) were collected over a mixed suburban and rural area with urban residential features, roadways, rivers, and agricultural fields.  The first image (May) was acquired at a relatively small off-nadir (14.0\degman) angle early in the summer season.  The second (Aug) and third (Nov) images were collected at much higher off nadir angles, 43.6\degman and 29.3\degman, respectively.  In each of the final two images, one dark and one white tarp (20$\times$20 m each) were introduced as anomalous changes. 

This creates two anomalous change image sets on which to test the proposed methods with varying degree of both angular and seasonality distortions: (1) May/Aug: High off-nadir difference, moderate seasonality change; and (2) May/Nov: Moderate off-nadir difference, large seasonality change.  While the white and black tarps that are introduced into the change images are highly anomalous, the spectral change is not unrepresentative of real-world problems.  Additionally, the ability to more accurately model changes in highly distorted images provides a unique test case for these proposed methods. 

\begin{figure}[t!]
\centering
\subfloat[May/Aug High Off-Nadir Experiment]{\includegraphics[width=3.0in.]{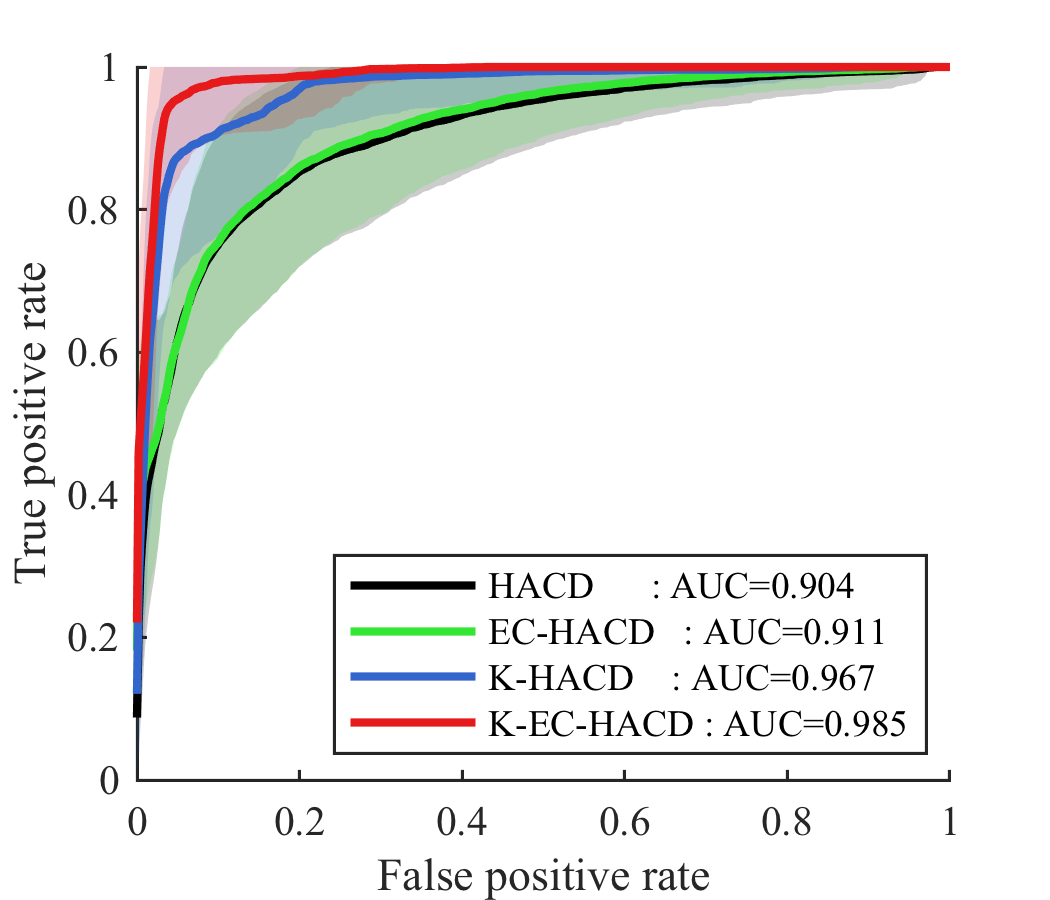}}\\
\subfloat[May/Nov Large Seasonality Experiment]{\includegraphics[width=3.0in.]{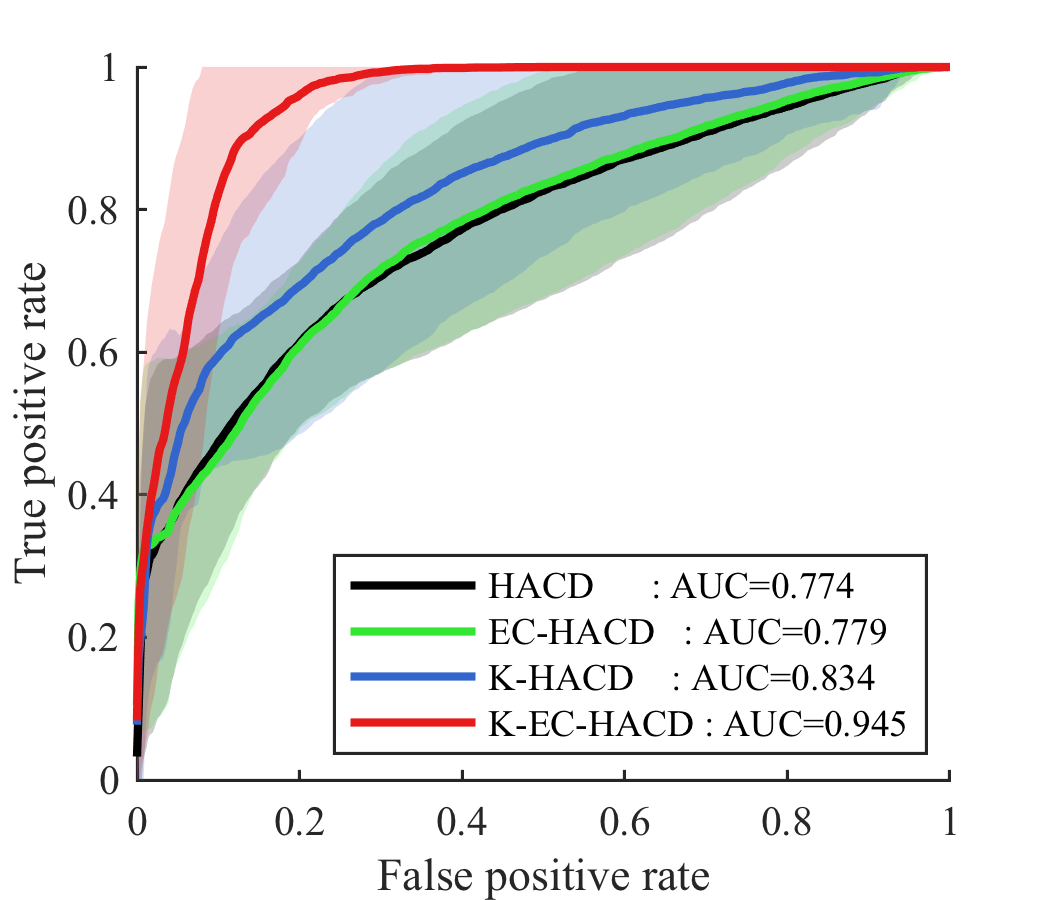}}
\caption{\small ROC curves for the two experiments of Section~\ref{sec:exp2}.  The mean value of the experimental runs is plotted with the standard deviation of each detection algorithm represented by the shaded region.} 
\label{fig:wv2ROCs}
\end{figure}

For each experiment, $50$ non-anomalous pixels were randomly selected from the stacked image sets to model the data space using the proposed algorithms.  $500$ randomly selected (training samples held out) non-anomalous and all anomalous pixels (May/Aug:$153$, May/Nov:$144$) were select for testing. These random selections were collected for $50$ independent runs. 
The mean ROC curves are reported in Fig.~\ref{fig:wv2ROCs} and the statistics for AUC are reported in Table~\ref{table:wv2AUCs}.  As was reported earlier, the parameters $\nu$ and $\sigma$ were tuned through standard cross-validation. The results are shown for independent test sets. 
In both of the experiments, the HACD and EC-HACD methods had almost identical average ROC curves. The parameter search for $\nu$ used in the EC-HACD method favored very large values, indicating that the data space is Gaussian and does not particularly benefit from elliptical modeling. This is most likely due to the anomalousness of the tested anomalous targets.  Each of the tarp spectral signatures are highly anomalous (very dark and very bright) presenting a relatively simplified modeling space. However, the kernel methods did outperform the non-kernel methods by a statistically significant +8\% and +17\% as measured by mean AUC.

\begin{table}[t!]
\hspace{-1cm}
\centering
\caption{Area Under the Curve Statistics for the WorldView-2 View-Angle and Seasonality Experiments.}
\label{table:wv2AUCs}
\scriptsize
\begin{tabular}{ |l|l|l| } 
\hline
\hline
\rowcolor[HTML]{A9A9A9} \bf{METHODS}  & \bf{May-Aug Large Off-Nadir} & \bf{May-Nov Large Seasonality}\\ \hline\hline
\rowcolor[HTML]{D8D8D8}  \multicolumn{3}{|c|}{Longmont, Colorado}      \\
\hline\hline
HACD & \textit{0.90 $\pm 0.06$} & \textit{0.77 $\pm 0.08$}  \\ \hline
EC-HACD & \textit{0.91 $\pm 0.06$} & \textit{0.78 $\pm 0.08$}  \\ \hline
K-HACD & \textit{0.97 $\pm 0.04$} & \textit{0.83 $\pm 0.11$}  \\ \hline
K-EC-HACD & \textit{0.99 $\pm 0.02$} & \textit{0.95 $\pm 0.04$}  \\ \hline

\end{tabular}
\end{table}

\subsection{Experiment 3: Real and Natural Changes}

This experiment deals with the detection of anomalous changes that can be found naturally in a real environment.

\subsubsection{Data collection}

\begin{table}[t!]
\centering
\caption{Images attributes in the experimentation dataset. }\label{table:database}
\label{ROC_table}
\begin{tabular}{ |l|l|c|l|l|} 
\hline\hline
\rowcolor[HTML]{A9A9A9} \bf{Images} & \bf{Sensor}&\bf{Size} & \bf{Bands}  & \bf{SR}\\ \hline \hline
\rowcolor[HTML]{D8D8D8} \multicolumn{5}{|l|}{\bf Experiment 1}
\\ \hline \hline

KSC                 & AVIRIS    & 512 x 614      &  224  & 18m\\ 
\hline \hline

\rowcolor[HTML]{D8D8D8}\multicolumn{5}{|l|} {\bf Experiment 2}  \\ \hline \hline
Longmont (May) & Worldview-2  &  1156 x 1563  & 8    & 2.0m
\\\hline
Longmont (Aug) & Worldview-2  &  710 x 1021  & 8    & 3.6m
\\\hline
Longmont (Nov) & Worldview-2  &  1074 x 1149  & 8    & 2.4m
\\\hline\hline

\rowcolor[HTML]{D8D8D8}\multicolumn{5}{|l|}{\bf Experiment 3}\\ \hline \hline                   

Argentina           & Sentinel-2    & 1257 x 964      &  12  & 10m-60m 
\\\hline
Australia           & Sentinel-2    & 1175 x 2031     &  12 & 10m-60m
\\\hline
California          & Sentinel-2    &  332 x 964     &   12   & 10m-60m          
\\\hline
Poopo Lake               & MODIS\footnotemark       & 326 x 201      & 7  & 250m-1km
\\\hline            
  Denver              & QuickBird    &  500 x 684      &   4 & 1m-4m\\
\hline

\end{tabular}
\end{table}

\footnotetext{Only bands in the visible part of the spectrum were used.}

\begin{figure*}[b]
\centering 

\subfloat[Jul 2016]{\includegraphics[width= 5cm]{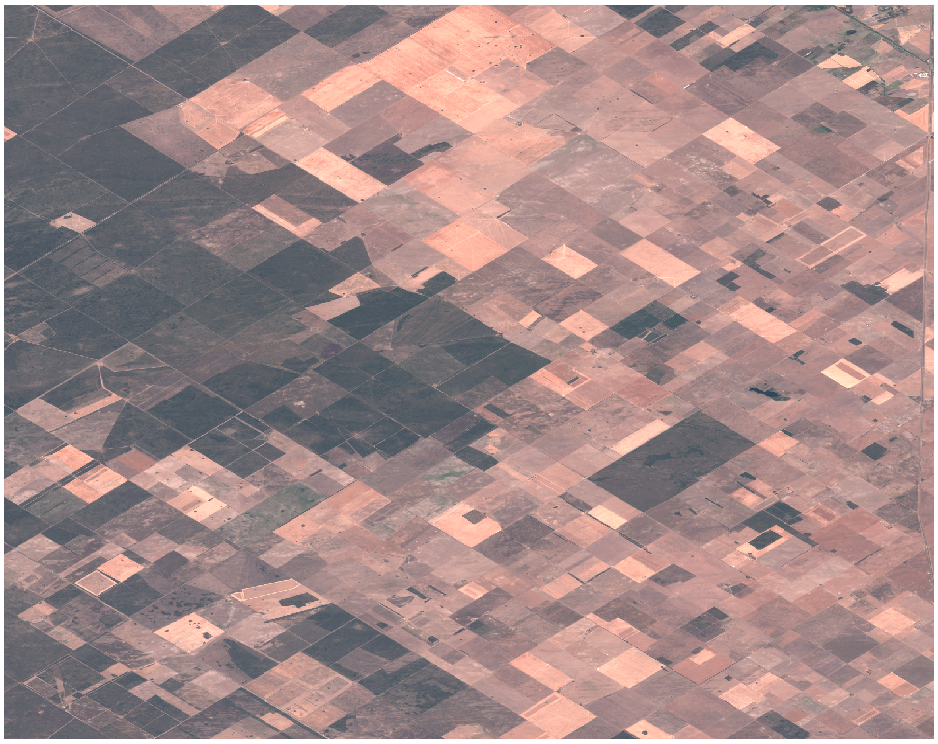}} \hspace{1mm}
\subfloat[Aug 2016]{\includegraphics[width= 5cm]{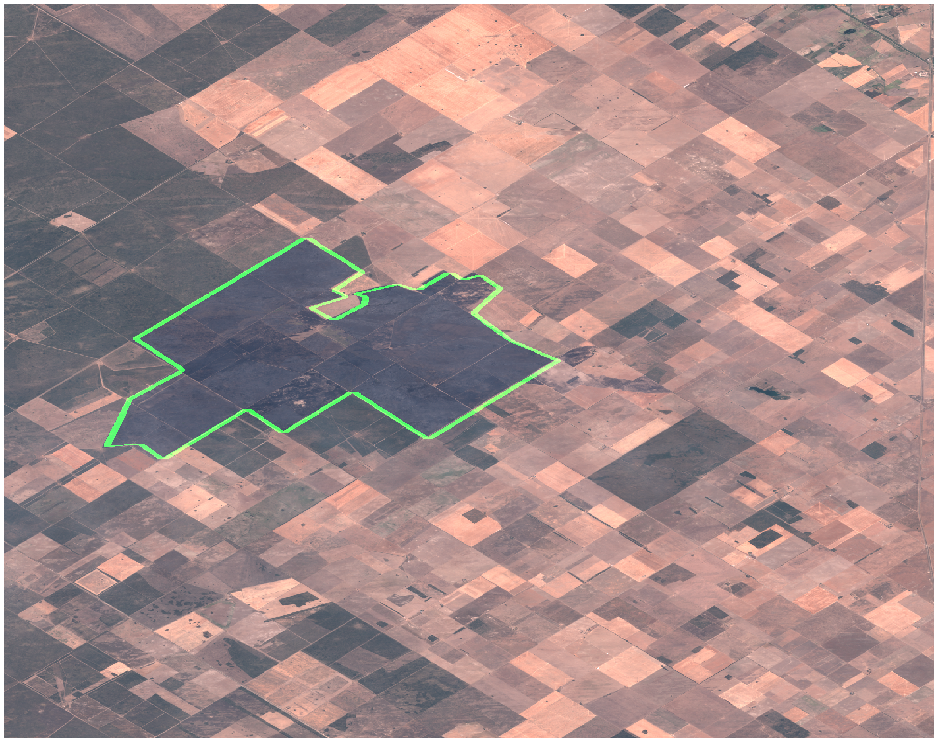}} \hspace{1mm}

\vspace{-0.3cm}
\subfloat[Mar 2017]{\includegraphics[width= 5cm]{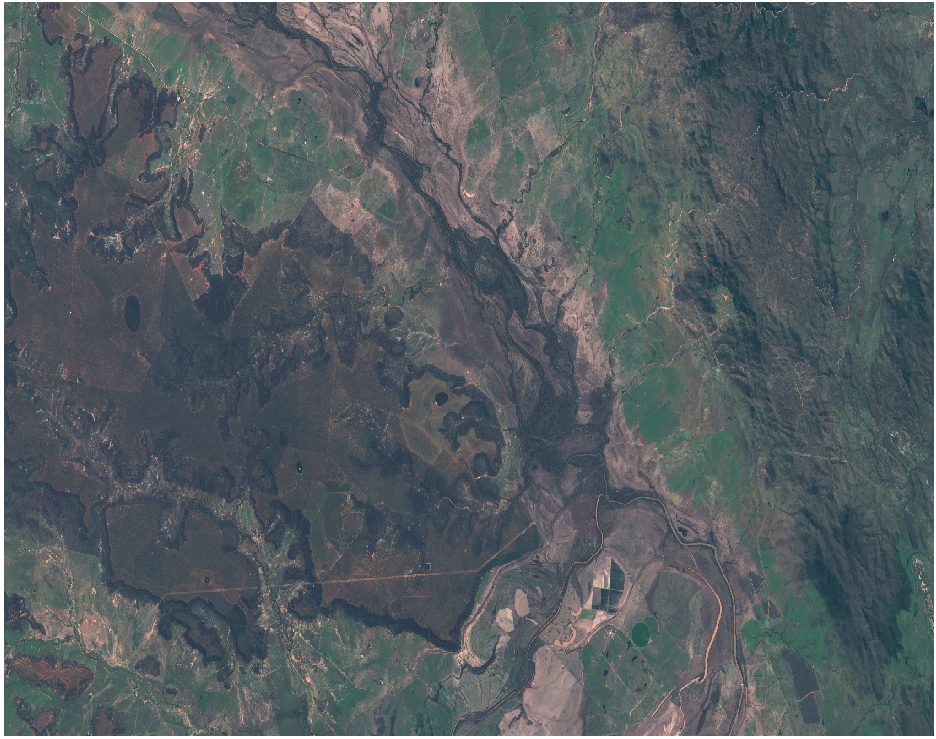}} \hspace{1mm}
\subfloat[May 2017]{\includegraphics[width= 5cm]{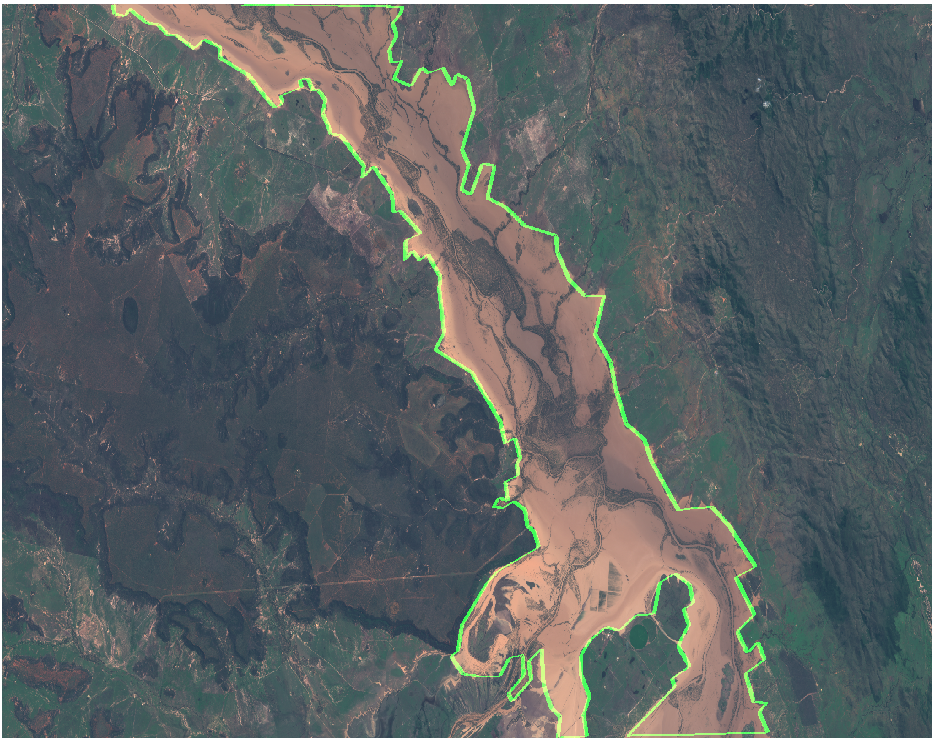}} \hspace{1mm}

\vspace{-0.3cm}
\subfloat[Aug 8th 2017]{\includegraphics[width=5cm]{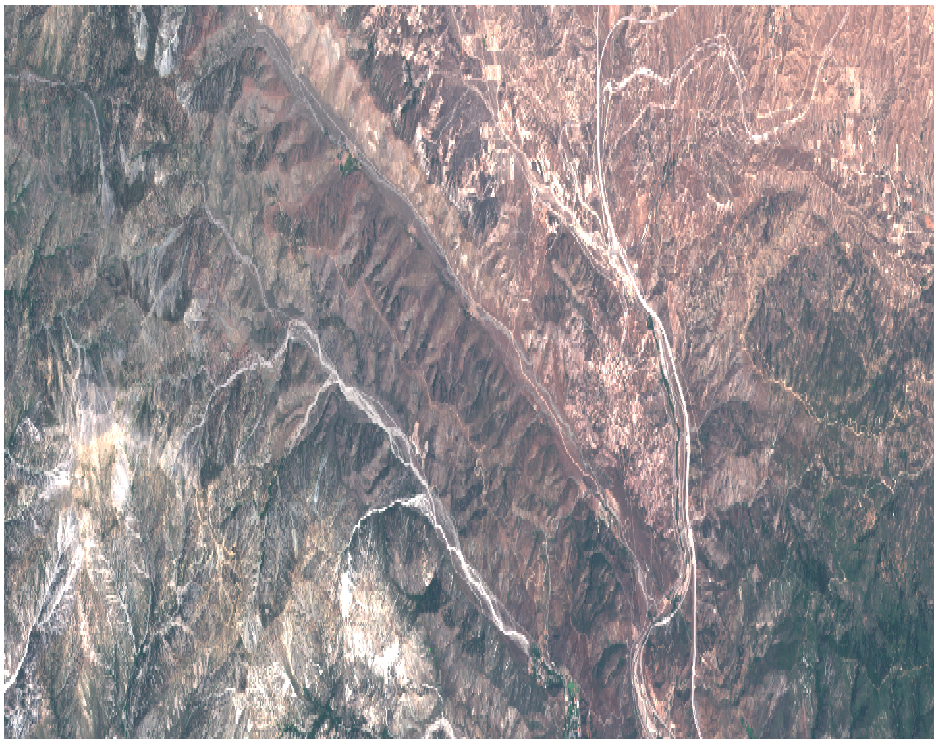}}\hspace{1mm}
\subfloat[Aug 28th 2017]{\includegraphics[width=5cm]{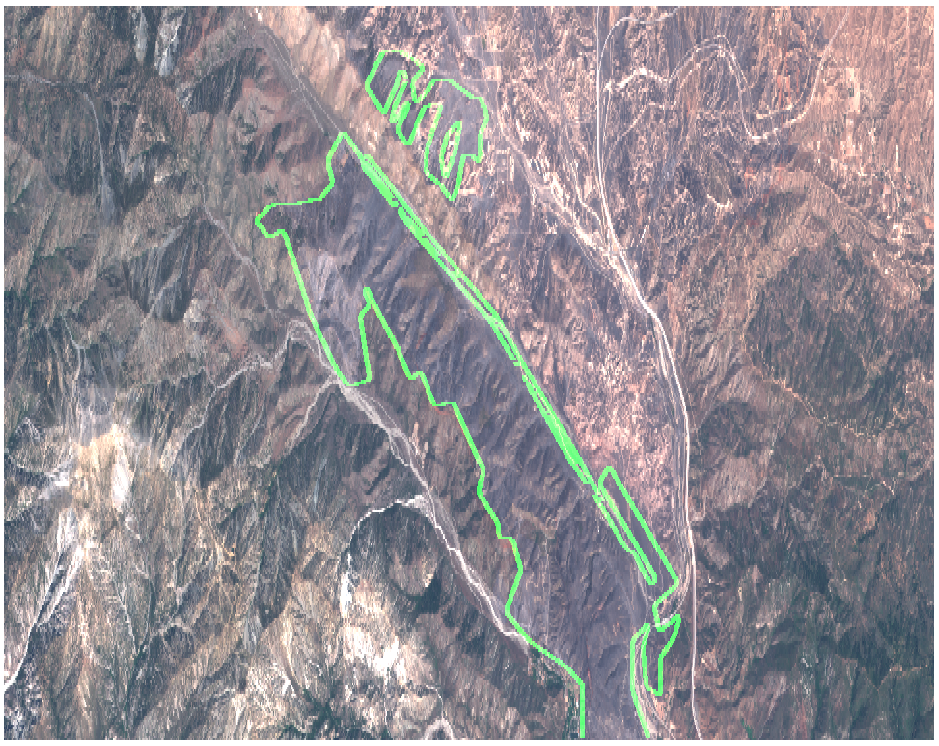}}\hspace{1mm}

\vspace{-0.3cm}
\subfloat[Jul 17th 2002]{\includegraphics[width= 5cm]{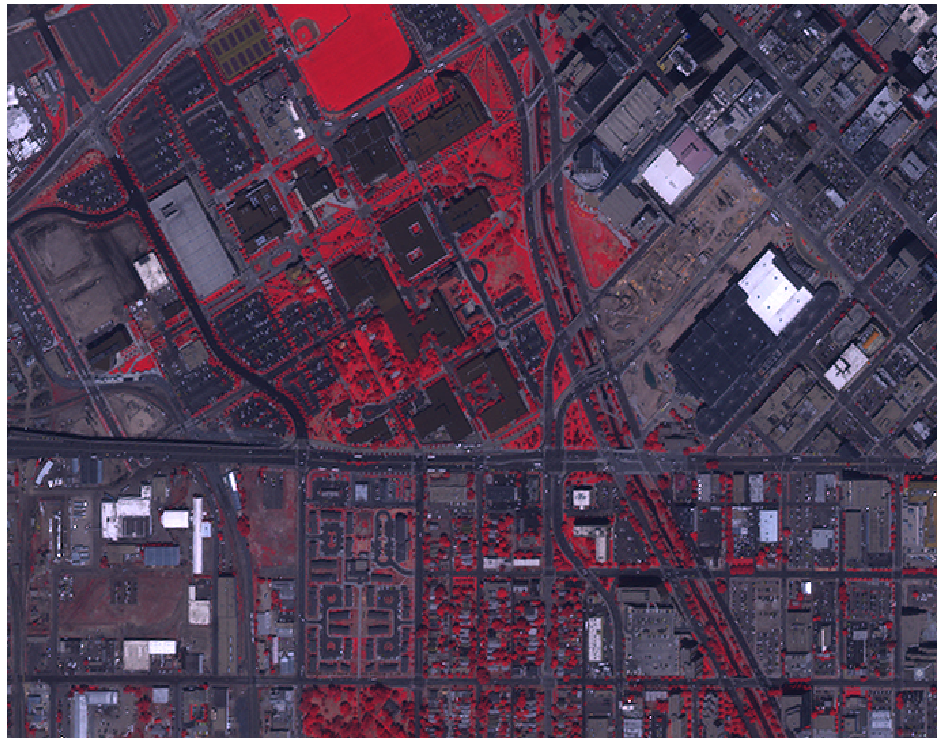}} \hspace{1mm}
\subfloat[Aug 22nd 2008]{\includegraphics[width= 5cm]{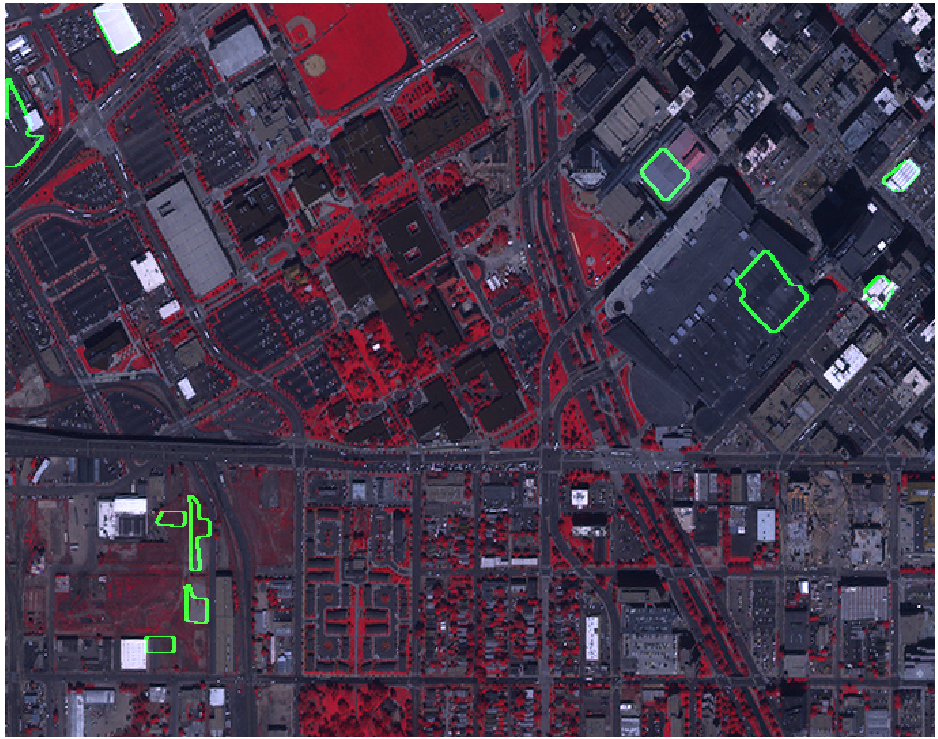}} \hspace{1mm}
\\

\subfloat[Oct 4th 2015]{\includegraphics[width= 5cm]{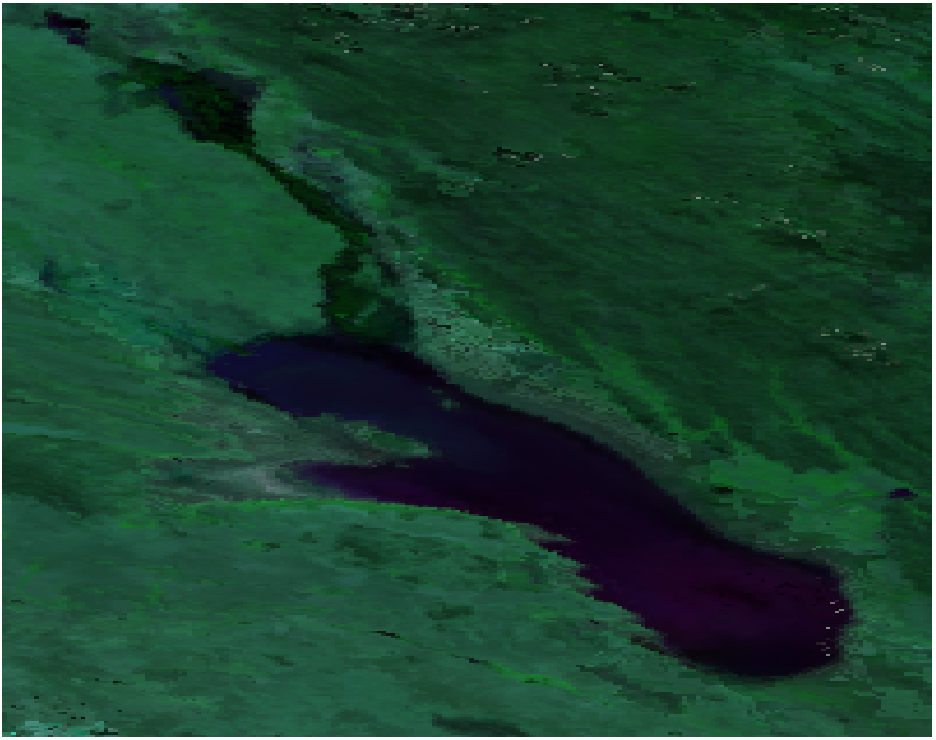}} \hspace{1mm}
\subfloat[Feb 20th 2016]{\includegraphics[width= 5cm]{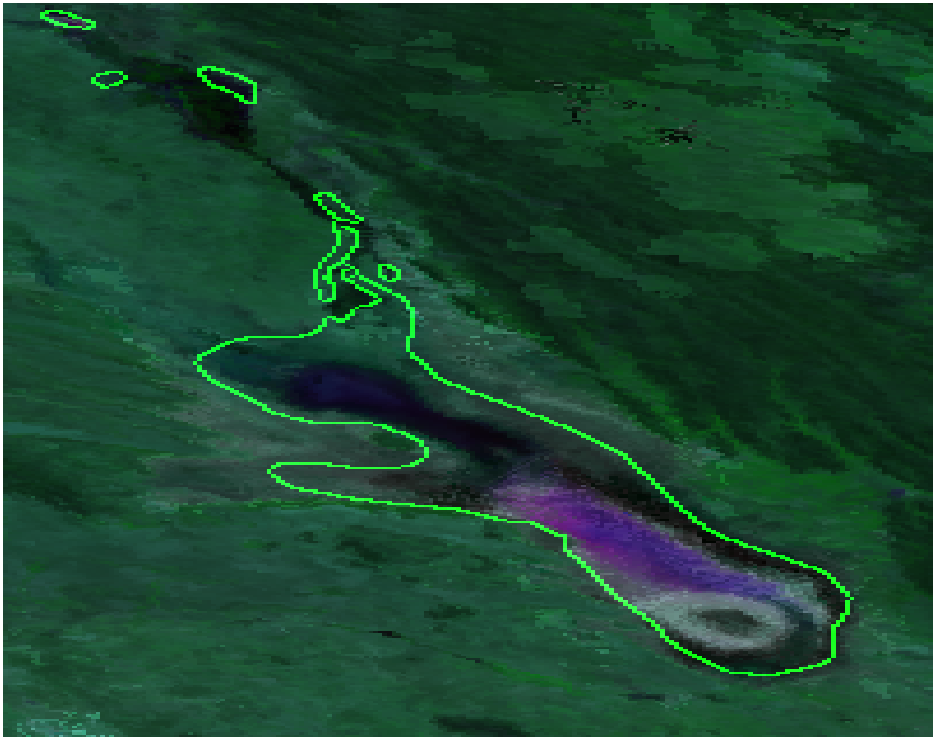}} \hspace{1mm}

\caption{\small Images with {\em natural} anomalies. Left column: images without anomaly. Right column: images with anomalous change and their corresponding ground truth in green. First row: we can see an area burned in Argentina between the months of July and August 2016. Second row: natural floods caused by Cyclone Debbie in Australia 2017. Third row: consequences of the fire in a mountainous area of California (USA). Fourth row: Quickbird multispectral images acquired over Denver city (USA) where appears an urbanized area. Last row: drying of Poopo Lake in Bolivia at the end of 2015.\label{fig:rgb2}}
\end{figure*}

We collected pairs of multispectral images, each pair consists of images taken at the same location but at different times. We selected the images in such a way that an anomalous change happened between the two acquisition times. We manually labeled the ground truth of each pair of images by photo-interpretation, and consequently we know where the anomalous change is present. All images contain changes of different nature, which allow us to study how the different algorithms perform in a diversity of realistic scenarios.
Table~\ref{table:database} exposes different descriptors of the images in the database. 
Fig.~\ref{fig:rgb2} shows the RGB composites of the pairs of images and the corresponding groundtruth.


\subsubsection{Numerical comparison} 
\begin{figure*}
     \centering
     \vspace{0.2mm}
     {\includegraphics[width= 4cm]{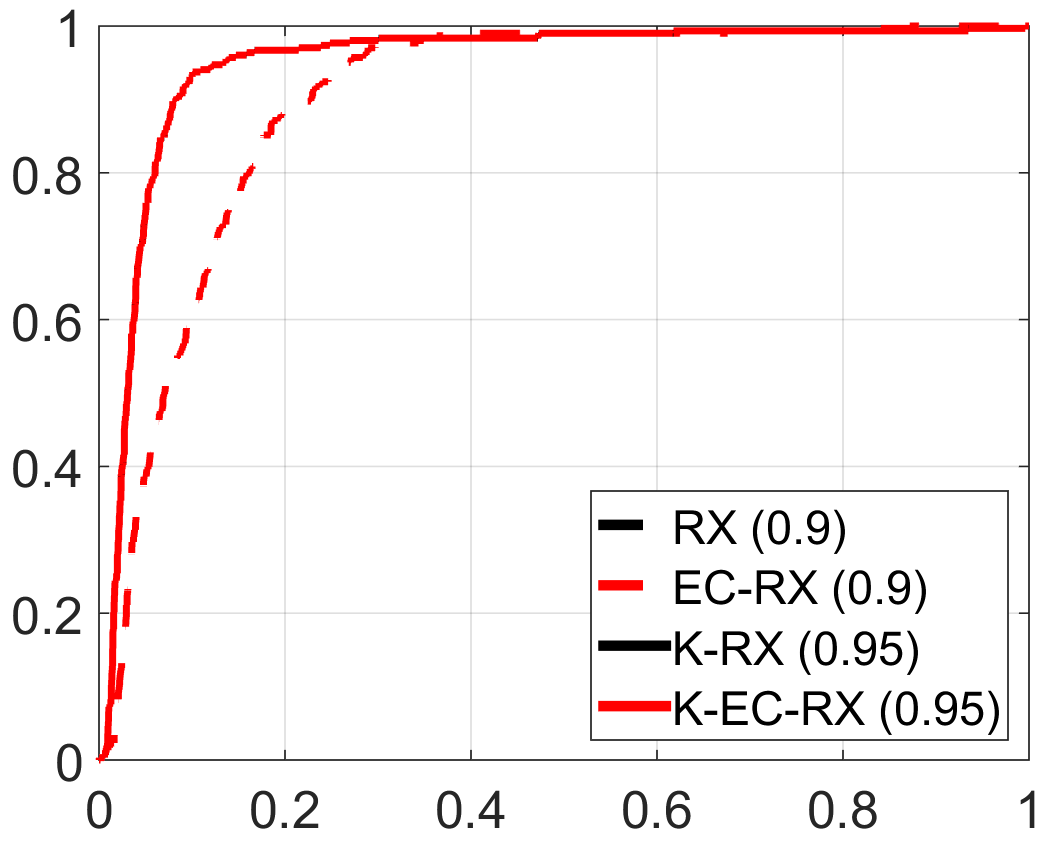}}
     \hspace{0.1mm}
     {\includegraphics[width= 4cm]{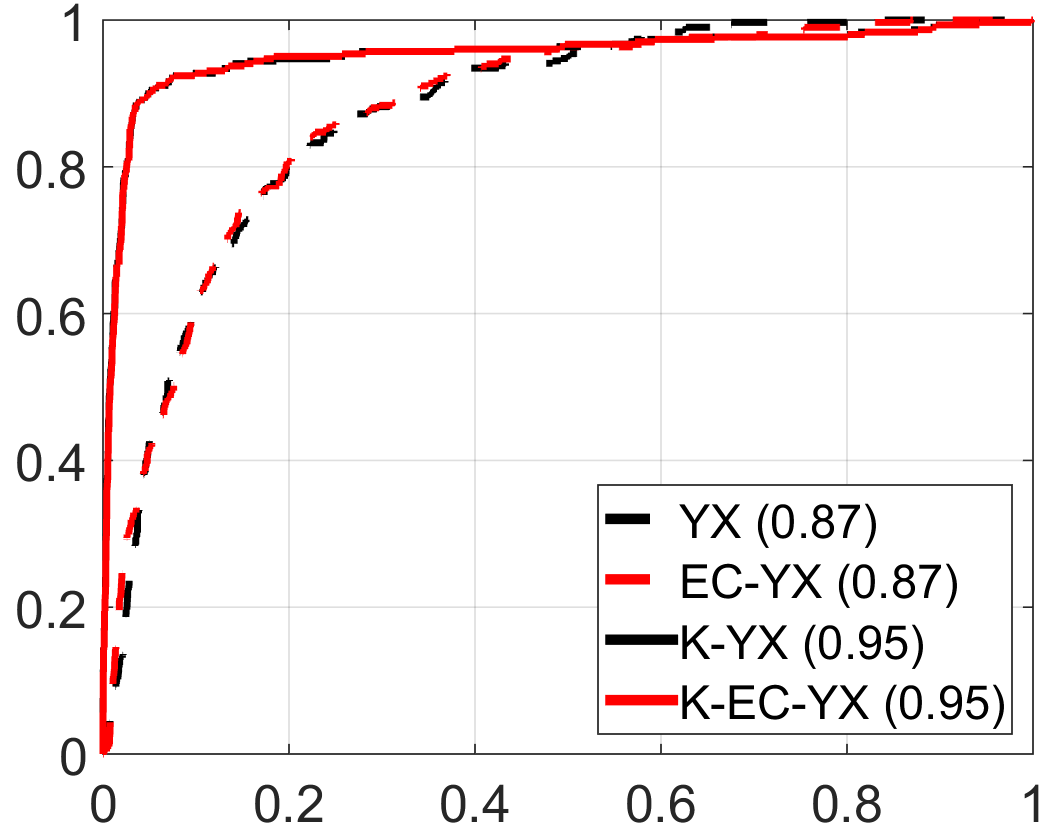}}
     \hspace{0.1mm}
     {\includegraphics[width= 4cm]{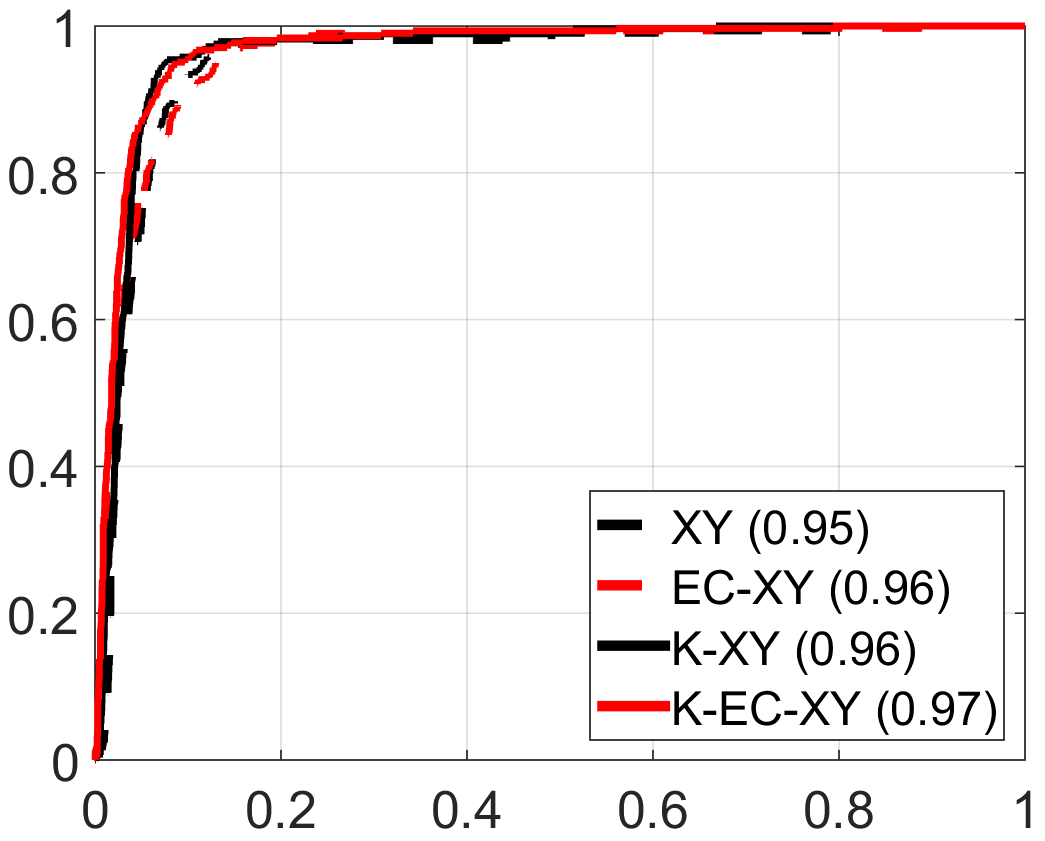}}
     \hspace{0.1mm}
     {\includegraphics[width= 4cm]{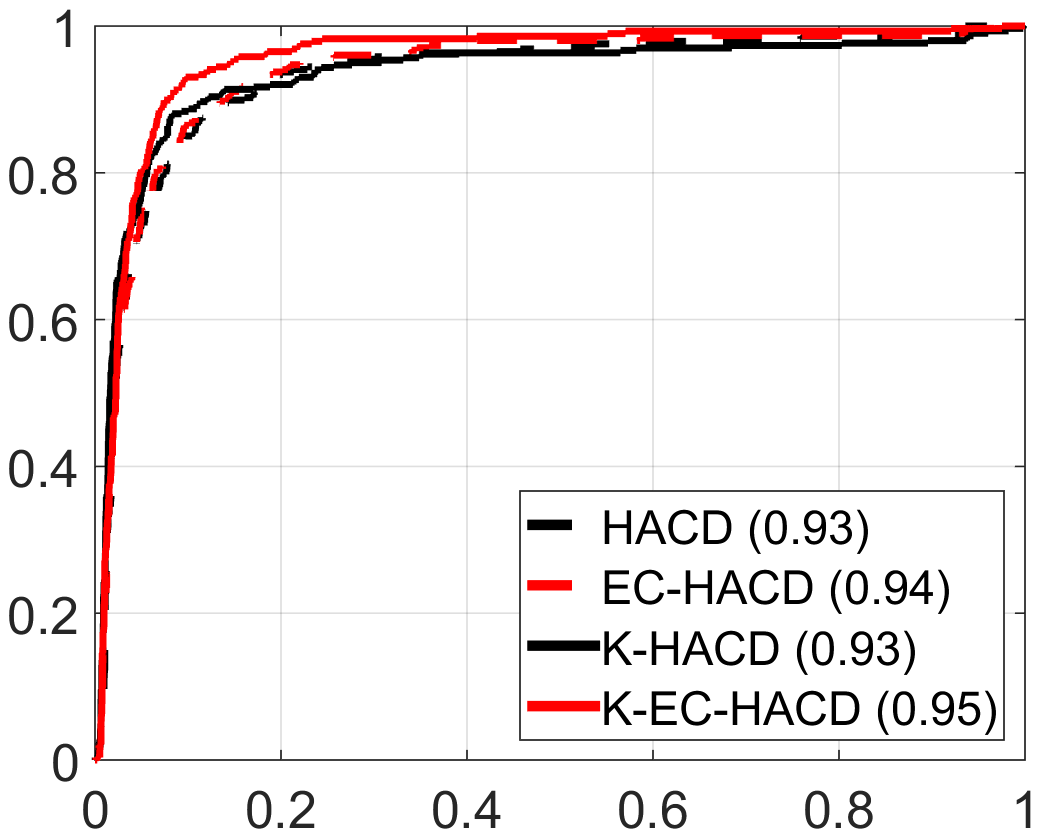}} \\
     \vspace{0.4mm}
     {\includegraphics[width= 4cm]{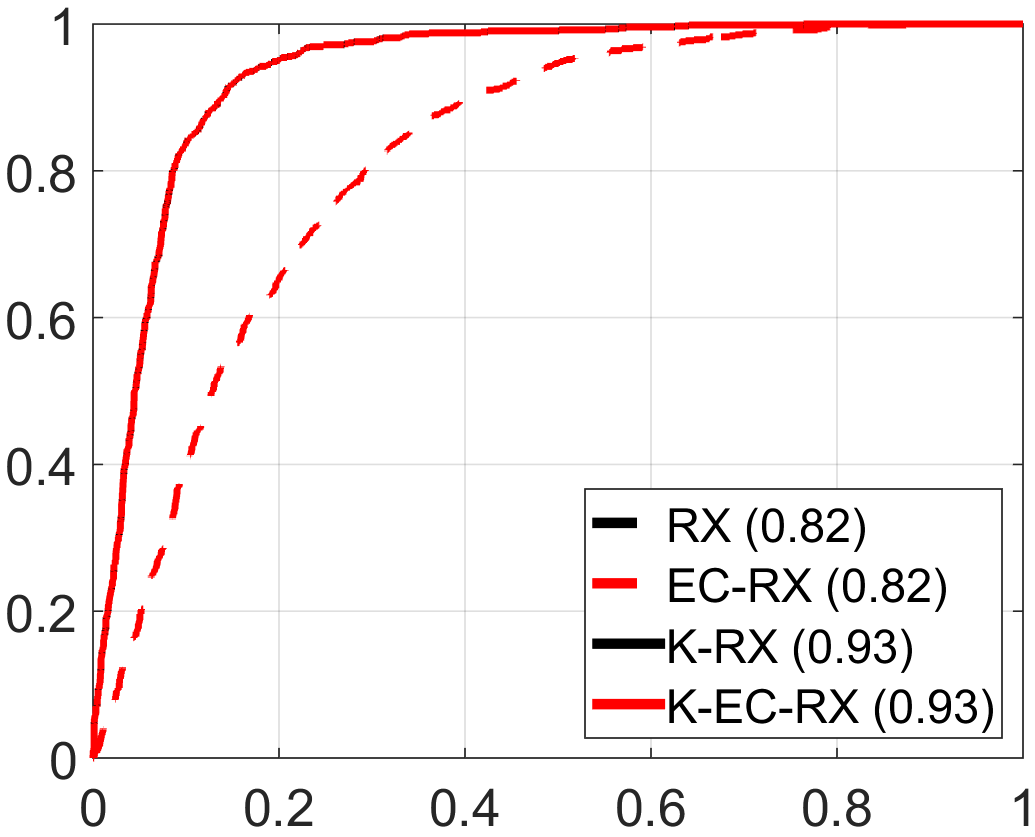}}
     \hspace{0.01mm}
     {\includegraphics[width= 4cm]{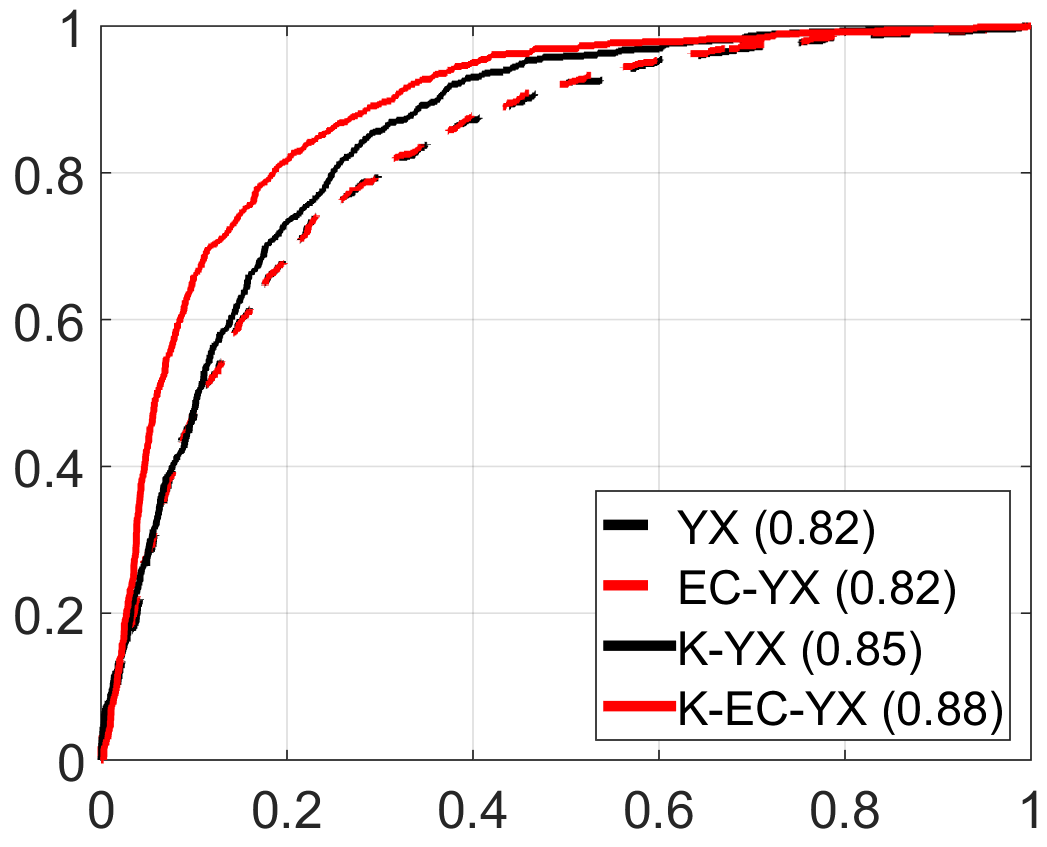}}
     \hspace{0.01mm}
     {\includegraphics[width= 4cm]{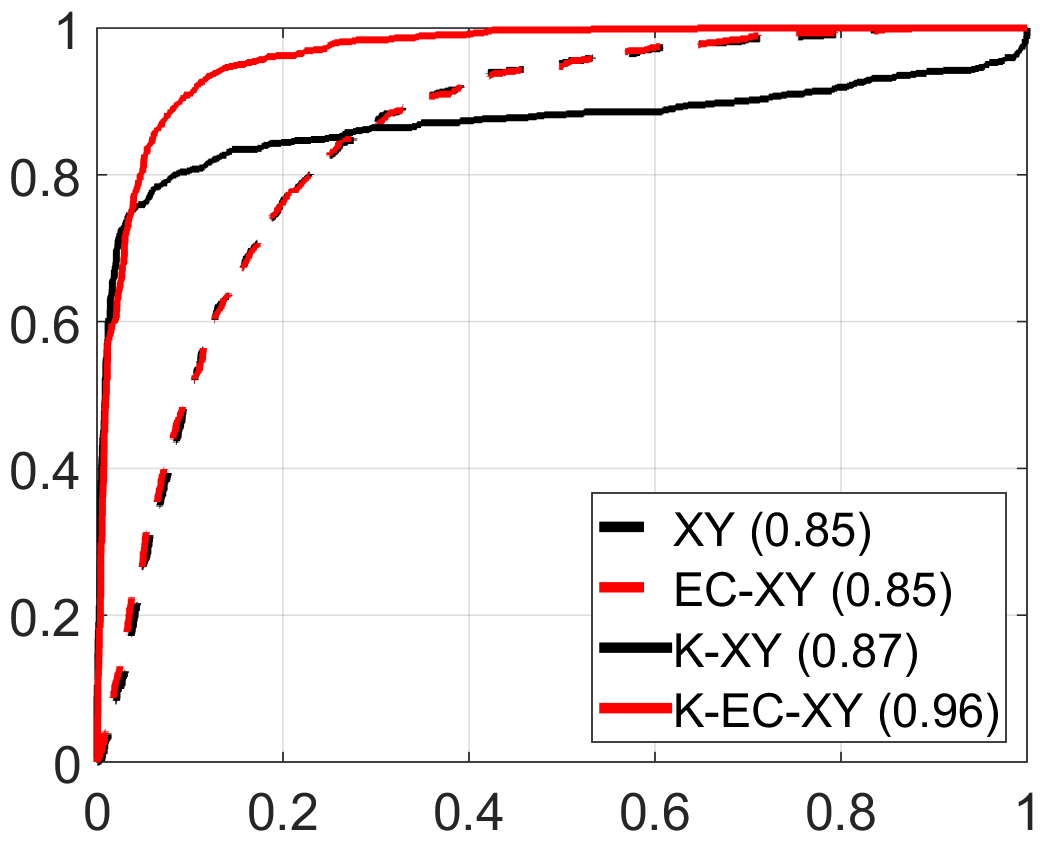}}
     \hspace{0.01mm}
     {\includegraphics[width= 4cm]{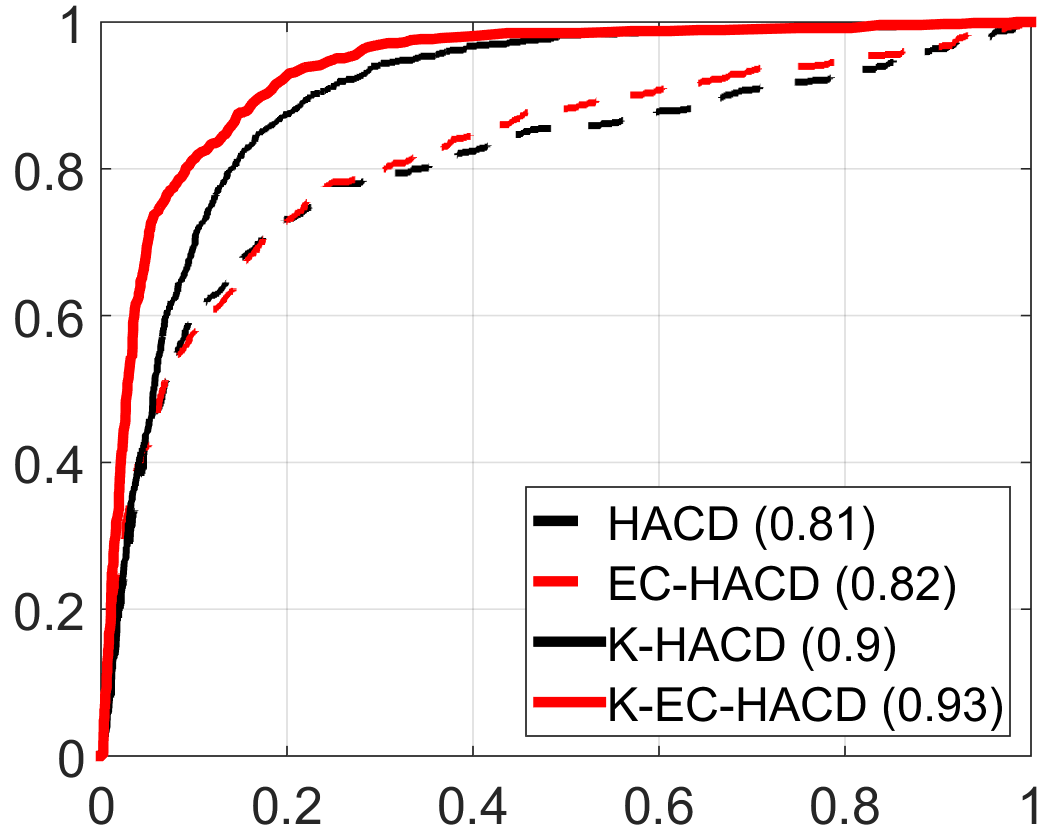}} \\
     \vspace{0.4mm}
      {\includegraphics[width= 4cm]{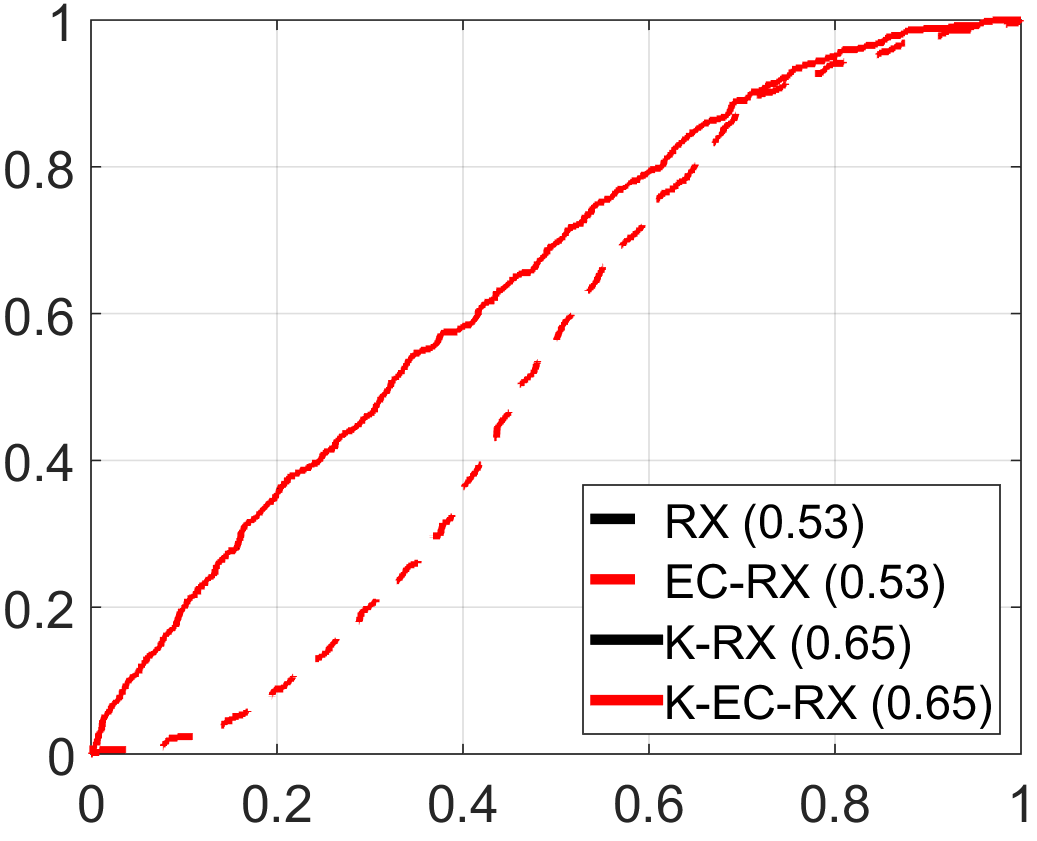}}
     \hspace{0.01mm}
     {\includegraphics[width= 4cm]{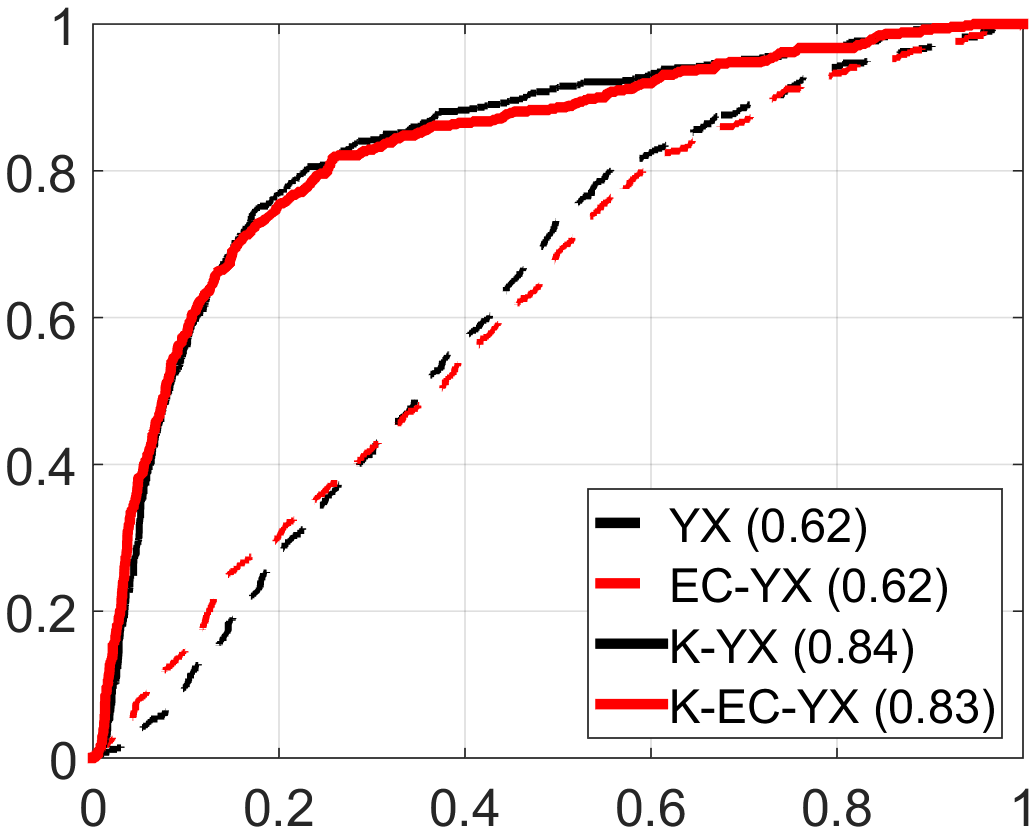}}
     \hspace{0.01mm}
     {\includegraphics[width= 4cm]{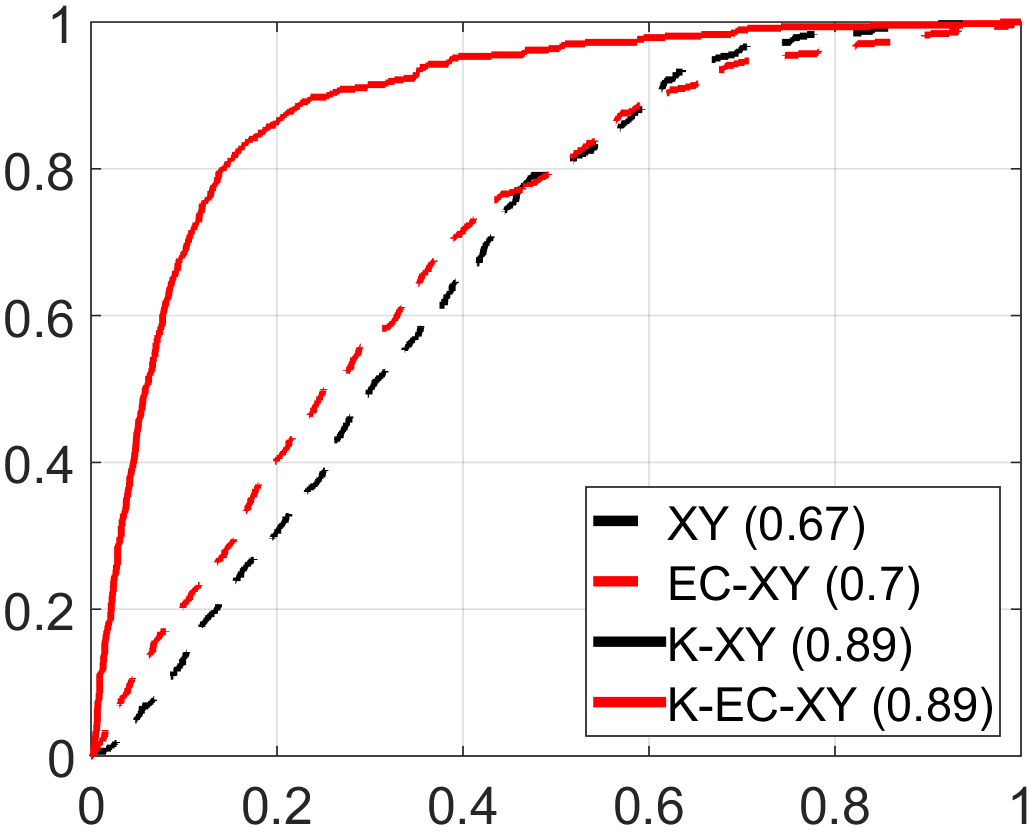}}
     \hspace{0.01mm}
     {\includegraphics[width= 4cm]{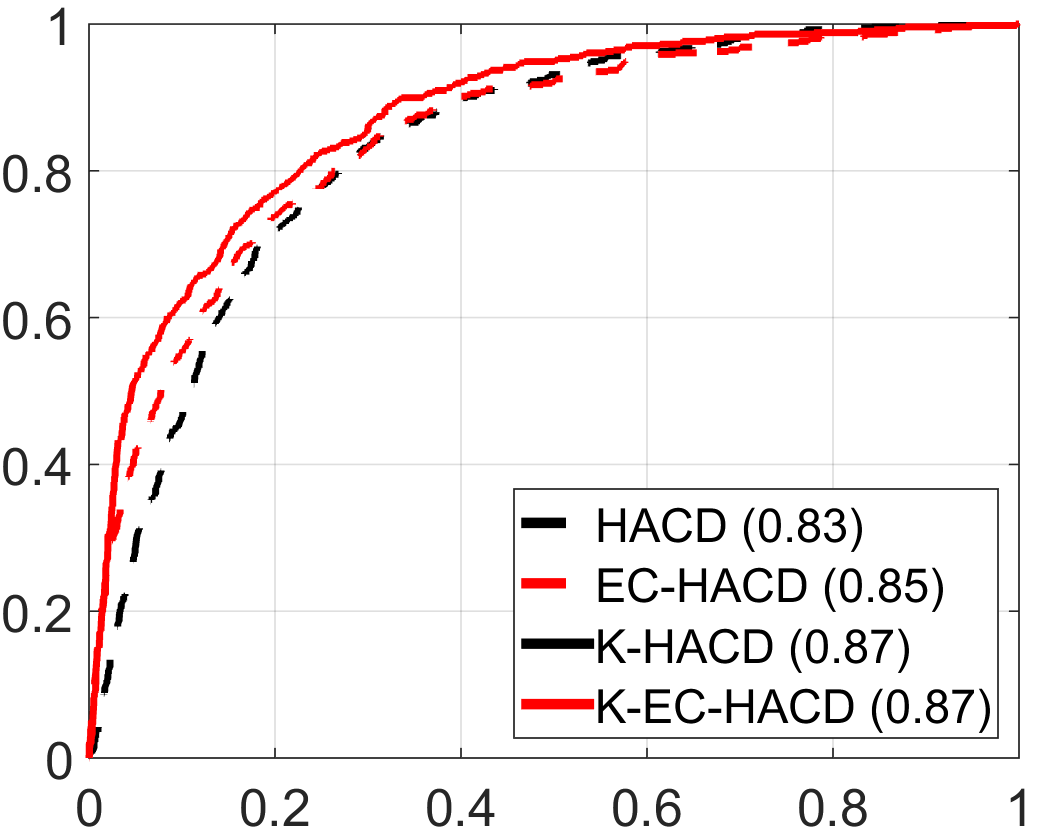}} \\
     \vspace{0.4mm}

      {\includegraphics[width= 4cm]{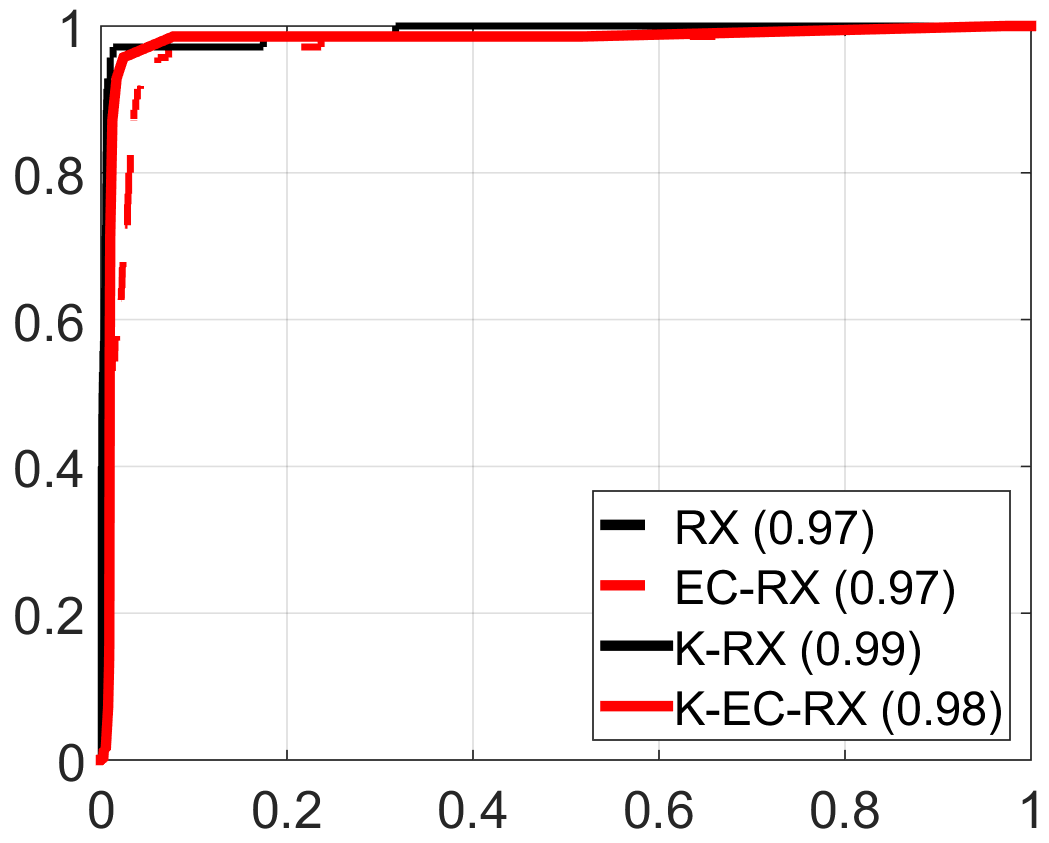}}
     \hspace{0.01mm}
     {\includegraphics[width= 4cm]{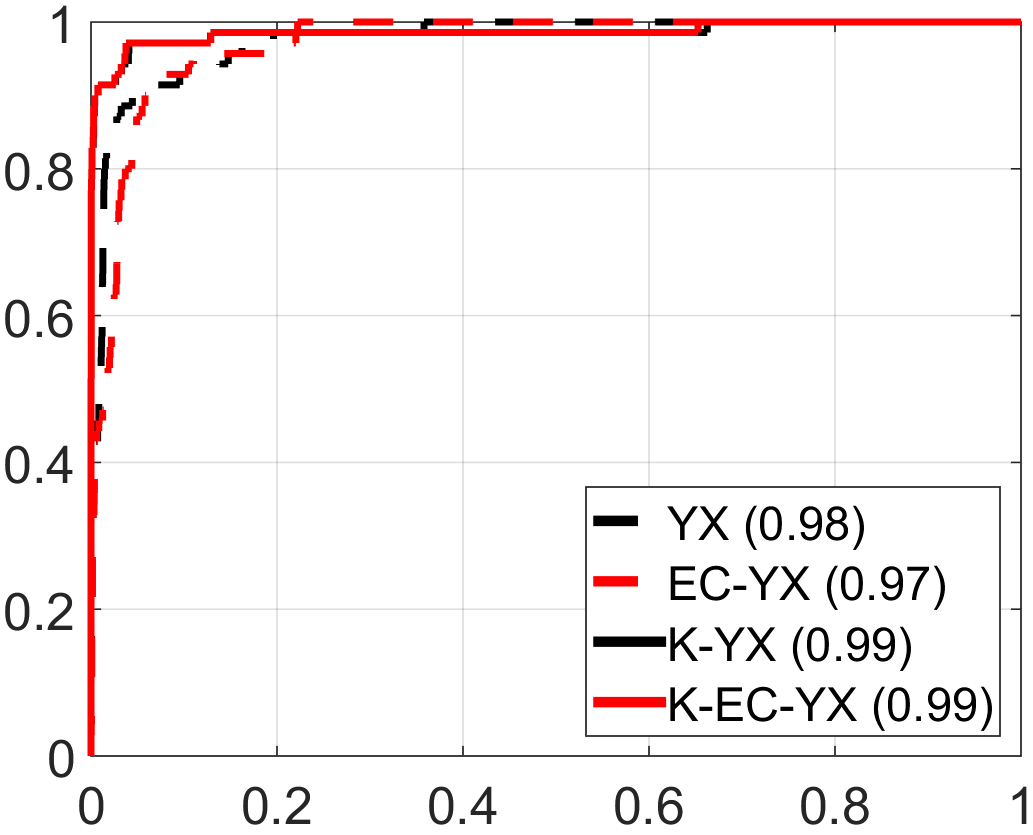}}
     \hspace{0.01mm}
     {\includegraphics[width= 4cm]{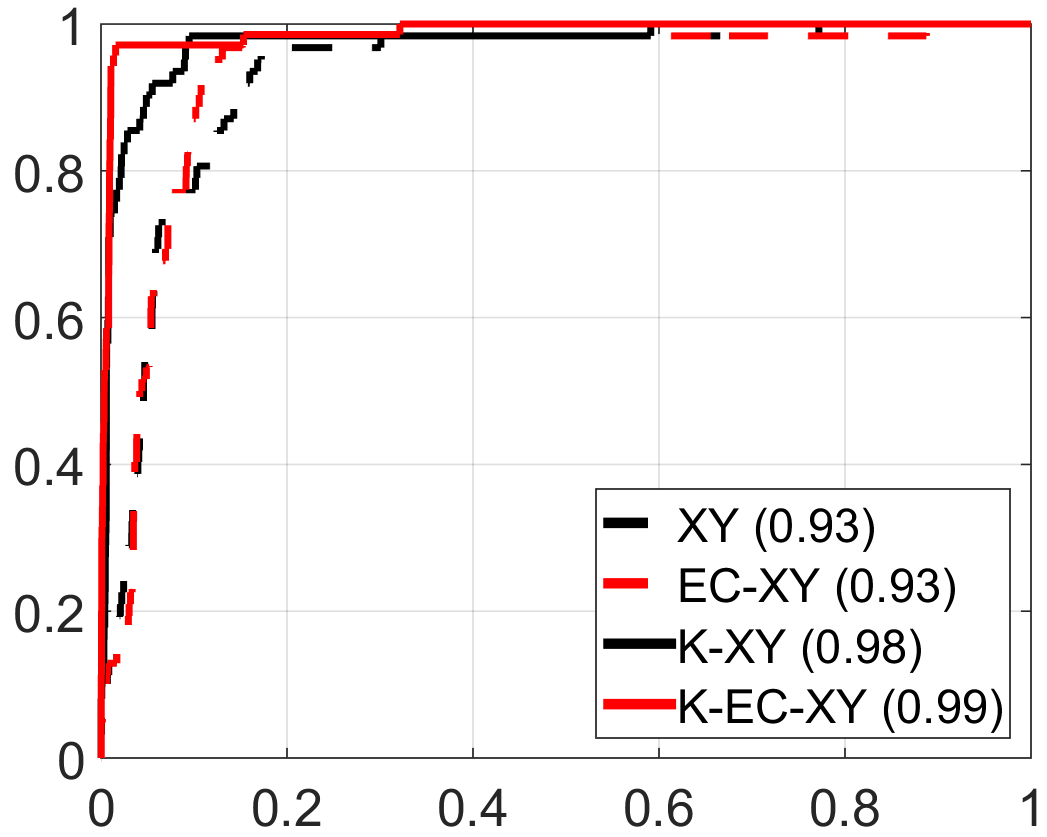}}
     \hspace{0.01mm}
     {\includegraphics[width= 4cm]{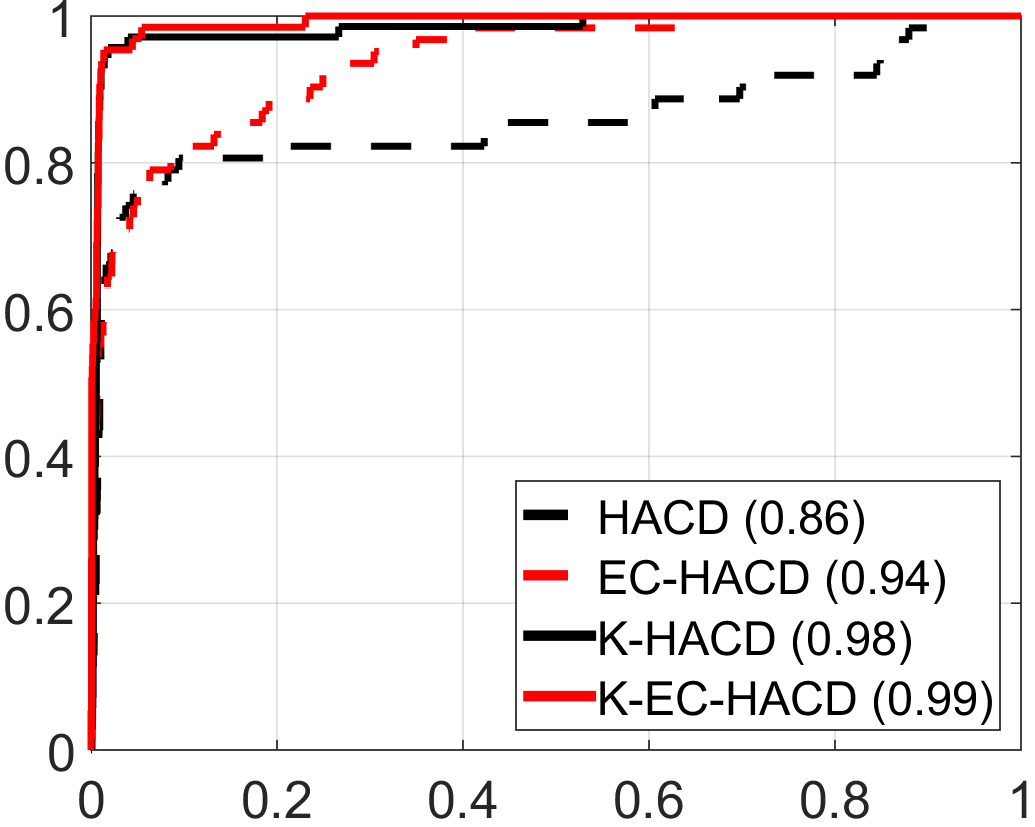}} \\
     \vspace{0.4mm}

      {\includegraphics[width= 4cm]{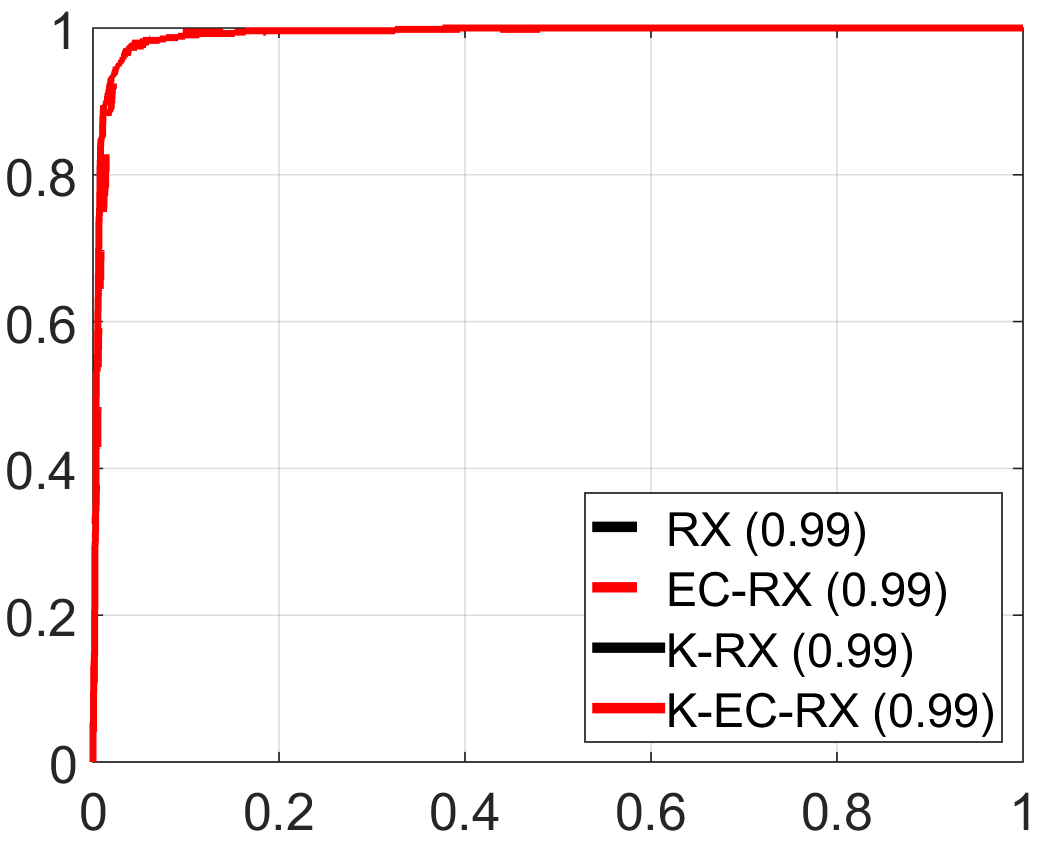}}
     \hspace{0.01mm}
     {\includegraphics[width= 4cm]{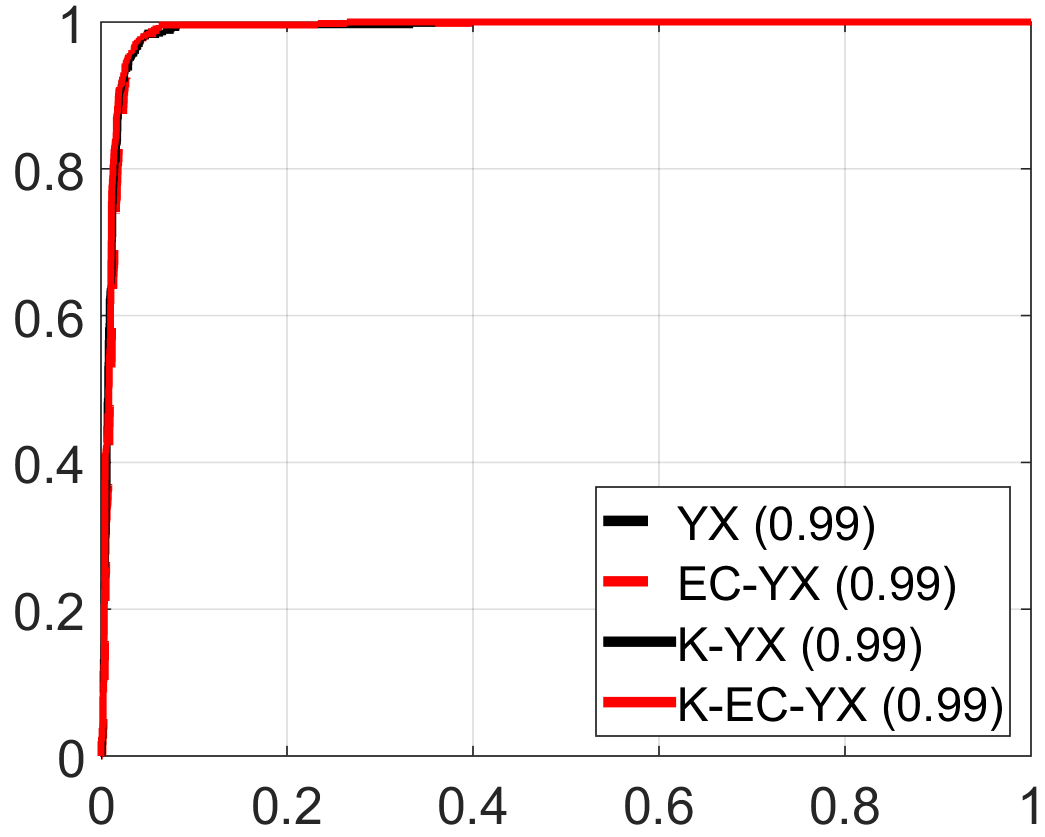}}
     \hspace{0.01mm}
     {\includegraphics[width= 4cm]{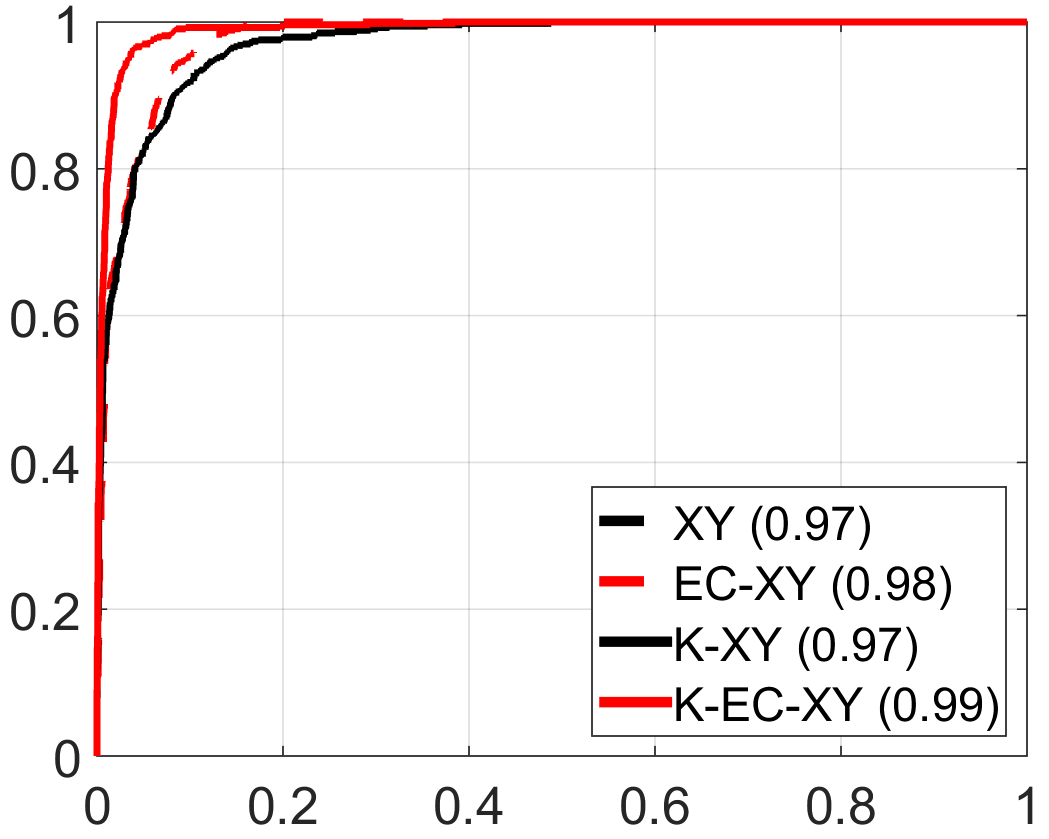}}
     \hspace{0.01mm}
     {\includegraphics[width= 4cm]{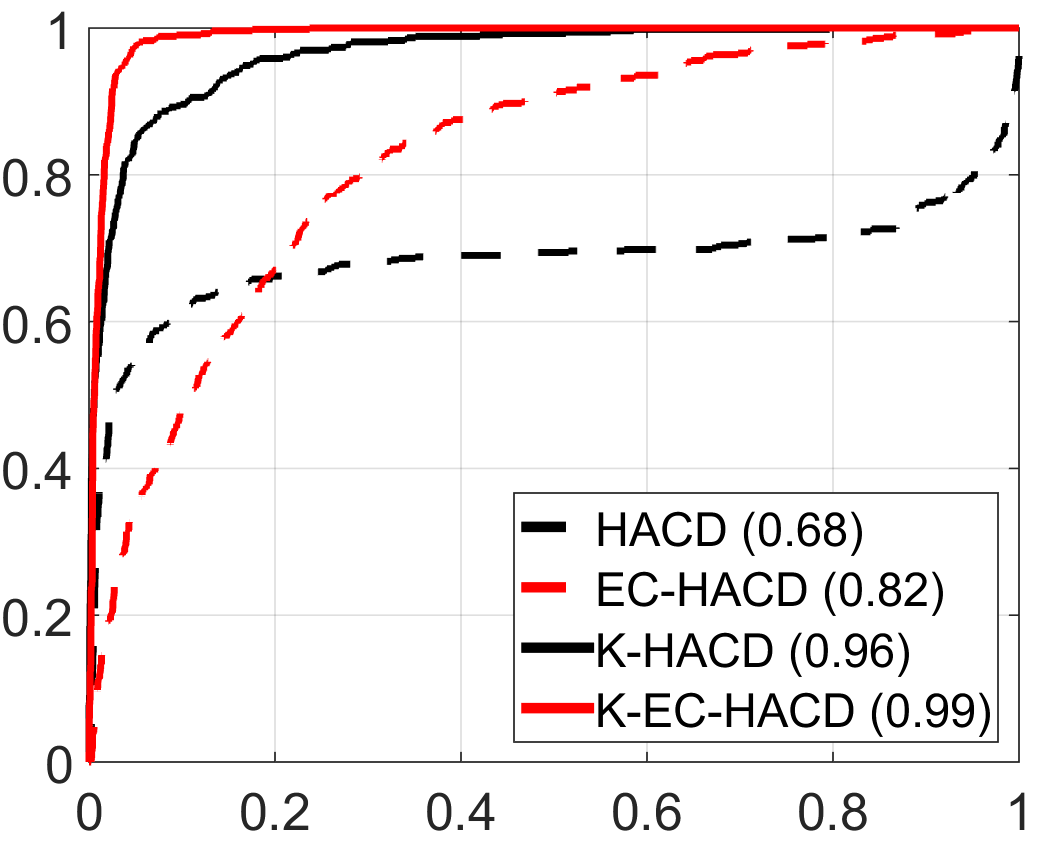}} \\
     \vspace{0.4mm}

     \caption{ROC curves and AUC obtained for natural changes on multispectral images. In rows the images from Argentina, Australia, California, Denver to Lake (see table \ref{table:database} for details) and in the columns from left to right the RX, YX, XY and HACD methods respectively.}
     \label{fig:ROC2}
 \end{figure*}

\begin{figure*}
     \centering
     
     
     \subfloat[HACD 50$\%$]{\includegraphics[width= 5cm]{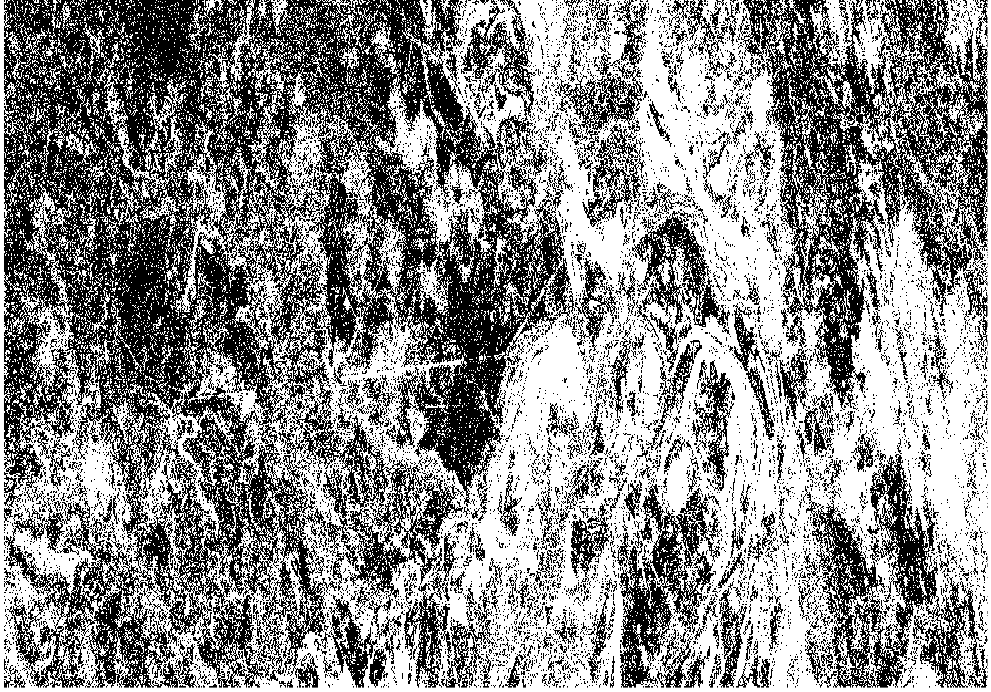}}
     \hspace{0.01mm}
     \subfloat[HACD 82$\%$]{\includegraphics[width= 5cm]{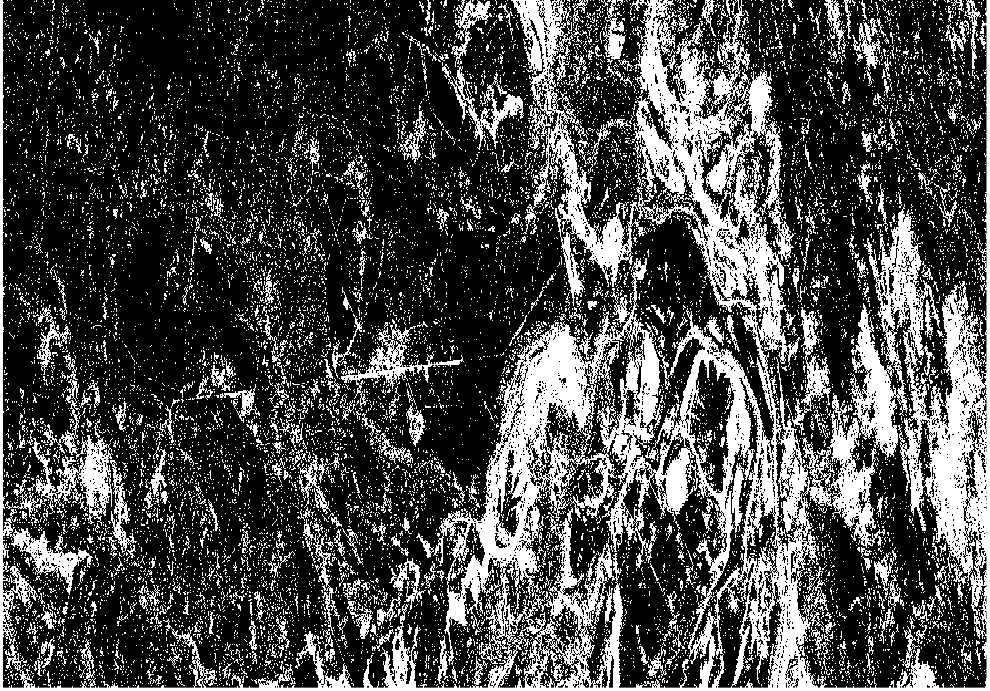}}
     \hspace{0.01mm}
     \subfloat[HACD 97$\%$]{\includegraphics[width= 5cm]{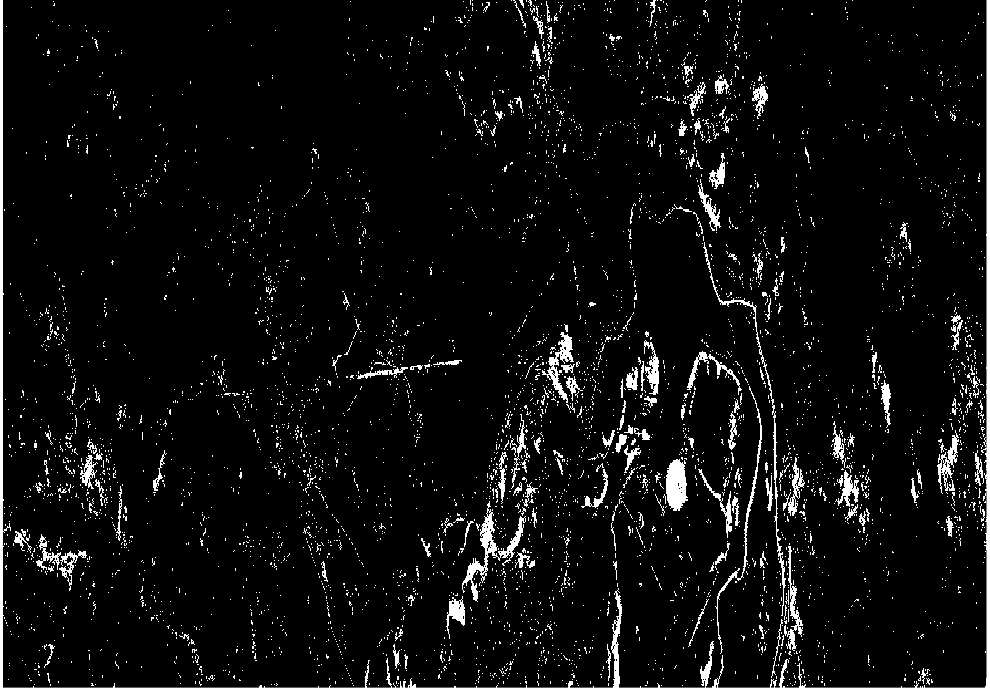}}\\
      \vspace{0.1mm}

     \subfloat[EC-HACD 50$\%$]{\includegraphics[width= 5cm]{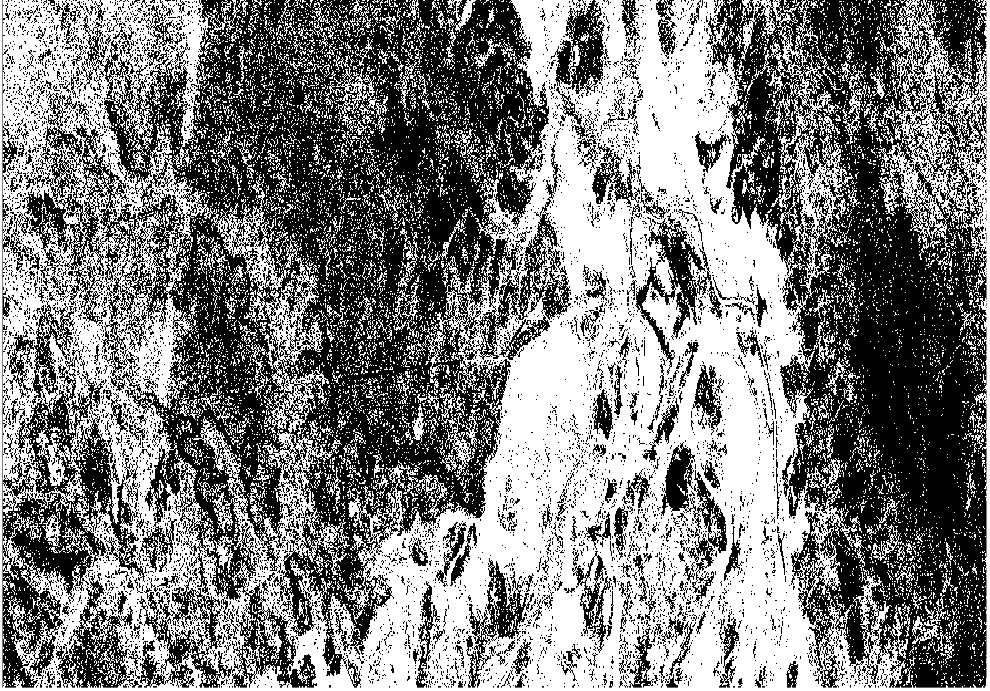}}
     \hspace{0.01mm}
     \subfloat[EC-HACD 82$\%$]{\includegraphics[width= 5cm]{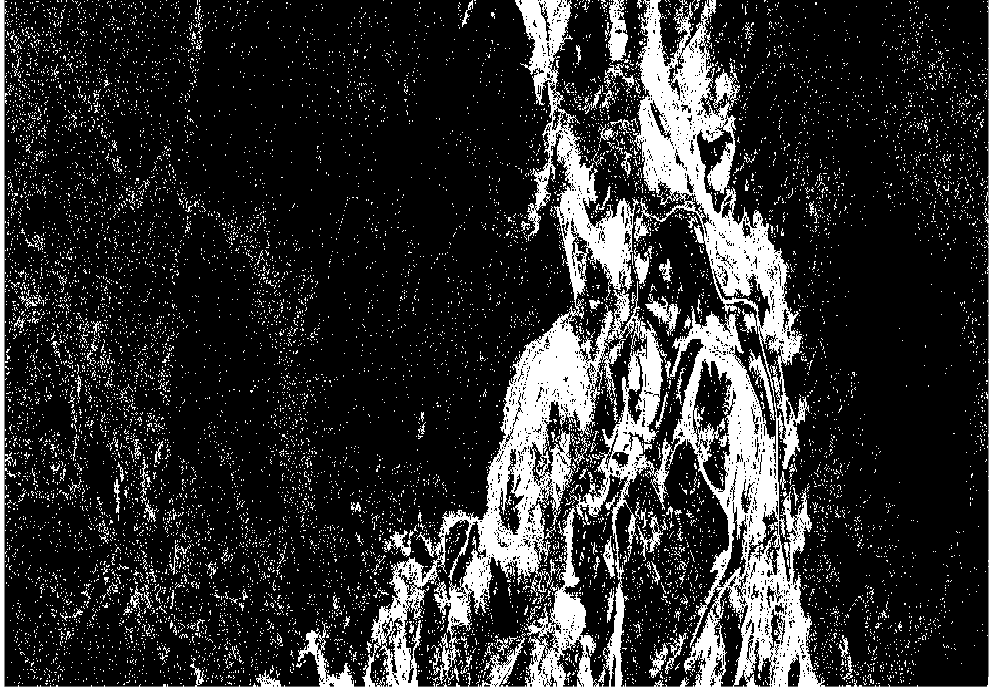}}
     \hspace{0.01mm}
     \subfloat[EC-HACD 97$\%$]{\includegraphics[width= 5cm]{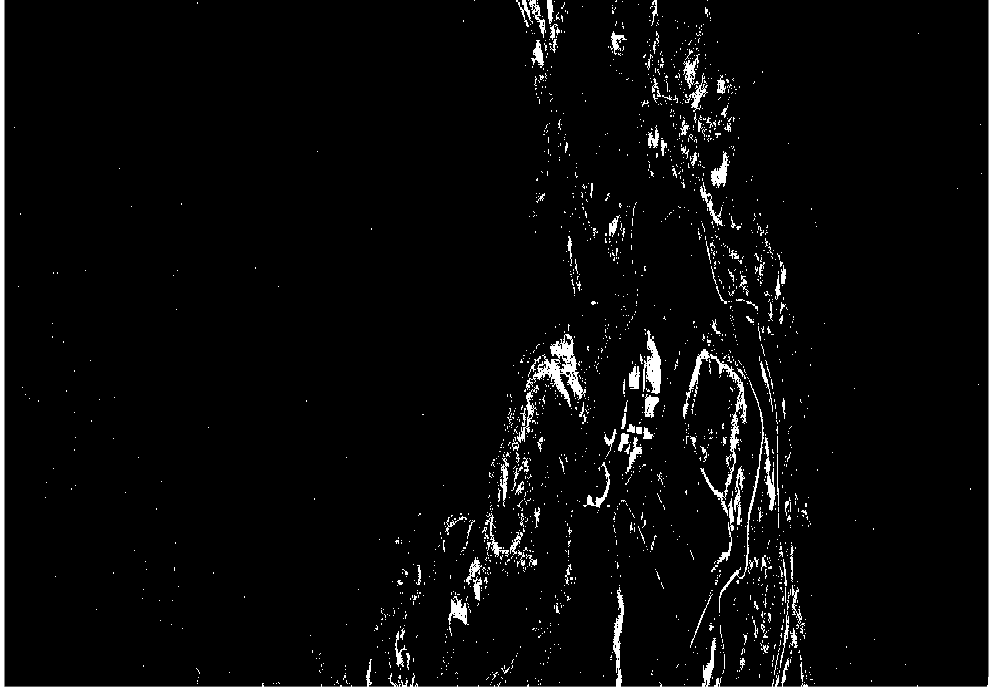}} \\
     \vspace{0.1mm}
     
     \subfloat[K-HACD 50$\%$]{\includegraphics[width= 5cm]{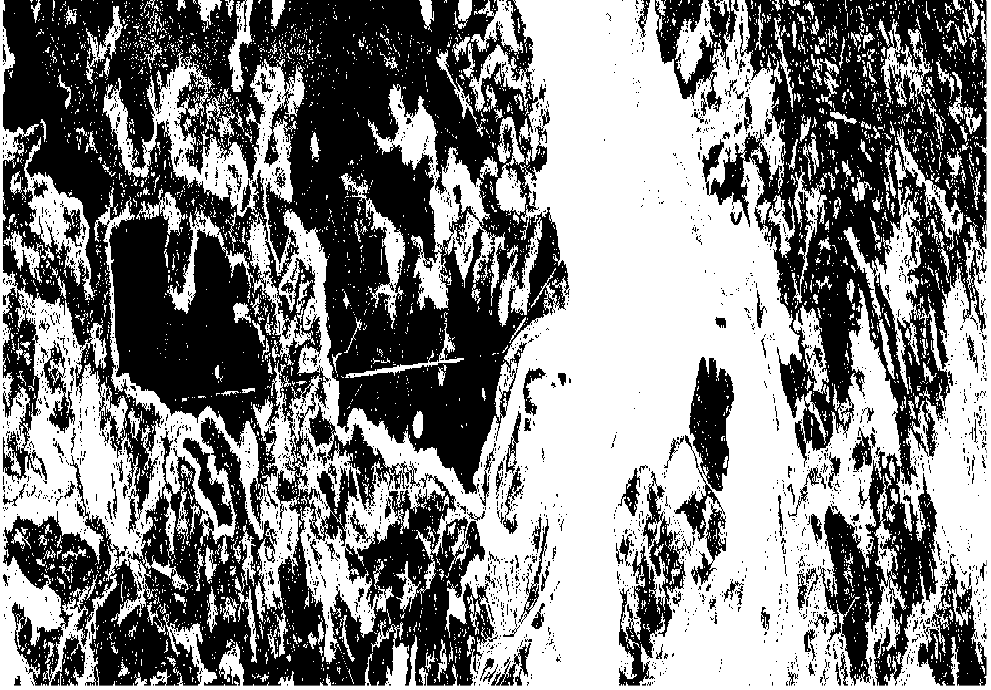}}
     \hspace{0.01mm}
     \subfloat[K-HACD 82$\%$]{\includegraphics[width= 5cm]{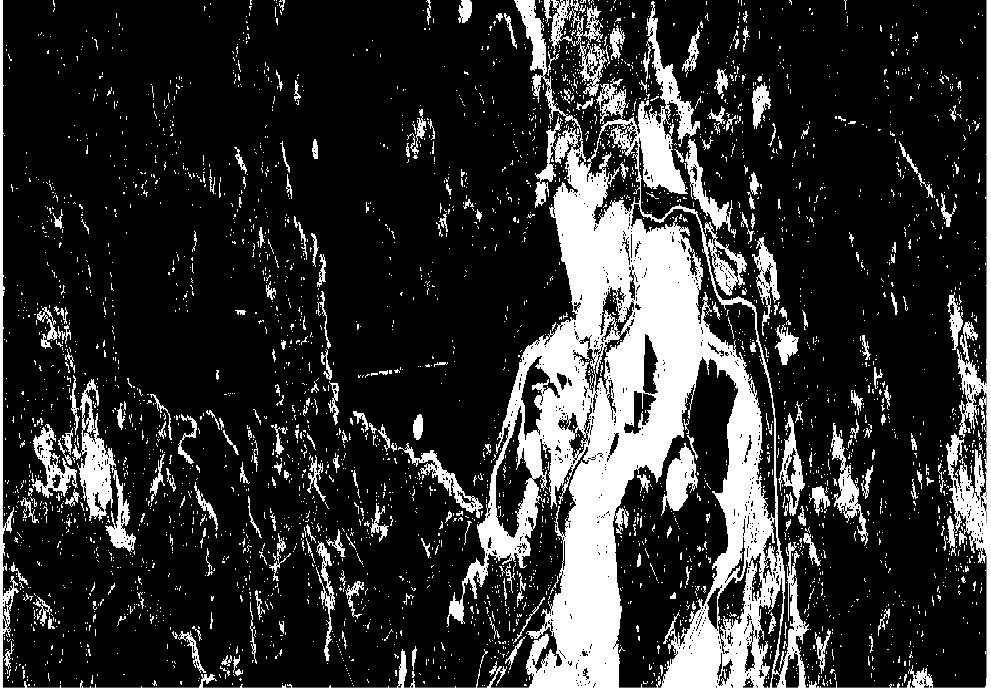}}
     \hspace{0.01mm}
     \subfloat[K-HACD 97$\%$]{\includegraphics[width= 5cm]{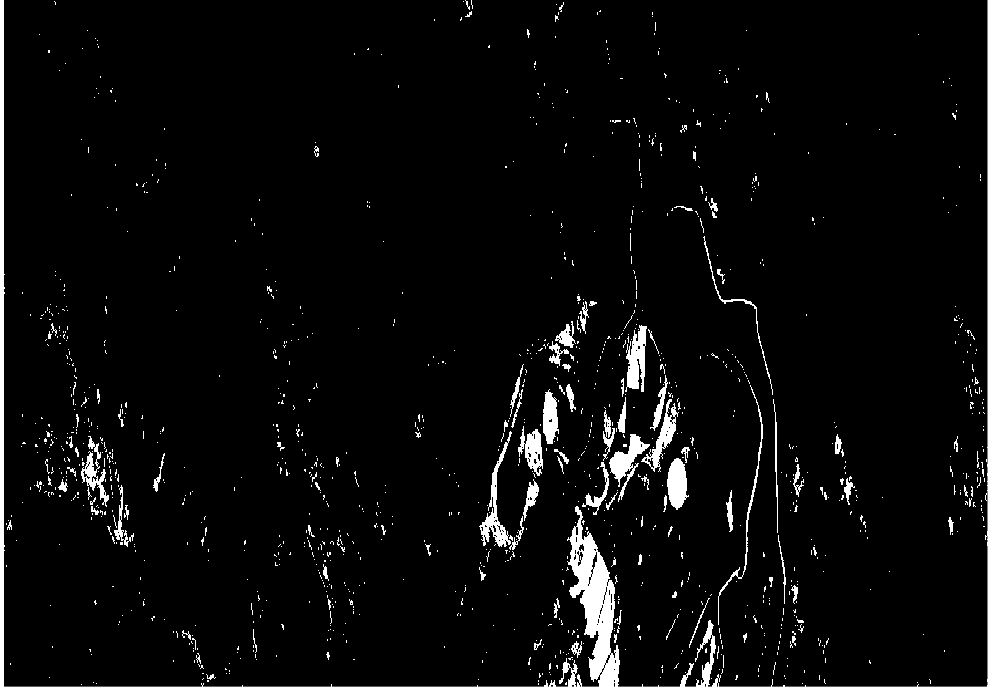}} \\
     \vspace{0.1mm}
     
     \subfloat[K-EC-HACD 50$\%$]{\includegraphics[width= 5cm]{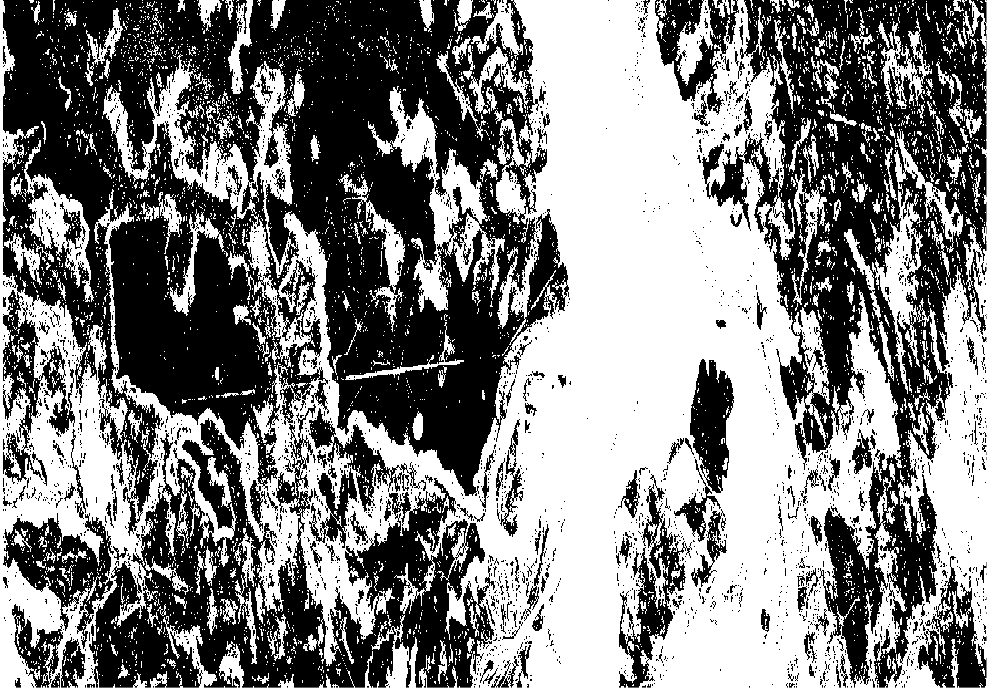}}
     \hspace{0.01mm}
     \subfloat[K-EC-HACD 82$\%$]{\includegraphics[width= 5cm]{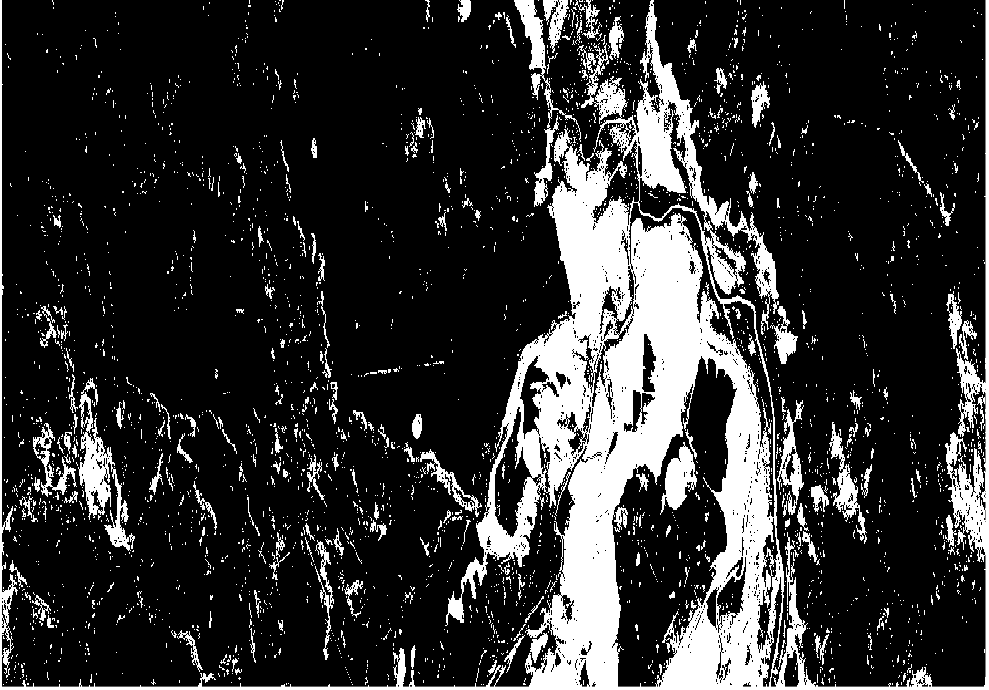}}
     \hspace{0.01mm}
     \subfloat[K-EC-HACD 97$\%$]{\includegraphics[width= 5cm]{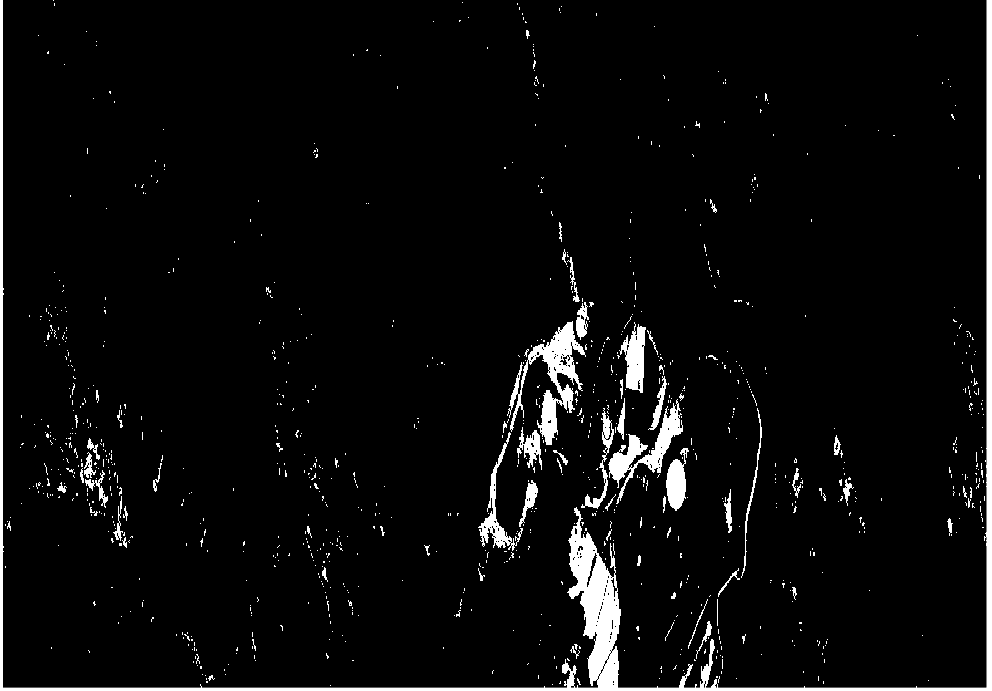}} \\
     \vspace{0.1mm}
     
     \caption{HACD Predictions Maps from Australia test. Left : lower threshold, Center : optimal threshold and Right : high threshold.}
     \label{fig:Pred1}
 \end{figure*}

Different considerations have to be taken when using the different algorithms. On the one hand the family of methods based on EC distribution involve the optimization of the ${\nu}$ parameter. On the other hand kernel methods involve fitting the kernel function parameters. In this experiment we use the classical $rbf$ kernel which is well suited for multispectral images and has only one parameter, $\sigma$. In addition an extra parameter $\lambda$ has to be fitted to regularize the matrix inversion. Selecting properly all these three parameters is an issue. An ideal situation would be having a rule of thumb to choose them. We performed preliminary experiments to explore the applicability of several existing rules to estimate the $\sigma$ parameter. For the different images and problems faced in this section we applied the heuristics and tried to find an heuristic for the ${\nu}$ and $\lambda$ parameters. In particular we investigated ten different heuristics: average distance between all samples, median of the distance between all samples, squared root of the dimensionality times variance per dimension averaged, median of the Silverman's rule \cite{Silverman86}, median of the Scott's rule per feature \cite{Scott}, maximum likelihood density estimation, maximum Bayes estimate, maximum entropy estimate, average estimate of marginal kernel density estimate, and kernel density estimation using Gaussian kernel. While some of them have good performance for particular problems none of the rules was useful in general (results not shown). This is a usual problem in ACD where for instance, instead of setting a particular anomaly threshold, it is usual to compute the ROC curve where all the thresholds are evaluated \cite{Theiler10}. Since we found that all the rules failed we proposed to use a cross-validation scheme to fit all the involved parameters: $\sigma$, $\lambda$, and $\nu$. The same procedure was used for all the algorithms.
\\
For each pair of images, we split them into two parts, and we use one for training and one for testing. We select the best parameters by greed search in a cross-validation scheme, using $1000$ training samples and $4000$ validation samples randomly selected form the training set. Each method implies different set of parameters. For the ${\nu}$ parameter we explore $100$ points logarithmically spaced between $[10^{-5},10^{10}]$. For $\sigma$ parameter, we explore around the heuristic of the mean of the Euclidean distance between pairs of points (which was the most successful in the preliminary experiments), we make a grid by taking 60 logarithmically spaced points respectively between [$10^{-3},10^{3}$] multiplied by the heuristic value. For the $\lambda$ parameter we use $30$ values logarithmically spaced between [$10^{-10},10^{2.5}$]. We optimized the different methods by maximizing the area under the curve (AUC) of the receiver operating characteristic (ROC) curve and use the best parameters in the training set for the validation set. 
\\
In Fig.~\ref{fig:ROC2} the ROC curves for all the images and all the families of methods are shown. Table~\ref{ROC_table} summarizes the AUC values of these ROC curves.  
Fig.~\ref{fig:Pred1} shows an example of the spatial distribution of the predictions for the HACD family. We use three different thresholds: one that produces a high rate of detection, one which detect the exact proportion of anomalous changes present in the groundtruth (82$\%$), and finally one that produces low detection rate. It can be seen how even in low and high detection rates the kernel methods produce more spatially compact detections which brings to much more realistic maps. Fig.~\ref{fig:Pred1} compares the ability of the best linear method against the best kernel method when using the optimal threshold. Again kernel methods produce maps with less both false positives and negative alarms.

As a summary, the kernel version achieves the best results in all the images when compared with its linear counterpart. Of the $16$ detectors under study there is not an overall winner because each detector has its own characteristics (that can relatively fit data particularities), and the parameters are adjusted according to the type of image. We can see that the K-ACD version obtains better performance both over the linear ACD, and over the linear EC-ACD. And the K-EC-ACD versions have better performance than the rest. For each type of detector (i.e. RX, XY, YX, or HACD) the AUC values can be ranked as: K-EC-ACD $\geqslant$ K-ACD $\geqslant$ EC-ACD $\geqslant$ ACD.


\begin{table}[t!]
\hspace{-1cm}
\centering
\caption{AUC results for all five images. First and second best values for each image and each member of the family are in bold. We provide the mean and the standard deviation (in parenthesis) for ten different trials, values marked with (*) had an outlier so we give the median instead of the mean.}
\label{ROC_table}
\scriptsize
\begin{tabular}{ |l|l|l|l|l| } 
\hline
\hline
\rowcolor[HTML]{8BCA8B} \bf{METHODS}  & \bf{RX} & \bf{YX} & \bf{XY} & \bf{HACD}   \\ \hline\hline
          \rowcolor[HTML]{8BCA8B}       \multicolumn{5}{|c|}{\bf{ARGENTINA}}      \\\hline 
ACD          & \textit{0.88 $\pm 0.008$} & \textit{0.86 $\pm 0.010$}  & \textit{0.95 $\pm 0.004$} & \textbf{0.93 $\pm 0.007$} \\ \hline
K-ACD          & \textbf{0.93 $\pm 0.009$} & \textbf{0.94 $\pm 0.007$}  & \textbf{0.95 $\pm 0.011$} & \textbf{0.93 $\pm 0.005$} \\ \hline
EC-ACD        & \textit{0.88 $\pm 0.008$} & \textit{0.86 $\pm 0.010$}  & \textit{0.95 $\pm 0.004$} & \it{0.93 $\pm 0.006$}     \\ \hline
K-EC-ACD      & \textbf{0.93 $\pm 0.009$} & \textbf{0.94 $\pm 0.008$}  & \textbf{0.96 $\pm 0.008$} & \bf{0.95 $\pm 0.007$}     \\ \hline\hline

 \rowcolor[HTML]{8BCA8B} 
                      \multicolumn{5}{|c|}{\bf{AUSTRALIA}} \\
\hline
ACD           & \textit{0.79 $\pm 0.019$}  & \textit{0.79 $\pm 0.018$}  & \textit{0.83 $\pm 0.015$}   & \textit{0.79 $\pm 0.012$} \\\hline
K-ACD         & \textbf{0.92 $\pm 0.010$}  & \textbf{0.82 $\pm 0.019$}  & \textbf{0.83 $\pm 0.049$}   & \textbf{0.89 $\pm 0.010$} \\\hline
EC-ACD       & \textit{0.79 $\pm 0.019$}  & \textit{0.80 $\pm 0.018$}  & \textit{0.83 $\pm 0.015$}   & \textit{0.80 $\pm 0.012$} \\\hline
K-EC-ACD     & \textbf{0.92 $\pm 0.010$}  & \textbf{0.86 $\pm 0.016$}  & \textbf{0.95 $\pm 0.008$}   & \textbf{0.87 $\pm 0.038$} \\\hline\hline

\rowcolor[HTML]{8BCA8B}                      \multicolumn{5}{|c|}{\bf{CALIFORNIA}} \\
\hline
ACD           & \textit{0.50 $\pm 0.015$}   & \textit{0.59 $\pm 0.017$}    & \textit{0.65 $\pm 0.018$}    & \textit{0.81 $\pm 0.014$} \\ \hline
K-ACD         & \textbf{0.61 $\pm 0.024$}   & \textbf{0.71 $\pm 0.048$}    & \textbf{0.85 $\pm 0.022$}    & \textbf{0.84 $\pm 0.013$} \\ \hline
EC-ACD       & \textit{0.50 $\pm 0.015$}   & \textit{0.59 $\pm 0.016$}    & \textit{0.66 $\pm 0.024$}    & \textit{0.82 $\pm 0.016$} \\\hline 
K-EC-ACD     & \textbf{0.61 $\pm 0.024$}   & \textbf{0.71 $\pm 0.047$}    & \textbf{0.85 $\pm 0.022$}    & \textbf{0.84 $\pm 0.013$} \\ \hline\hline

           \rowcolor[HTML]{8BCA8B}         \multicolumn{5}{|c|}{\bf{DENVER}} \\
\hline
ACD           & \textit{0.95 $\pm 0.013$}   & \textit{0.94 $\pm 0.014$}  & \textit{0.82 $\pm 0.059$}   & \textit{0.75 $\pm 0.058$} \\ \hline
K-ACD         & \textbf{0.96 $\pm 0.023$}   & \textbf{*0.94 $\pm 0.050$}  & \textbf{0.87 $\pm 0.017$}   & \textbf{0.96 $\pm 0.017$} \\ \hline
EC-ACD       & \textit{0.95 $\pm 0.013$}   & \textit{0.95 $\pm 0.011$}  & \textit{0.88 $\pm 0.027$}   & \textit{0.89 $\pm 0.023$} \\ \hline
K-EC-ACD     & \textbf{0.96 $\pm 0.019$}   & \textbf{*0.95 $\pm 0.037$}  & \textbf{0.97 $\pm 0.018$}   & \textbf{0.97$\pm 0.018$} \\ \hline\hline

             \rowcolor[HTML]{8BCA8B}        \multicolumn{5}{|c|}{\bf{POOPO LAKE}} \\
\hline
ACD           & \textit{0.99 $\pm 0.002$}   & \textit{0.98 $\pm 0.003$}  & \textit{0.96 $\pm 0.007$}    & \textit{0.63 $\pm 0.032$} \\ \hline
K-ACD         & \textbf{0.99 $\pm 0.002$}   & \textbf{*0.97$\pm 0.044$} & \textbf{0.96 $\pm 0.007$}     & \textbf{0.96 $\pm 0.005$} \\ \hline
EC-ACD       & \textit{0.99 $\pm 0.002$}   & \textit{0.98 $\pm 0.004$}  & \textit{0.97 $\pm 0.006$}    & \textit{0.79 $\pm 0.034$} \\ \hline
K-EC-ACD     & \textbf{0.99 $\pm 0.002$}   & \textbf{0.98 $\pm 0.013$}  & \textbf{0.99 $\pm 0.002$}    & \textbf{0.98 $\pm 0.004$} \\ \hline

\end{tabular}
\end{table}

  \begin{figure*}
     \centering
     \includegraphics[width= 4cm, height = 4cm]{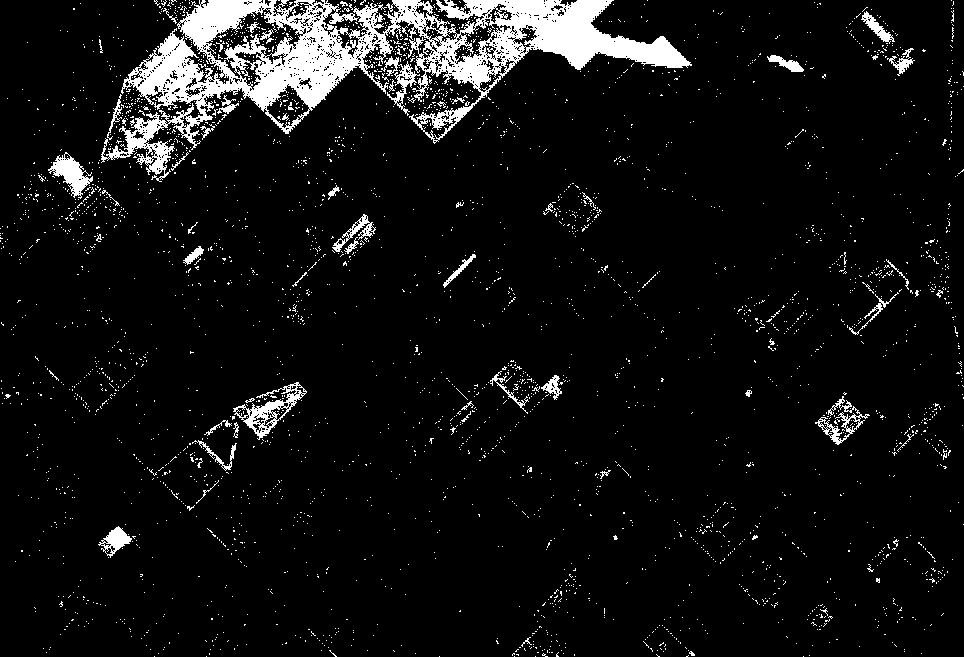} 
     \hspace{0.01mm}
     \includegraphics[width= 4cm,height = 4cm]{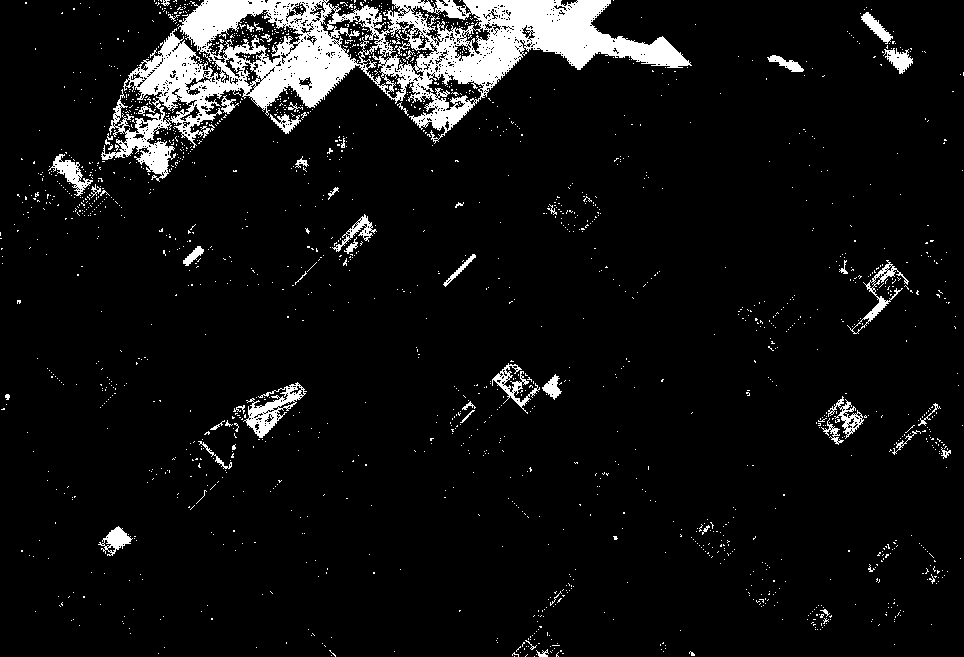}
     \hspace{0.01mm}
     \includegraphics[width= 4cm,height = 4cm]{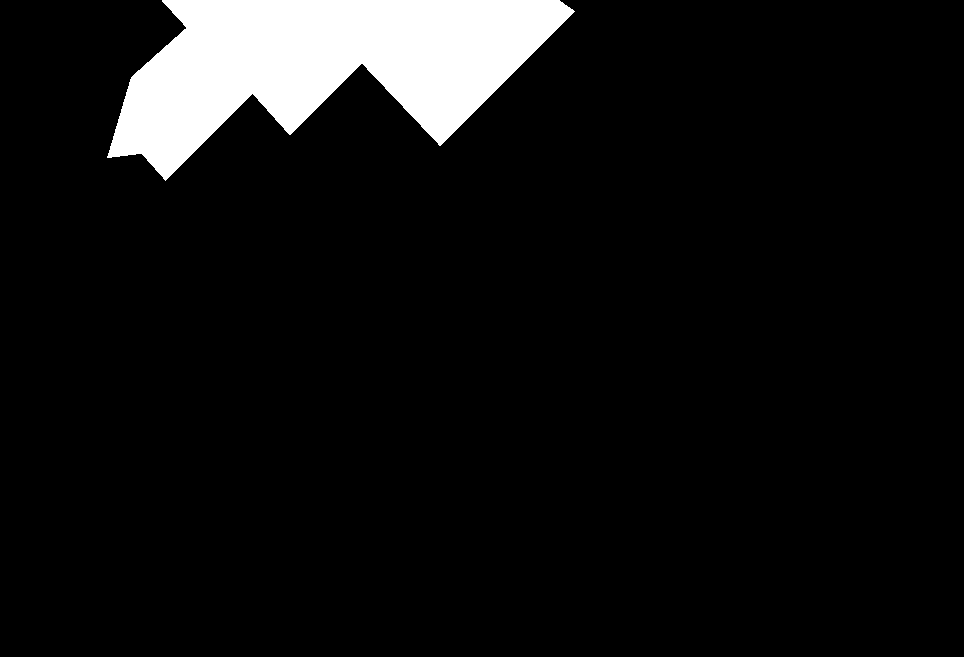} \\
     \vspace{0.2cm}
     \includegraphics[width= 4cm,height = 4cm]{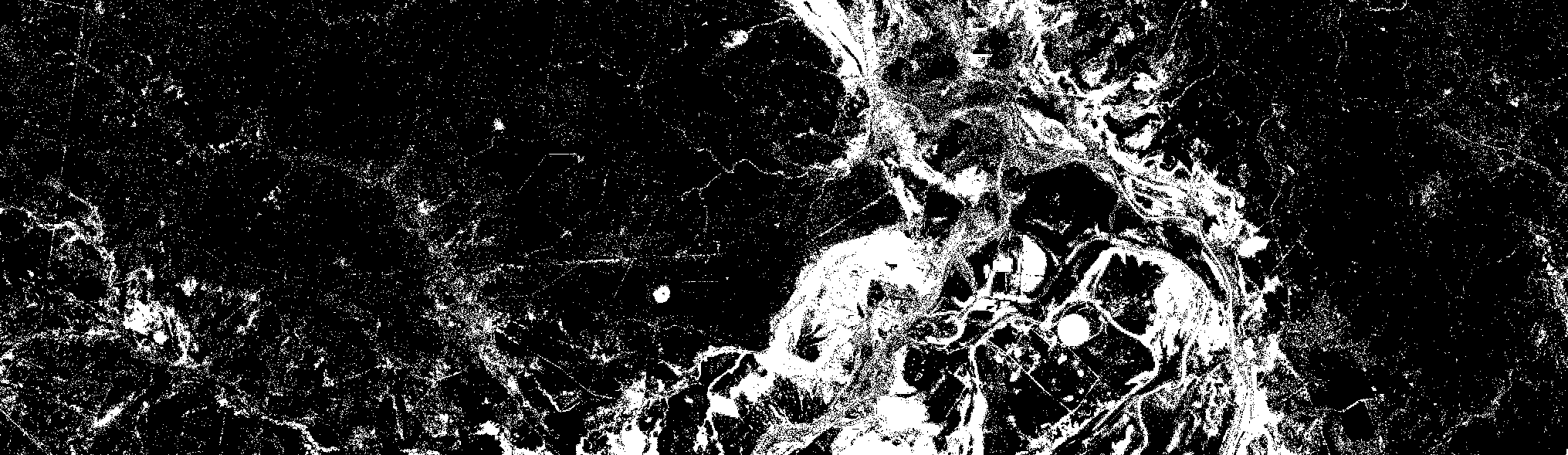} 
     \hspace{0.01mm}
     \includegraphics[width= 4cm,height = 4cm]{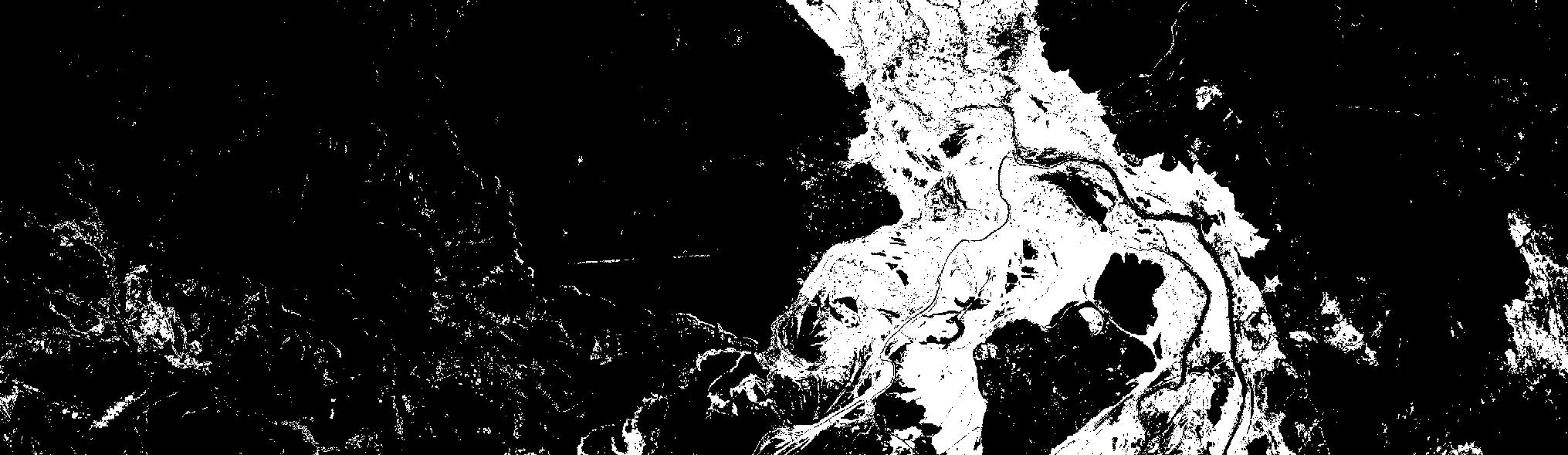} 
     \hspace{0.01mm}
     \includegraphics[width= 4cm,height = 4cm]{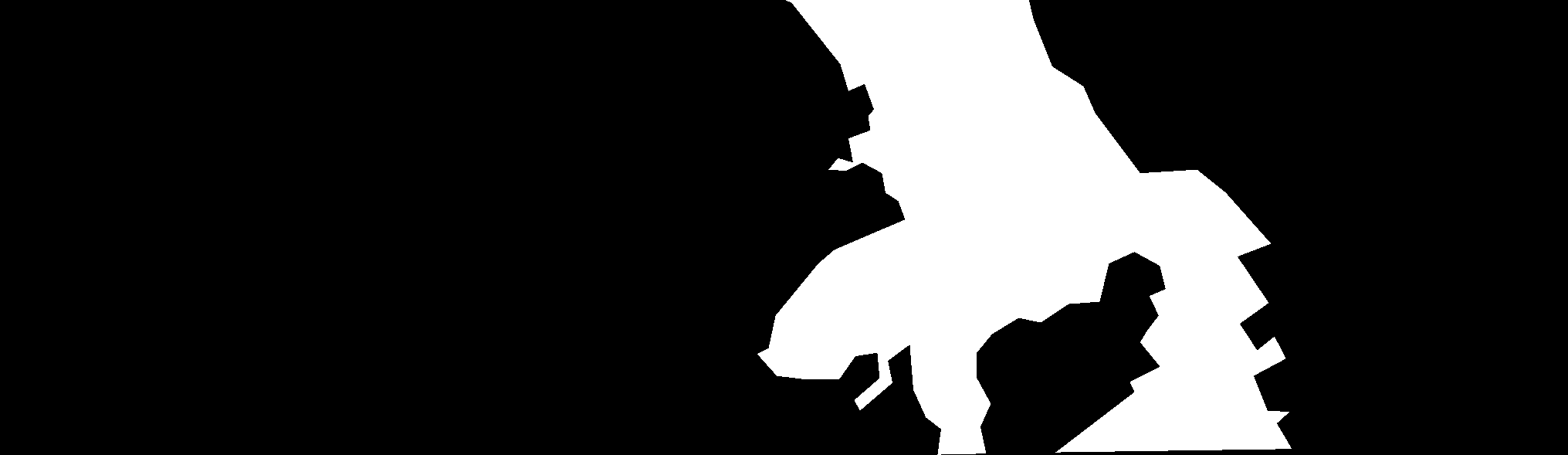} \\
     \vspace{0.2cm}
     \includegraphics[width= 4cm,height = 4cm]{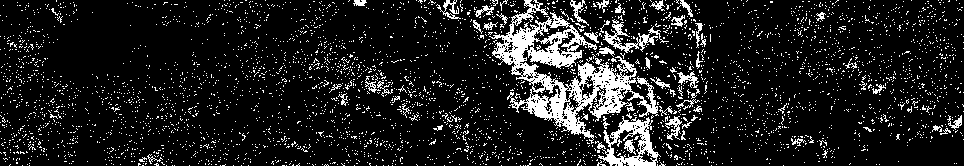} 
     \hspace{0.01mm}
     \includegraphics[width= 4cm,height = 4cm]{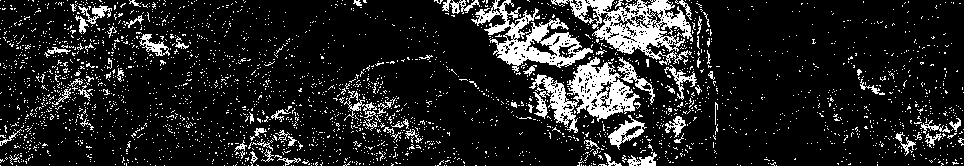} 
     \hspace{0.01mm}
     \includegraphics[width= 4cm,height = 4cm]{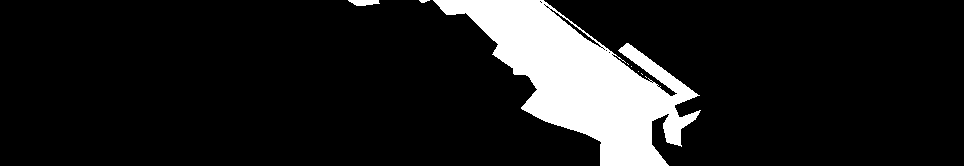} \\
     \vspace{0.2cm}
     \includegraphics[width= 4cm,height = 4cm]{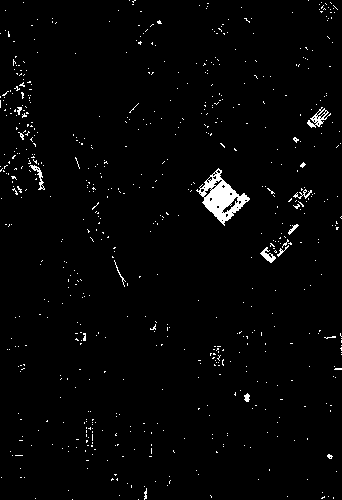} 
     \hspace{0.01mm}
     \includegraphics[width= 4cm,height = 4cm]{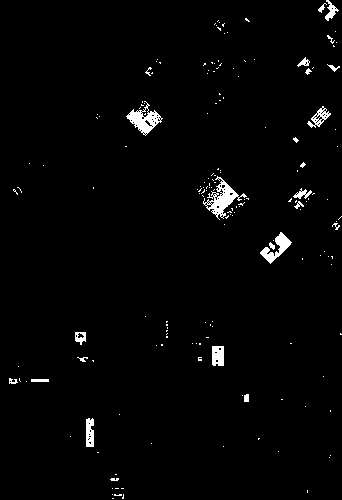} 
     \hspace{0.01mm}
     \includegraphics[width= 4cm,height = 4cm]{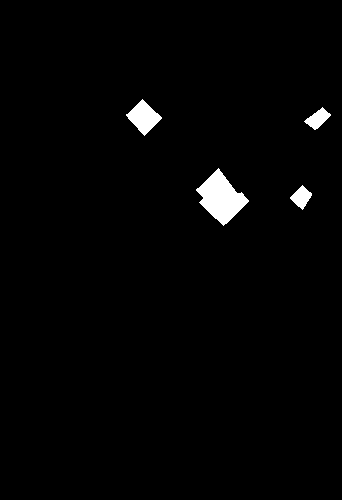} \\
     \vspace{0.2cm}
     \includegraphics[width= 4cm,height = 4cm]{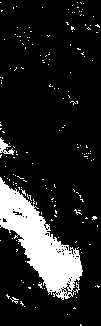} 
     \hspace{0.01mm}
     \includegraphics[width= 4cm,height = 4cm]{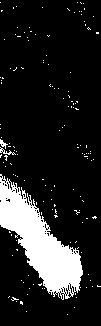} 
     \hspace{0.01mm}
     \includegraphics[width= 4cm,height = 4cm]{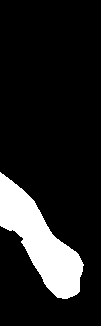} \\
    \vspace{0.2cm}
      
    
     \caption{Predictions Maps from best linear and kernel methods. Left column: best linear case, Center column: best kernel case and Right column: ground truth.}
     \label{fig:Pred2}
 \end{figure*}

\section{Conclusions}\label{sec:conclusions}

We introduced a family of kernel-based anomaly change detection algorithms. The family extends standard methods like the RX detector~\cite{Ree90,Lu97}, and many others in the literature~\cite{Theiler06,Theiler10}. The key in the proposed methodology is to redefine the anomaly detection in a reproducing kernel Hilbert space where the data are mapped to. This endorses the methods with improved capacity and flexibility since nonlinear feature relations (and hence anomalies) can be identified. The introduced methods generalize the previous ones since they account for higher-order dependencies between features. 
We provided implementations of the methods and a database of pairs of images with anomalous changes that can be found in real scenarios\footnote{\url{http://isp.uv.es/kacd.html}}. 

In practical terms, kernel ACD methods presented here yielded improved results over their linear counterparts in multi- and hyperspectral images in both real and simulated, pervasive and anomalous, changes. Results in a wide range of problems with varying complexity (control or not over the anomaly), dimensionality (multi- and hyperspectral sensors), as well as exposure to real and/or pervasive anomalies were presented. We adopted standard metrics for measure and compare methods robustness (AUC and detection) and averaged results over several runs to avoid skewed conclusions. 

Interestingly, the EC assumption may be still valid in Hilbert spaces, especially when high pervasive distortions mask anomalous targets. This observation opens the door to the study of the anomalies distribution in Hilbert spaces in the future. A second important conclusion of this work to be highlighted is that, among all 16 methods implemented, we did not observe an overall winner in all situations. After all, each problem has its own characteristics and the different methods adapt to different particularities. In the future we plan to extend the study with low-rank, sparse and scalable kernel versions to cope with high computational requirements.

\section*{Acknowledgements}

Research funded by the European Research Council (ERC) under the ERC-CoG-2014 SEDAL project (grant agreement 647423) and the Spainish Ministry of Economy, Industry and Competitiveness under the `Network of Excellence' program (grant code TEC2016-81900-REDT). Jose A. Padr\'on was supported by the Grisolia grant from Generalitat Valenciana (GVA) with code GRISOLIA/2016/100.
\bibliographystyle{plain}
\bibliography{target,BOOKbib}

\end{document}